\documentclass[a4paper,twoside]{article}

\usepackage{subcaption}
\usepackage{calc}
\usepackage{amstext}
\usepackage{amsthm}
\usepackage{multicol}
\usepackage{pslatex}
\usepackage{apalike}
\usepackage{algorithm2e}
\usepackage[bottom]{footmisc}
\usepackage{amsmath,amsfonts,amssymb}
\usepackage{graphicx}
\usepackage[table, dvipsnames]{xcolor}
\usepackage{epsfig}
\usepackage{algorithmic}
\usepackage{array}
\usepackage{arydshln}
\usepackage{textcomp}
\usepackage{stfloats}
\usepackage{url}
\usepackage{verbatim}
\usepackage{balance}
\usepackage[export]{adjustbox}
\usepackage{hyperref}
\usepackage{times}
\usepackage{booktabs}
\usepackage{tabularx}
\usepackage{multirow}
\usepackage{SCITEPRESS}  

\begin{document}

\title{Investigating Color Illusions from the Perspective of Computational Color Constancy}

\author{\authorname{Oguzhan Ulucan, Diclehan Ulucan and Marc Ebner}
\affiliation{Institut für Mathematik und Informatik, Universität Greifswald, Germany}
\email{\{oguzhan.ulucan, diclehan.ulucan, marc.ebner\}@uni-greifswald.de}
}

\keywords{Computational color constancy, color assimilation illusions, color illusion perception}

\abstract{Color constancy and color illusion perception are two phenomena occurring in the human visual system, which can help us reveal unknown mechanisms of human perception. For decades computer vision scientists have developed numerous color constancy methods, which estimate the reflectance of the surface by~\textit{discounting} the illuminant. However, color illusions have not been analyzed in detail in the field of computational color constancy, which we find surprising since the relationship they share is significant and may let us design more robust systems. We argue that any model that can reproduce our sensation on color illusions should also be able to provide pixel-wise estimates of the light source. In other words, we suggest that the analysis of color illusions helps us to improve the performance of the existing global color constancy methods, and enable them to provide pixel-wise estimates for scenes illuminated by multiple light sources. In this study, we share the outcomes of our investigation in which we take several color constancy methods and modify them to reproduce the behavior of the human visual system on color illusions. Also, we show that parameters purely extracted from illusions are able to improve the performance of color constancy methods. A noteworthy outcome is that our strategy based on the investigation of color illusions outperforms the state-of-the-art methods that are specifically designed to transform global color constancy algorithms into multi-illuminant algorithms.}

\onecolumn \maketitle \normalsize \vfill

\section{\uppercase{Introduction}} \label{sec:intro}
More than $20\%$ of the human brain is devoted to visual processing~\cite{Sheth/Young:2016}. Color vision might be the~\textit{simplest} visual attribute to understand as stated by Semir Zeki~\cite{Zeki:1993}, hence understanding color processing might be the key to unraveling how the brain works as a whole. Color illusion perception and color constancy are two phenomena related to color processing, whose mechanisms are still not entirely discovered. These phenomena can help us to reveal unknown mechanisms of the brain and let us design artificial systems one step closer to mimicking the human visual system. 

\begin{figure}
\centering
\setlength{\tabcolsep}{1.5pt} %
\renewcommand{\arraystretch}{1} 
\begin{tabular}{c c}
    \multicolumn{1}{c}{Assimilation Illusion} 
    & \multicolumn{1}{c}{Target Region} 
    \\
    \includegraphics[width=.468\linewidth, height = 2cm]{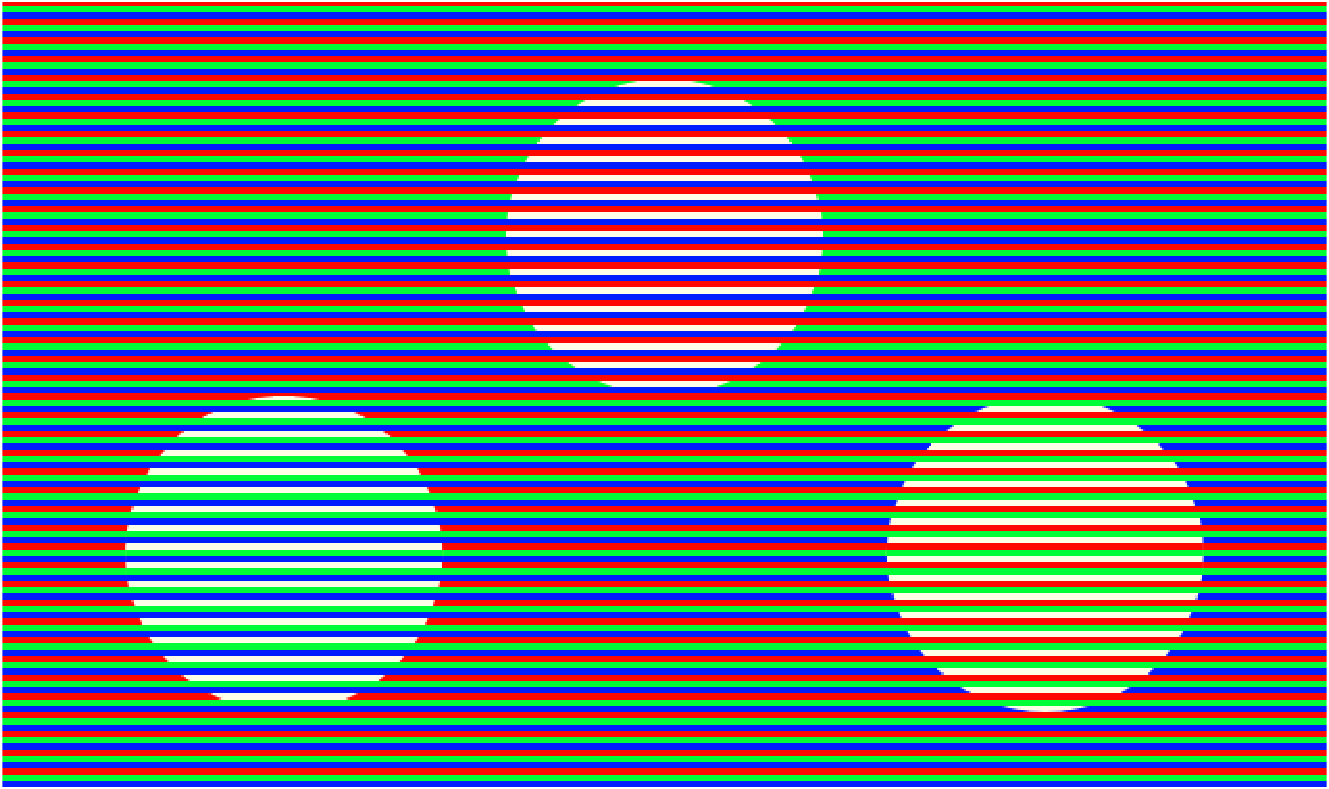}
     &\includegraphics[width=.468\linewidth, height = 2cm]{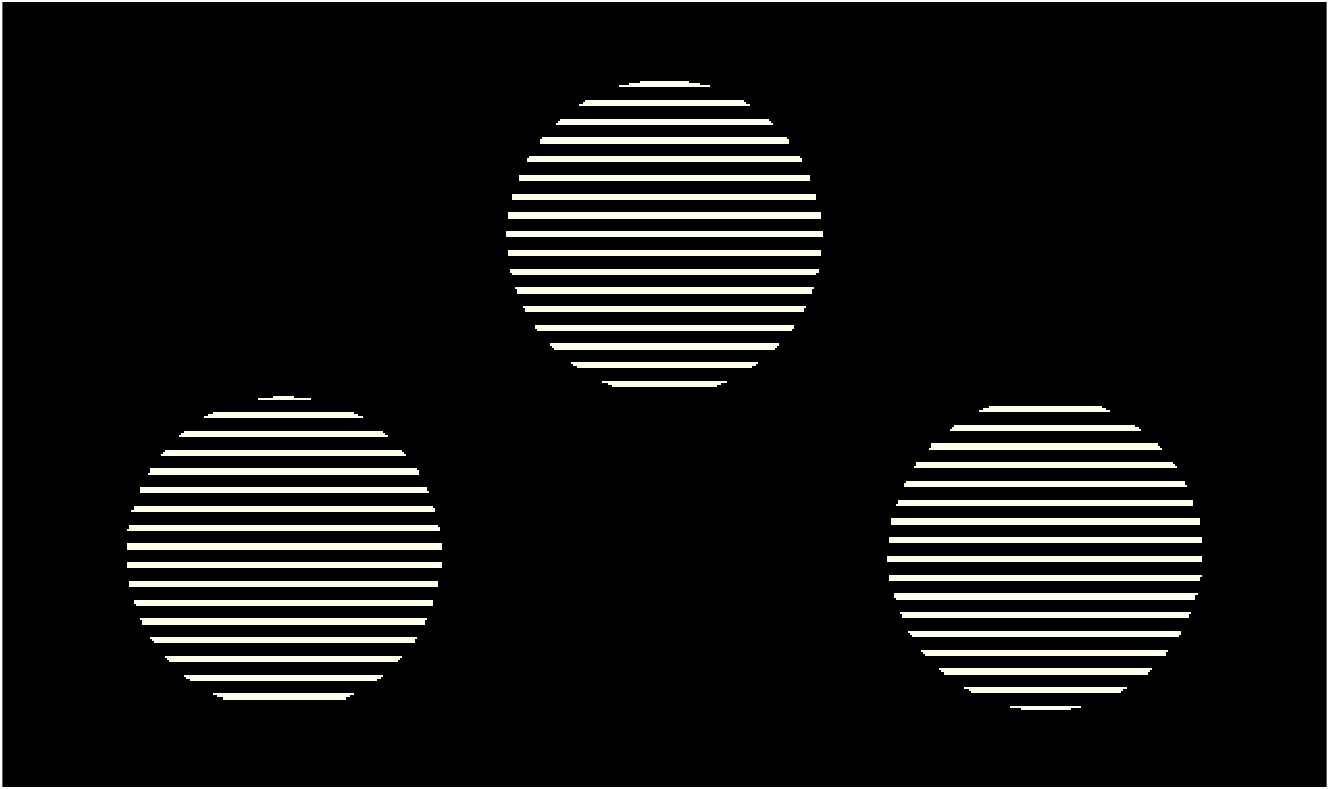}
\end{tabular}
\caption{Color assimilation illusion~\cite{BachWP}. Although we perceive the colors of the disks as if they have different colors, they are in fact the same when we remove the inducers, i.e., context, as shown in the target image.}
\label{fig:teaser}
\end{figure}

One of the interesting facts of color processing is that under some circumstances, the perceived color can be quite different than the actual physical reflectance of the object. The way we are~\textit{fooled} by color illusions can be given as an example of the difference between the perceived color and actual reflectance of an object. An example color illusion can be seen in Fig.~\ref{fig:teaser}. We perceive the colors of the disks as bluish, pinkish, and yellowish, while all the disks have the same reflectance. The reason behind this illusion, called the color assimilation illusion, is that the perceived color of the target shifts towards that of its local neighbors.

Another fascinating aspect of color processing is that the human visual system estimates the reflectance of a scene by discounting the illuminant. Hence, the perceived color of an object remains constant regardless of the illumination conditions~\cite{Gegenfurtner:1999,Brainard/Radonjic:2004,Ebner:2007,Hurlbert:2007}. This phenomenon is called color constancy. For years, researchers in the fields of neuroscience and computer vision have carried out numerous studies related to color constancy~\cite{Land:1977,Buchsbaum:1980,Brainard/Longere/Delahunt/Freeman/Kraft/Xiao:2006,Corney/Lotto:2007,Weijer/Gevers/Gijsenij:2007,Ebner:2011,Dixon/Shapiro:2017,Qian/Kamarainen/Nikkanen/Matas:2019,Afifi/Barron/LeGendre/Tsai/Bleibel:2021}. While in the former it is desired to understand human color constancy, in the latter it is aimed to estimate the illuminant of the scene.
 
The connection between color illusions and color constancy has been pointed out in both computational biology and computer vision~\cite{Marini/Rizzi:2000,Corney/Lotto:2007,Gomez/Martin/Vazquez/Bertalmio:2019,Ulucan/Ulucan/Ebner:2022a}, but never investigated together deeply in the field of computational color constancy. As presented in computational biology studies, it is important to both reproduce the behavior of the human visual system on color illusions and to perform color constancy~\cite{Corney/Lotto:2007}. Yet, computational biology studies aiming at explaining color illusion perception only accomplish to mimic the human visual system's response to color illusions but they do not provide any detailed analysis on illuminant estimation~\cite{Marini/Rizzi:1997,Marini/Rizzi:2000}, whereas computational color constancy studies only aim at estimating the illuminant. We find it rather surprising that even though color constancy methods can estimate the illuminant, the analysis of color illusions is neglected, which could provide significant benefits. In this study, by using a learning-free approach we attempt to show that there is indeed a relationship between color illusions and color constancy, in particular between the color assimilation illusions and multi-illuminant color constancy since the relationship between the spatially close regions is important for both phenomena. We argue that if we analyze color illusions from the perspective of computational color constancy, we can highlight the natural link between these two phenomena and learn from the outcomes of this investigation to modify global color constancy algorithms so that they can provide pixel-wise estimations, and hence perform multi-illuminant color constancy. Also, we argue that by using the results of the investigation of color illusions we can enhance the efficiency of the global color constancy algorithms. Furthermore, since color illusions and color constancy are two important phenomena that occur due to the processing of the human brain, the outcomes of investigating these phenomena together may help us to design artificial systems that are one step closer to mimicking human visual perception~\cite{Bach/Poloschek:2006,Gomez/Martin/Vazquez/Bertalmio:2019}. It is important to note here that despite numerous studies conducted in this field, the biological explanation for the link between color illusions and color constancy is still unknown. If we could explain this link biologically as well as how the human visual system discounts the illuminant and is fooled by the illusions, then we would basically have a correct model of human color processing, hence, we would be able to develop algorithms for both digital photography and computer vision applications that would satisfy the consumers around the world~\cite{Ebner:2009}.

To the best of our knowledge, this is the first study in the field of computational color constancy that stresses the importance of the relationship between color assimilation illusions and color constancy in detail.

Our contributions can be summarized as follows;
\begin{itemize}
    \item We investigate color illusions from the perspective of computational color constancy algorithms.
    \item We introduce a simple yet effective approach based on our findings from the investigation of color illusions that transforms the global color constancy algorithms into multi-illuminant color constancy methods, which do not require any information about the scene.
    \item We show that our approach improves the performance of existing color constancy methods. 
\end{itemize}

This paper is organized as follows. In Sec.~\ref{sec:relwork}, we briefly explain the field of computational color constancy. In Sec.~\ref{sec:prop}, we introduce our approach. In Sec.~\ref{exp: illusions}, we discuss the results for color assimilation illusions, and in Sec.~\ref{exp: colorconstancy} we provide the outcomes for both multiple and single illuminant color constancy. In Sec.~\ref{sec:conc}, we summarize our work.

\section{\uppercase{Background}} \label{sec:relwork}

Before introducing our proposed approach we would like to give a brief summary of the widely used image formation model in the field of color constancy and color constancy algorithms since throughout the paper we will investigate color illusions from the perspective of color constancy algorithms. 

\subsection{Theory of Color Image Formation}

We begin to process visual information when the light entering the eye is measured by the photoreceptors, e.g., cones, present in the retina, whereas artificial systems start to process visual information when an array of sensors measures the incident light. In case the camera contains three different sensors, then each sensor responds to a specific light from the distinct parts of the visible spectrum, i.e., short-, middle-, and long-wavelength. Let us suppose that the camera sensors are capturing every spatial location of the scene. Then, the measured energy of the signal at the spatial location $(x,y)$ can be formulated as follows;
\begin{equation}
    I(x,y) = \int_{w} E(x,y;\lambda) S_i(\lambda) d\lambda
\end{equation}
where, $E(x,y;\lambda)$ is the irradiance hitting the sensors of the capturing device, and $S_i(\lambda)$ is the sensor specifications of the camera that contains the responses of the sensors for a specific wavelength, with $i \in$ \{long, middle, short\}, and $\lambda$ is the wavelength of the visible spectrum.

In color constancy studies, we mostly assume that the surface is equally reflecting the light into all directions, i.e., Lambartian surface, and the scene is illuminated by a point light source $L(x,y;\lambda)$. Thus, the irradiance hitting the sensors of the camera can be expressed as follows; 
\begin{equation}
    E(x,y;\lambda) = G(x,y) R(x,y;\lambda) L(x,y;\lambda)
\end{equation}
where, $R(x,y;\lambda)$ is the (shaded) reflectance, and $G(x,y)$ is the scaling factor that can be expressed as $cos~\alpha$, where $\alpha$ is the angle between the surface normal vector and a vector pointing in the direction of the light source. 

Consequently, an image captured by the camera is generally modeled by using the model of Lambertian image formation, and it can be formulated as follows;
\begin{equation} \label{color_constancy}
    I(x,y) = G(x,y) \int_{w} R(x,y;\lambda) L(x,y;\lambda) S_i(\lambda)d\lambda.
\end{equation}

The aim of color constancy is to estimate the illuminant $L$ from the input image $I$ with a color cast. However, as we can see from the Eqn.~\ref{color_constancy}, color constancy is framed as a computational challenge, i.e., an ill-posed problem, since it depends on both the type of the sensors and the illumination, which we cannot be sure of. Therefore, most of the algorithms make relaxations to the ill-posed nature of color constancy by assuming that the camera sensor responses are narrow-band, i.e., they can be approximated by Dirac’s delta functions, and the scene is uniformly illuminated. Moreover, most of the algorithms do not consider the scene geometry $G(x,y)$. Thus, the image can be formed as the element-wise product of the reflectance $R$ and the global illumination source $\mathbf{L}$ as follows; 
\begin{equation}
    I(x,y) =  R(x,y) \cdot \mathbf{L}.
    \label{hadamard}
\end{equation}

It is noteworthy to point out that even though these assumptions are beneficial to tackle the ill-posed nature of color constancy to estimate the color vector of the light source, assuming that the geometry factor throughout the scene is uniform and there is a single light source illuminating the scene are strong assumptions which are usually violated in the real world due to the presence of interreflections, shadows, and multiple light sources in the scene~\cite{Ershov/Tesalin/Ermakov/Brown:2023}.

\subsection{Related Work}

Over the decades, numerous global color constancy algorithms have been proposed. The white-patch Retinex and the gray world algorithm are well-known methods, which are inspired by the mechanisms of the human visual system since the human visual system might be discounting the illuminant of the scene based on the highest luminance patch, and space-average color~\cite{Land:1977,Buchsbaum:1980}. The white-patch Retinex method computes the maximum responses of the image channels separately to estimate the illuminant of the scene, while the gray world algorithm averages the pixels of each channel independently to output an illuminant estimate. These two methods have been modified and used in several other studies due to their simplicity and efficacy. For example, the shades of gray algorithm assumes that the mean of pixels raised to a certain power is gray~\cite{Finlayson/Trezzi:2004}. The gray edge method and weighted gray edge algorithm stress that the gradient features of the image are beneficial cues for estimating the illuminant of the scene~\cite{Weijer/Gevers/Gijsenij:2007,Gijsenij/Gevers/Weijer:2009}. The mean-shifted gray pixel method detects and uses the gray pixels for illumination estimation~\cite{Qian/Pertuz/Nikkanen/Kamarainen/Matas:2018}. The block-based color constancy method divides the image into non-overlapping blocks and transforms the task of illumination estimation into an optimization problem by making use of both the gray world and the white-patch Retinex assumptions~\cite{Ulucan/Ulucan/Ebner:2022b,Ulucan/Ulucan/Ebner:2023,ulucan2023multi}. The double-opponency-based color constancy algorithm is inspired by the mechanisms of the human visual system, i.e., physiological findings on color information processing~\cite{Gao/Yang/Li/Li:2015}. The principal component analysis-based color constancy method considers only the informative pixels, which are assumed to be the ones having the largest gradient in the data matrix, to estimate the color vector of the illuminant~\cite{Cheng/Prasad/Brown:2014}. The local surface reflectance statistics algorithm is based on the biological findings about the feedback modulation mechanism in the eye and the linear image formation model~\cite{Gao/Han/Yang/Li/Li:2014}. The biologically inspired color constancy method is based on the hierarchical structure of the perception mechanism of the human vision system~\cite{Ulucan/Ulucan/Ebner:2022a}. 

While most of the aforementioned global color constancy algorithms are traditional methods, there also exist learning-based algorithms~\cite{Afifi/Brown:2019,Laakom/Raitoharju/Iosifidis/Nikkanen/Gabbouj:2019,Afifi/Brown:2020,Afifi/Barron/LeGendre/Tsai/Bleibel:2021,Afifi/Brubaker/Brown:2022,Domislovic/Vrsnak/Subasic/Loncarc:2022}. As discussed in several studies,  these algorithms outperform classical methods on well-known benchmarks. However, their effectiveness tends to decrease when they face images with different statistical distributions and/or images captured by cameras with unknown specifications~\cite{Gao/Zhang/Li/Li:2017,Qian/Kamarainen/Nikkanen/Matas:2019,Ulucan/Ulucan/Ebner:2022b}. The reason behind this performance decline can be explained by the facts that~\textit{(i)} well-known benchmarks are mostly formed with cameras, whose sensor response characteristics are similar,~\textit{(ii)} illumination conditions in most datasets do not significantly differ, for instance, lights outside and on the edges of the color temperature curve are seldomly considered, and~\textit{(iii)} learning-based methods expect their training and test sets to be somehow similar~\cite{Ulucan/Ulucan/Ebner:2022b,Buzzelli/Zini/Bianco/Ciocca/Schettini/Tchobanou:2023}. 

The assumption of having a single light source enables us to estimate the illuminant of the scene, however, it is usually violated in the real world since most scenes do not have uniform illumination due to interreflections, multiple light sources, and shadows~\cite{Ebner:2009,Gijsenij/Lu/Gevers:2011,bleier2011color}. To avoid this violation both traditional and learning-based multi-illuminant color constancy algorithms have been proposed. These methods output local estimates for each spatial location in the image. For instance, local space average color is one of the first methods, which finds the pixel-wise estimates to arrive at the color constant descriptor of the scene~\cite{Ebner:2003,Ebner:2004}. The patch-based methodology and conditional random fields-based algorithm propose strategies, which can be applied to existing color constancy algorithms~\cite{Gijsenij/Lu/Gevers:2011,Beigpour/Riess/Weijer/Angelopoulou:2013}. The retinal mechanism-inspired model mimics the color processing mechanisms in a particular retina level and outputs a white-balanced image without explicitly providing an illumination estimate~\cite{Zhang/Gao/Li/Du/Li/Li:2016}. The image regional color constancy weighing factors based algorithm separates the image into multiple parts and uses the normalized average absolute difference of each part as a measure of confidence~\cite{Hussain/Akbari:2018}. The image texture-based algorithm utilizes the texture to find the pixels, which have adequate color variation and uses these to discount the illuminant of the scene~\cite{Hussain/Akbari/Halpin:2019}. The algorithm inspired by the human visual system makes use of the bottom-up and top-down mechanisms to find the color of the light source~\cite{Gao/Ren/Zhang/Li:2019}. The grayness index-based color constancy algorithm finds the gray pixels based on the dichromatic reflection model and uses these gray pixels to find both global and local illumination estimates~\cite{Qian/Kamarainen/Nikkanen/Matas:2019}. The N-white balancing method determines the number of white points in the image to estimate the illuminant of the scene~\cite{Akazawa/Kinoshita/Shiota/Kiya:2022}. The convolutional neural networks (CNNs) based approach contains a detector that can determine the number of illuminants present in the scene~\cite{Bianco/Cusano/Schettini:2017}. The physics-driven and generative adversarial networks (GAN) based method models the light source estimation task as an image-to-image domain translation problem~\cite{Das/Liu/Karaoglu/Gevers:2021}. 

The main drawback of many multi-illuminant color constancy algorithms is the assignment of the number of clusters/segments~\cite{Wang/Wang/Wu/Gao:2022}. In other words, the provision of the number of illuminants to the algorithm before processing starts. We argue that in real-world scenarios, where the number of lights might be unknown or for arbitrary images for which the number of illuminants is not provided, algorithms depending on the number of light sources may not work effectively. Therefore, a multi-illuminant color constancy approach that can estimate the illuminants without having the knowledge of the number of lights is necessary.

\section{\uppercase{Proposed Approach}} \label{sec:prop}

Global color constancy algorithms estimate the color vector of a single light source uniformly illuminating the scene. Therefore, it cannot be expected that they reproduce our behavior on color assimilation illusions without any modification. Our method modifies the global color constancy algorithms with a simple yet effective approach so that they provide pixel-wise estimates of the input image, thus reproduce our sensation on color assimilation illusions. Since locality is an important feature for both assimilation illusions and multi-illuminant color constancy, our motivation is that the parameters purely extracted from the analysis of the color illusions should be beneficial for color constancy. Our strategy can transform the global color constancy methods into algorithms that can provide local estimates of the light source illuminating the scenes.  

\begin{figure}
\centering
\setlength{\tabcolsep}{1.5pt} %
\renewcommand{\arraystretch}{0.5} 
\begin{tabular}{c c}
\multicolumn{1}{c}{Input Image} 
& \multicolumn{1}{c}{Input Target} 
\\
\includegraphics[width=.468\linewidth, height = 1.8cm]{New_Teaser/bach_disks_in.png}\quad
&\includegraphics[width=.468\linewidth, height = 1.8cm]{New_Teaser/bach_disks_tar.png}\quad
\end{tabular}
\\
\begin{tabular}{ccc}
\multicolumn{1}{c}{Sparse Image $I_{s}$}
&\multicolumn{1}{c}{Estimates} 
& \multicolumn{1}{c}{Output Target} 
\\
\includegraphics[width=.312\linewidth, height = 1.8cm]{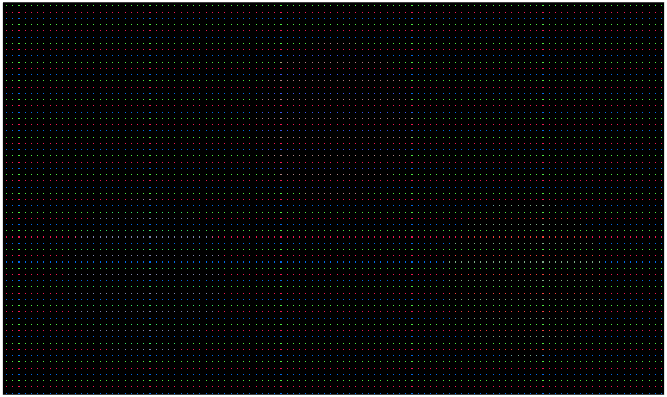}\quad
&\includegraphics[width=.312\linewidth, height = 1.8cm]{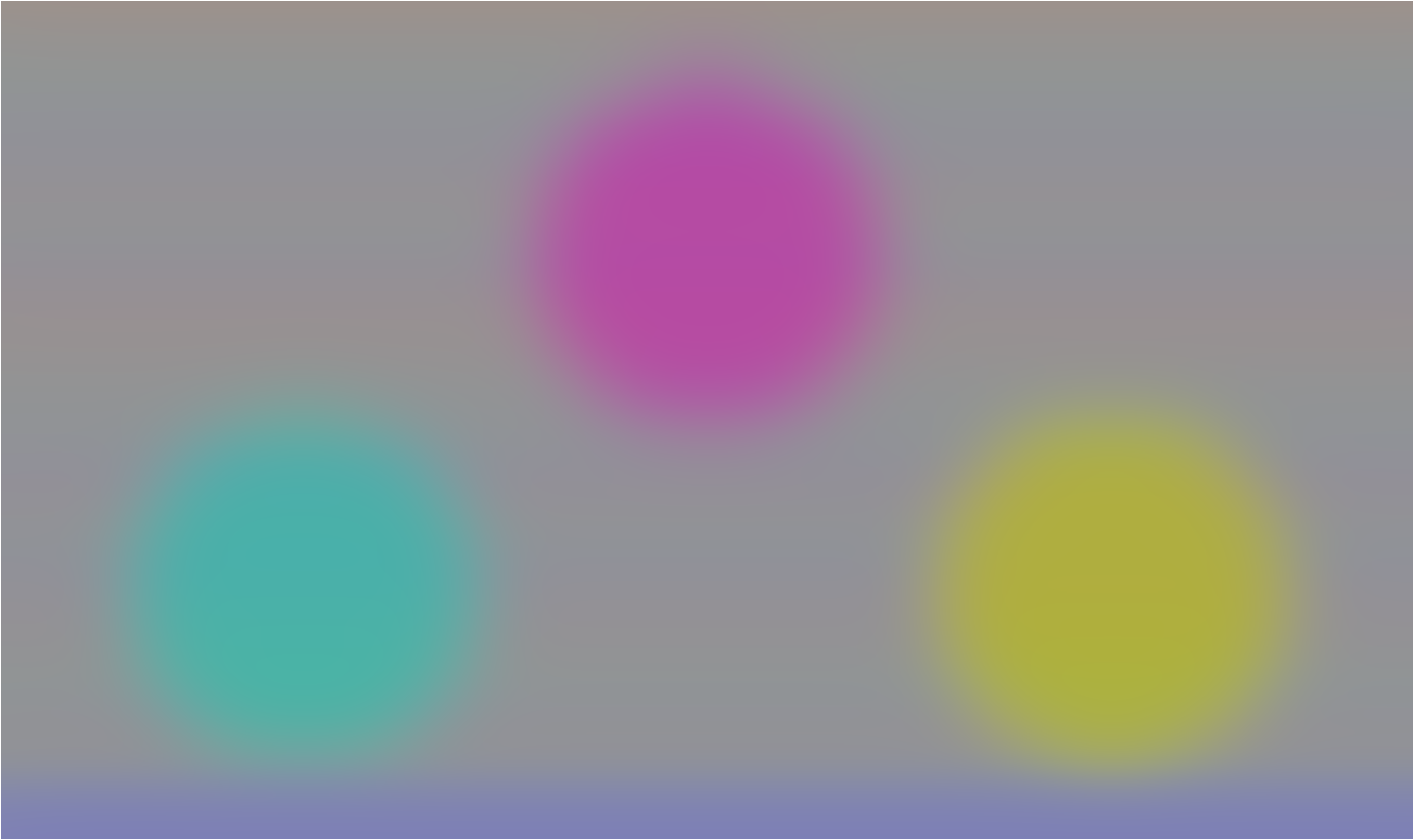}\quad
&\includegraphics[width=.312\linewidth, height = 1.8cm]{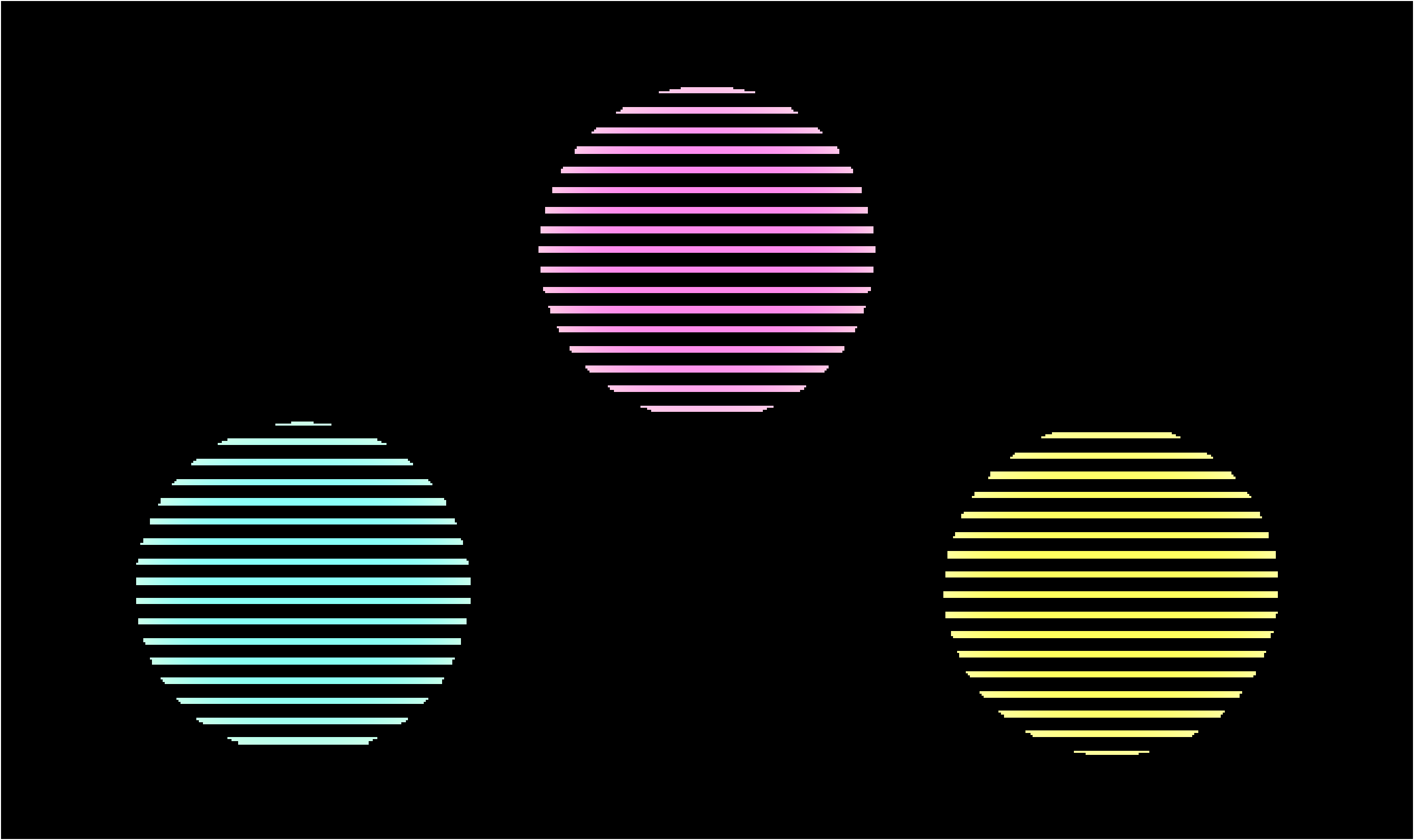}
\end{tabular}
\caption{Reproduction of color assimilation illusions. As shown in the input target, the RGB values of the disks are the same, yet we perceive them in the input image as if they are different. After we apply our method, as given in the output target the disks have colors closer to our perception.}
\label{fig: res}
\end{figure}

In our method, we take the linearized input image $I$ and divide it into $\beta \times \beta$ non-overlapping blocks, and $\beta$ is chosen as $8$ (the parameter selection is detailed in Sec.~\ref{exp: illusions}). After we divide the image into non-overlapping blocks, we apply a color constancy algorithm to each block to find the estimate of that block. Subsequently, rather than assuming that all the spatial locations in the block have uniform estimates, we place the computed local estimation into the center pixel of the corresponding block and we obtain a sparsely populated image $I_{s}$ (Fig.~\ref{fig: res}). The reason behind this operation is to avoid sharp changes between adjacent blocks, i.e., to obtain smooth transitions between blocks. Afterwards, to fill the missing information in $I_{s}$, i.e., to obtain a dense image, we perform an interpolation between the adjacent center pixels. The interpolation is carried out by convolving $I_{s}$ with a Gaussian kernel $\mathcal{G}$ (Eqn.~\ref{eqn: interpolate}), hence for each spatial location a pixel-wise estimate is obtained.   

\begin{equation}
    I_p(x,y) =  I_{s}(x,y) * \mathcal{G}(x,y;\sigma)
    \label{eqn: interpolate}
\end{equation}
where, $I_p$ contains the pixel-wise estimations normalized to unit norm, $*$ denotes the convolution operation, and $\mathcal{G} = \frac{1}{2\pi\sigma^2} exp\left(-\frac{x^2 + y^2}{2\sigma^2}\right)$. The controlling parameter $\sigma$ of $\mathcal{G}$ needs to be sufficiently large so that at least three estimates are inside of the area of support, hence it is chosen as $24$ (discussion on selecting $\sigma$ is provided in Sec.~\ref{exp: illusions}). 

Consequently, our simple yet effective approach enables us to simulate the behavior of the human visual system on color assimilation illusions by using computational color constancy algorithms as shown in Fig.~\ref{fig: res}. 

It is worth mentioning here that we only modify the traditional color constancy techniques due to the input size requirement of the learning-based methods. In other words, neural networks-based methods need a fixed input image size, while we use $8 \times 8$ blocks in our approach. During our experiments, we have observed that upsampling the block sizes does not result in satisfying outcomes, thus we do not apply our approach to neural networks-based algorithms.   

In the remainder of this section, we slightly adjust our approach for color constancy purposes. While we make our adjustments, we utilize the parameters, i.e., block size and controlling parameter of the Gaussian kernel, which we extract from our analysis on color assimilation illusions.

\subsection{Application to Color Constancy}

After we simulate the behavior of the human visual system on color assimilation illusions, we analyze whether the proposed approach relying on parameters purely extracted from color illusions can transform the global color constancy methods into multi-illumination color constancy algorithms. Also, we investigate whether our method can improve the performance of the existing color constancy methods on single illumination conditions.

It is known that not all pixels are beneficial to perform color constancy, i.e., the performance of algorithms tends to decrease when the images contain dominant sky regions. Hence, instead of using all pixels in the images, choosing the most informative ones increases the efficiency of the color constancy methods as proposed in several studies~\cite{Finlayson/Hordley:2001,Yang/Gao/Li:2015,Qian/Pertuz/Nikkanen/Kamarainen/Matas:2018,Ulucan/Ulucan/Ebner:2023}. Therefore, to adapt our proposed approach for color assimilation illusions to color constancy, we form a confidence map by using the whitest pixels, e.g., pixels having the highest luminance in the image. The reason for using the whitest pixels in the scene can be explained by biological findings, i.e., our visual system might be discounting the illuminant by making use of the areas that have the highest luminance rather than the darkest regions~\cite{Land/McCann:1971,Linnell/Foster:1997,Uchikawa/Fukuda/Kitazawa/MacLeod:2012,Morimoto/Kusuyama/Fukuda/Uchikawa:2021}, and from the perspective of digital photography, i.e., the illuminant can be determined easier from the achromatic colored areas such as white regions in the scene rather than the colored areas~\cite{Drew/Joze/Finlayson:2012,Joze/Drew/Finlayson/Rey:2012,Qian/Kamarainen/Nikkanen/Matas:2019,Ono/Kondo/Sun/Kurita/Moriuchi:2022,Ulucan/Ulucan/Ebner:2023}.

To detect the whitest pixels, we take the mean of each color channel separately and scale the input image according to this color vector. Then, we form a pixel-wise whiteness map $\mathcal{W}$ by computing the pixel-wise angular error between the scaled image $I_{temp}$ and the white vector $\mathbf{w} = [1~1~1]$ as follows; 
\begin{equation}
    \mathcal{W}(x,y) = cos^{-1} \begin{pmatrix} 
    \frac{\mathbf{w} \cdot I_{temp}(x,y)}{\left\| \mathbf{w} \right\| \left\| I_{temp}(x,y) \right\|}
\end{pmatrix}.  
\end{equation}

Subsequently, we form the confidence map $\mathcal{C}$ by adaptively weighting $\mathcal{W}$ at each spatial location using a Gaussian curve (Eqn.~\ref{eqn:confmap}) since the contribution of the brightest pixels to color constancy may not be equal throughout the scene. This approach coincides with the biological findings that brighter regions have a greater effect on human color constancy than darker regions~\cite{Uchikawa/Fukuda/Kitazawa/MacLeod:2012}. 
\begin{equation}
    \mathcal{C}(x,y) = \frac{1}{2\pi\sigma_{\mathcal{W}}^2} exp\left(-\frac{ (\mathcal{W}(x,y) - \mu_{\mathcal{W}})^2}{2\sigma_{\mathcal{W}}^2}\right)
    \label{eqn:confmap}
\end{equation}
where, $\mu_{\mathcal{W}}$ and $\sigma_{\mathcal{W}}$ are the mean and the standard deviation of $\mathcal{W}$, respectively.

The pixel-wise estimates of the light source $\hat L_{est}$ are obtained by applying an interpolation similar to Eqn.~\ref{eqn: interpolate} as follows;
\begin{equation}
     L_{est}(x,y) = \left (I_s(x,y) \times \mathcal{C}(x,y) \right) * \mathcal{G}(x,y).
\end{equation}

In global color constancy benchmarks, the color of the ground truth light vectors are provided as an $R,G,B$ triplet. Although our approach provides pixel-wise estimations, one can obtain a global estimate of the light source by averaging the pixels in $I_s$ for each color channel individually.

\section{\uppercase{Experiments on Color Illusions}} \label{exp: illusions}
In this section, we explain the formation of the color illusion set and the extraction of the parameters. Subsequently, we discuss our outcomes on the reproduction of color illusions by using various global color constancy methods. 

\subsection{Formation of the Color Illusion Set}
\begin{figure}
    \centering
    \setlength{\tabcolsep}{1.5pt} %
    \renewcommand{\arraystretch}{0.5} 
    \begin{tabular}{ccc}
    \includegraphics[width=.468\linewidth, height = 2cm]{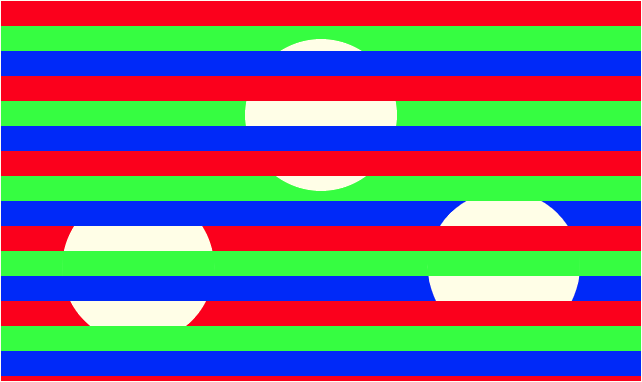} 
    &\includegraphics[width=.468\linewidth, height = 2cm]{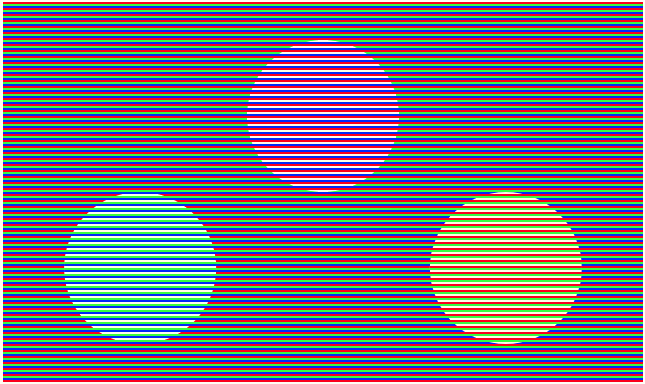} 
    \end{tabular}
    \caption{Evoking an illusion sensation highly depends on the inducer's characteristics. As given in the first column, the image having a low inducer frequency and a large inducer thickness does not evoke an illusion sensation. However, the illusion sensation is easily observable when we increase the frequency and decrease the thickness of the inducer as given in the second column.}
    \label{fig:inv_illusion}
\end{figure}

The most important two features of color assimilation illusions that affect our perception are the inducer's frequency of occurrence and its thickness. As seen in Fig.~\ref{fig:inv_illusion}, there is no illusion effect in the image given in the first column. On the other hand, illusion sensations are observable in the image present in the second column. It is important to note that while analyzing the illusions, we recommend zooming in to the figure and fixating our focal vision on each image individually. The former is recommended since when the observer views the image as small-sized or from a distance the image will appear as if its inducer has sufficient frequency to evoke an illusion sensation although there is no (or a weak) illusion sensation present in the image. The latter is advised since our peripheral vision is locally disordered~\cite{Koenderink/Doorn:2000}, i.e., blurred. This characteristic feature of the peripheral vision can ease the perception of color assimilation illusions, i.e., even a very weak illusion effect can be perceived as a strong illusion sensation, since in color assimilation illusions the color of the target region shifts towards that of its local neighbors. 

\begin{figure}
    \centering
    \setlength{\tabcolsep}{1.5pt} %
    \renewcommand{\arraystretch}{0.5} 
    \begin{tabular}{ccc}
    \includegraphics[width=.312\linewidth, height = 1.8cm]{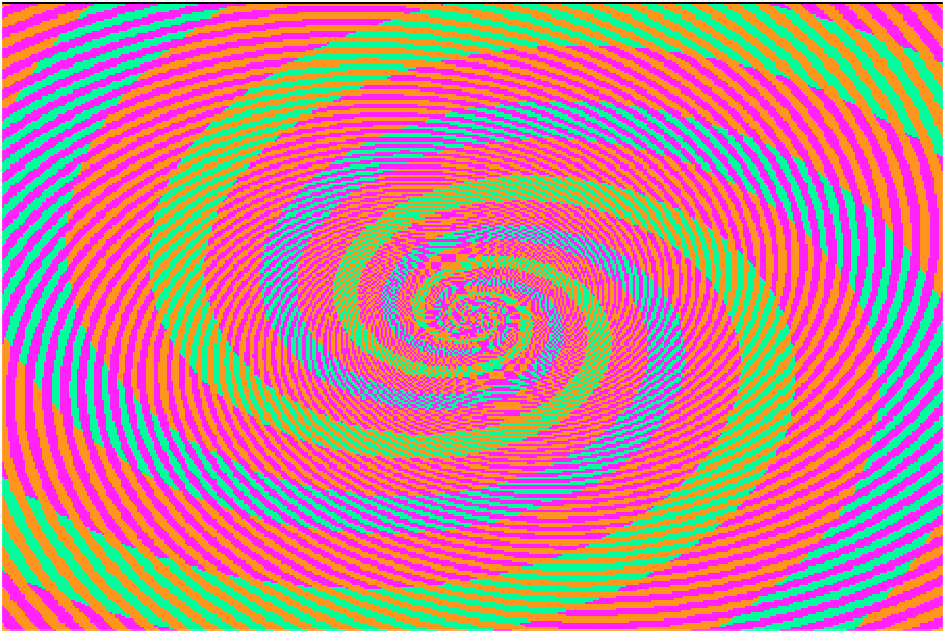}
    &\includegraphics[width=.312\linewidth, height = 1.8cm]{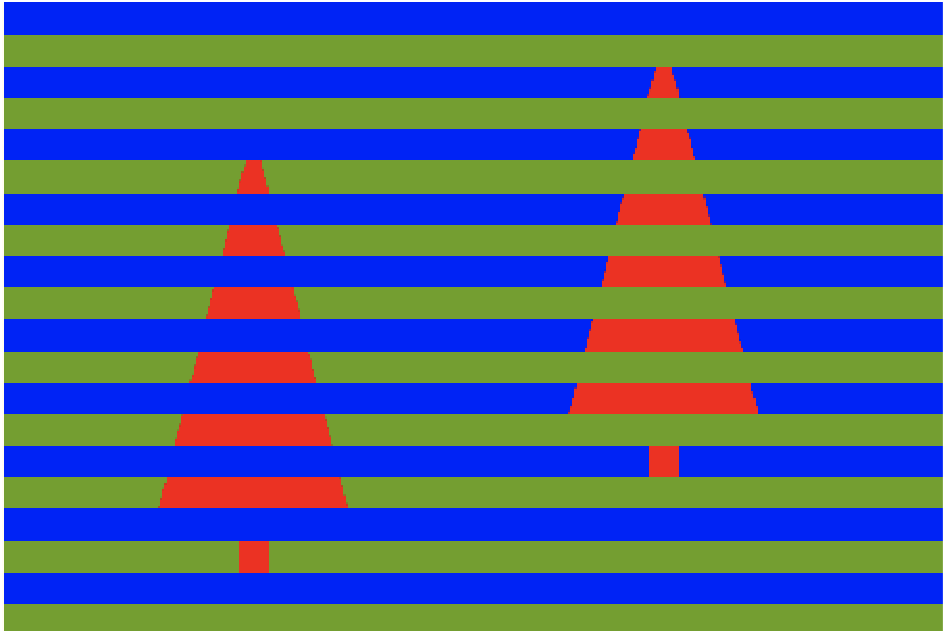}
    &\includegraphics[width=.312\linewidth, height = 1.8cm]{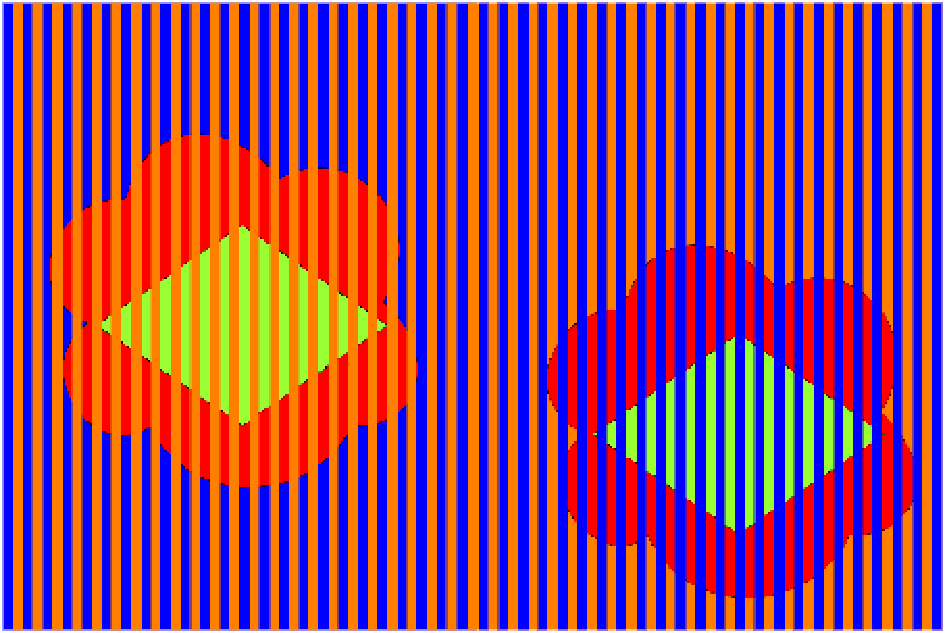}
    \\
    \includegraphics[width=.312\linewidth, height = 1.8cm]{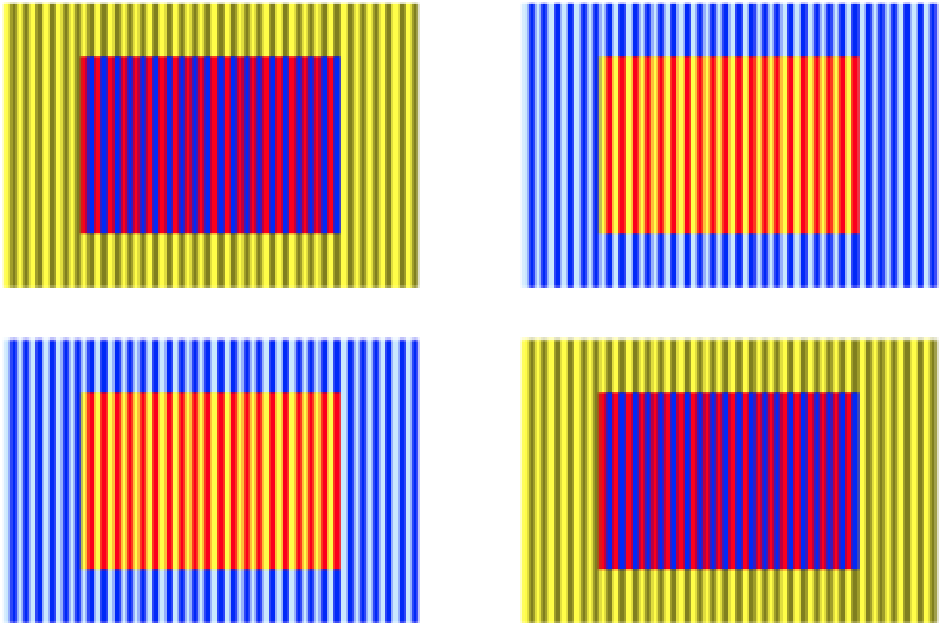} 
    &\includegraphics[width=.312\linewidth, height = 1.8cm]{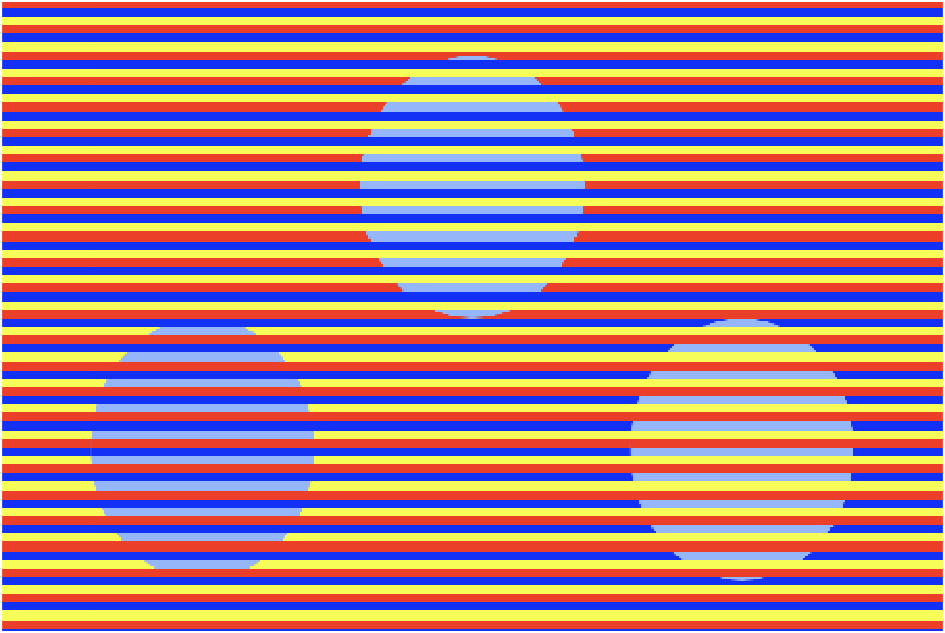} 
    &\includegraphics[width=.312\linewidth, height = 1.8cm]{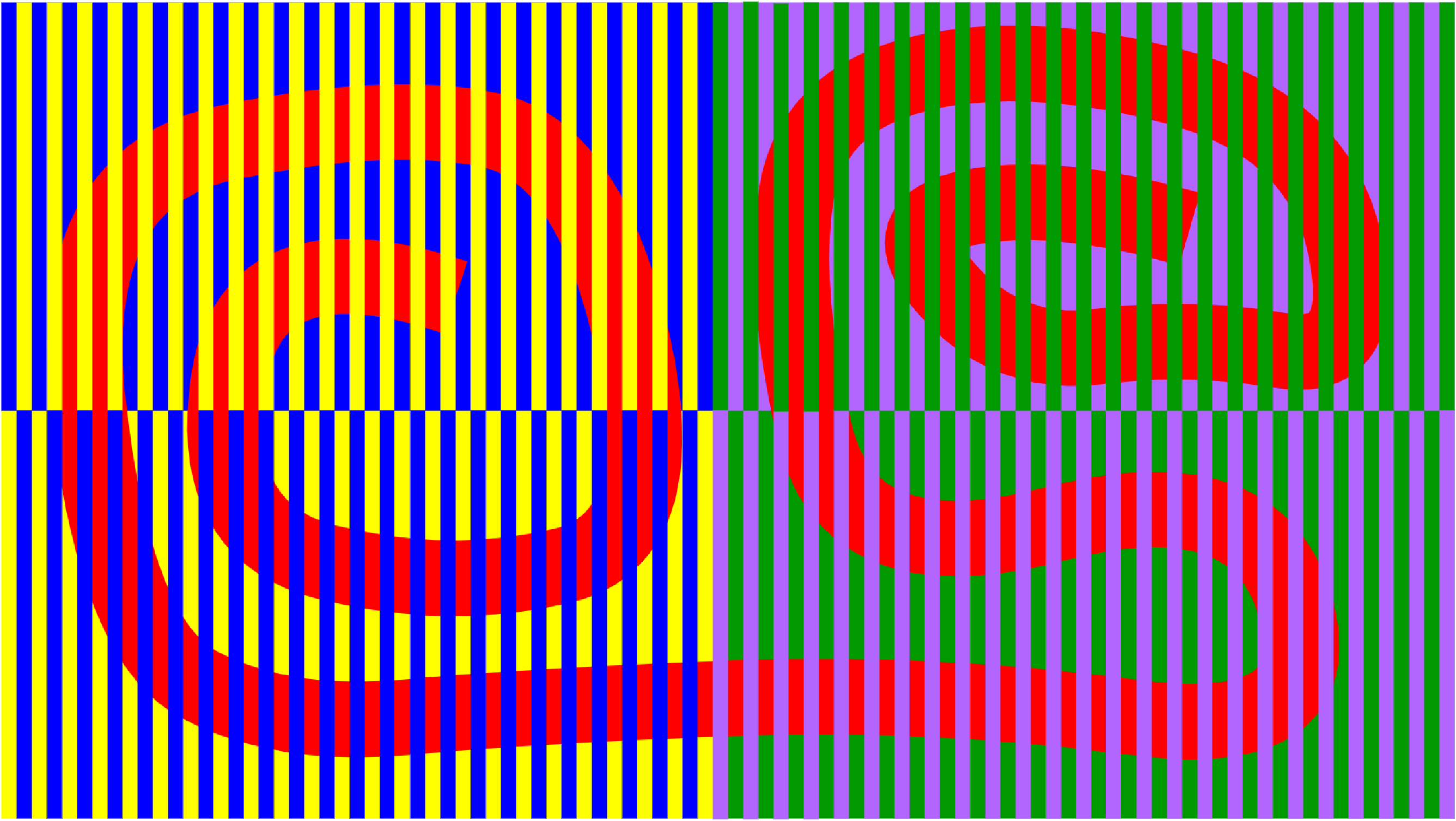}
    \end{tabular}
    \caption{Example assimilation illusions with different shapes and colors used in algorithm design~\cite{Ulucan/Ulucan/Ebner:2022a,BachWP,KitaokaWP}.}
    \label{fig:illusions}
\end{figure}

During the formation of our illusion set, we selected color assimilation illusions with various shapes and colors by considering the inducer's frequency of occurrence and its thickness. We do not include images that have no illusion effect, since it is not meaningful to reproduce images that do not cause an illusion sensation. Thus, we create a set containing a range of images that start to evoke an illusion sensation to images with a strong illusion effect (Fig.~\ref{fig:illusions}). 

\subsection{Discussion on Color Illusions}
After we form our set, we extract the block size $\beta$ and the controlling parameter of the Gaussian kernel $\sigma$ by analyzing the reproduction of distinct color assimilation illusions having various shapes, inducer frequencies, and colors via computational color constancy algorithms. While analyzing the reproduction, we prefer to carry out a visual investigation by taking the intensities of the pixels in the target regions into account since there is no error metric designed for this task. The lack of an objective quantitative evaluation method is not surprising, since color illusions are a sensation and sensations cannot be quantitatively measured. In other words, distinct observers do not perceive illusions entirely the same even if we only consider observers with a normal vision since the sensory processing of individuals differs from each other~\cite{Emery/Webster:2019}, i.e., even if distinct observers perceive a target as blue, the perceived shade of blue may differ between individuals, which is why studies mimicking illusions usually provide the intensity change within the target image~\cite{Marini/Rizzi:2000,Funt/Ciurea/McCann:2004,Corney/Lotto:2007}. It is worth mentioning that while we cannot measure sensations quantitatively, the closest alternative to provide quantitative measurements might be to conduct experiments similar to color-matching experiments with observers. In these experiments, the observers might be asked to match the target's colors they perceive with a test patch whose RGB triplet can be controlled by the observers. Thus, ground truth images can be obtained for illusions, which can be used for the evaluations. Nevertheless, performing such experiments and forming a dataset containing ground truths is beyond the scope of this work.

\begin{figure}
    \centering
    \setlength{\tabcolsep}{1.5pt} 
    \renewcommand{\arraystretch}{0.5} 
    \begin{tabular}{c c c c}
    \multicolumn{1}{c}{\footnotesize \bf{$\beta = 2$}} 
    & \multicolumn{1}{c}{\footnotesize \bf{$\beta = 8$}} 
    & \multicolumn{1}{c}{\footnotesize \bf{$\beta = 16$}}
    & \multicolumn{1}{c}{\footnotesize \bf{$\beta = 32$}}
    \\ 
    \includegraphics[width=.234\linewidth, height = 1.3cm]{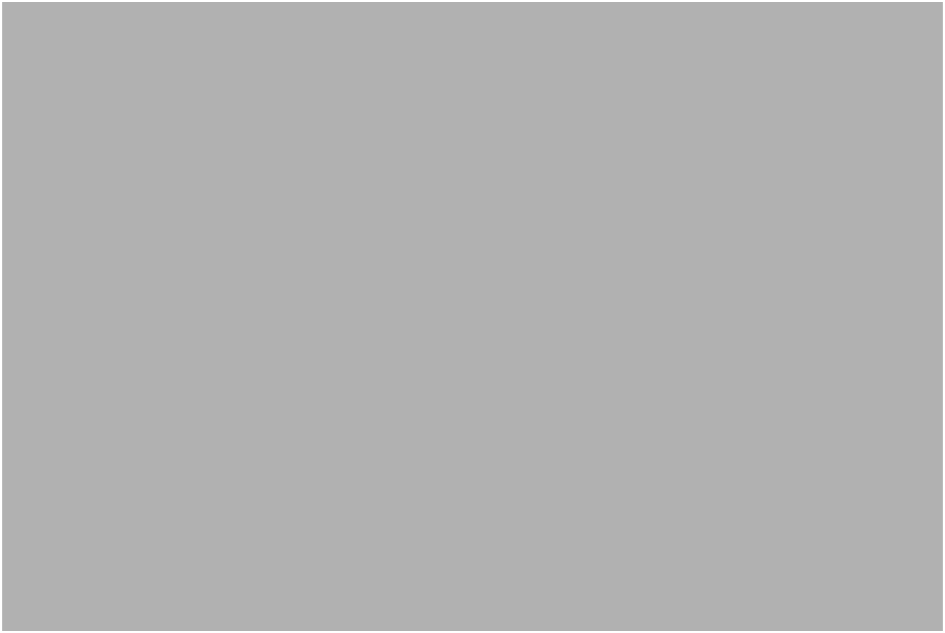} 
    & \includegraphics[width=.234\linewidth, height = 1.3cm]{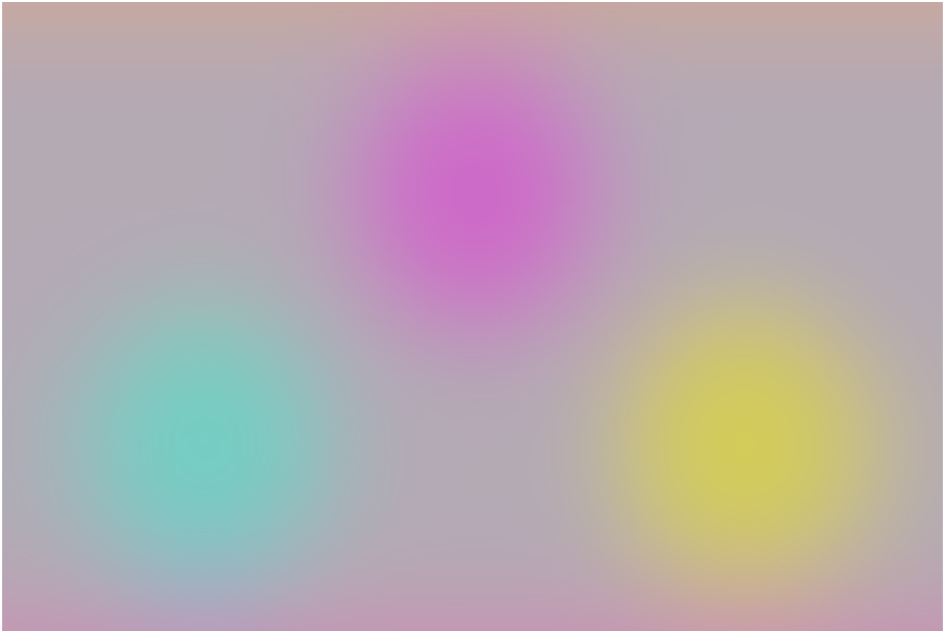} 
    & \includegraphics[width=.234\linewidth, height = 1.3cm]{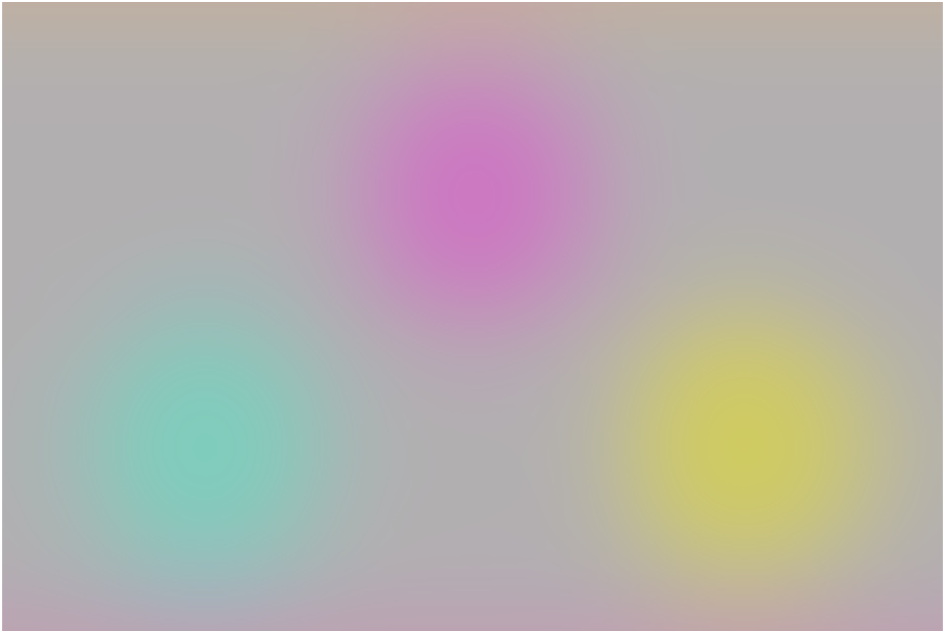}
    & \includegraphics[width=.234\linewidth, height = 1.3cm]{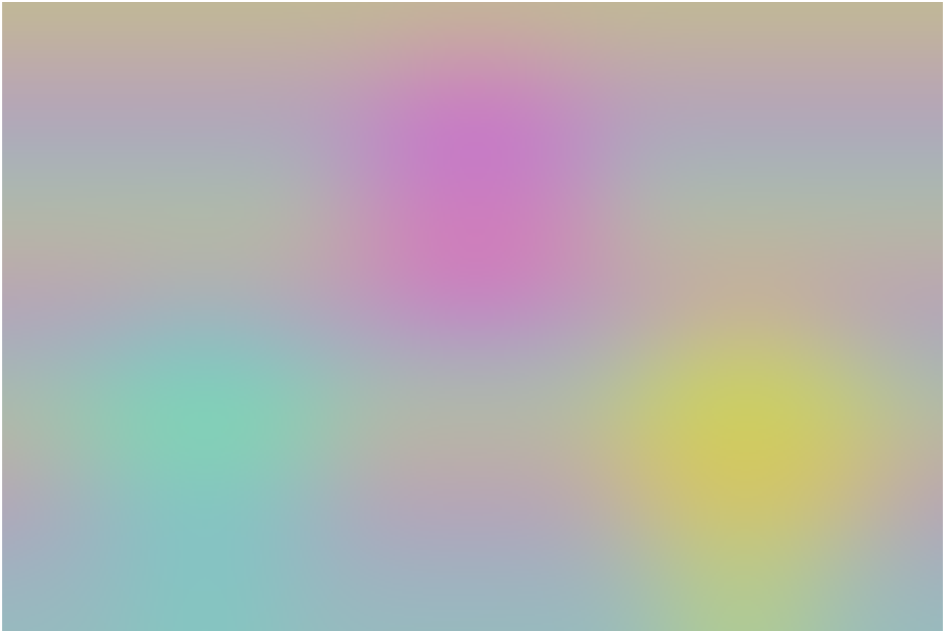} 
    \\
    \includegraphics[width=.234\linewidth, height = 1.3cm]{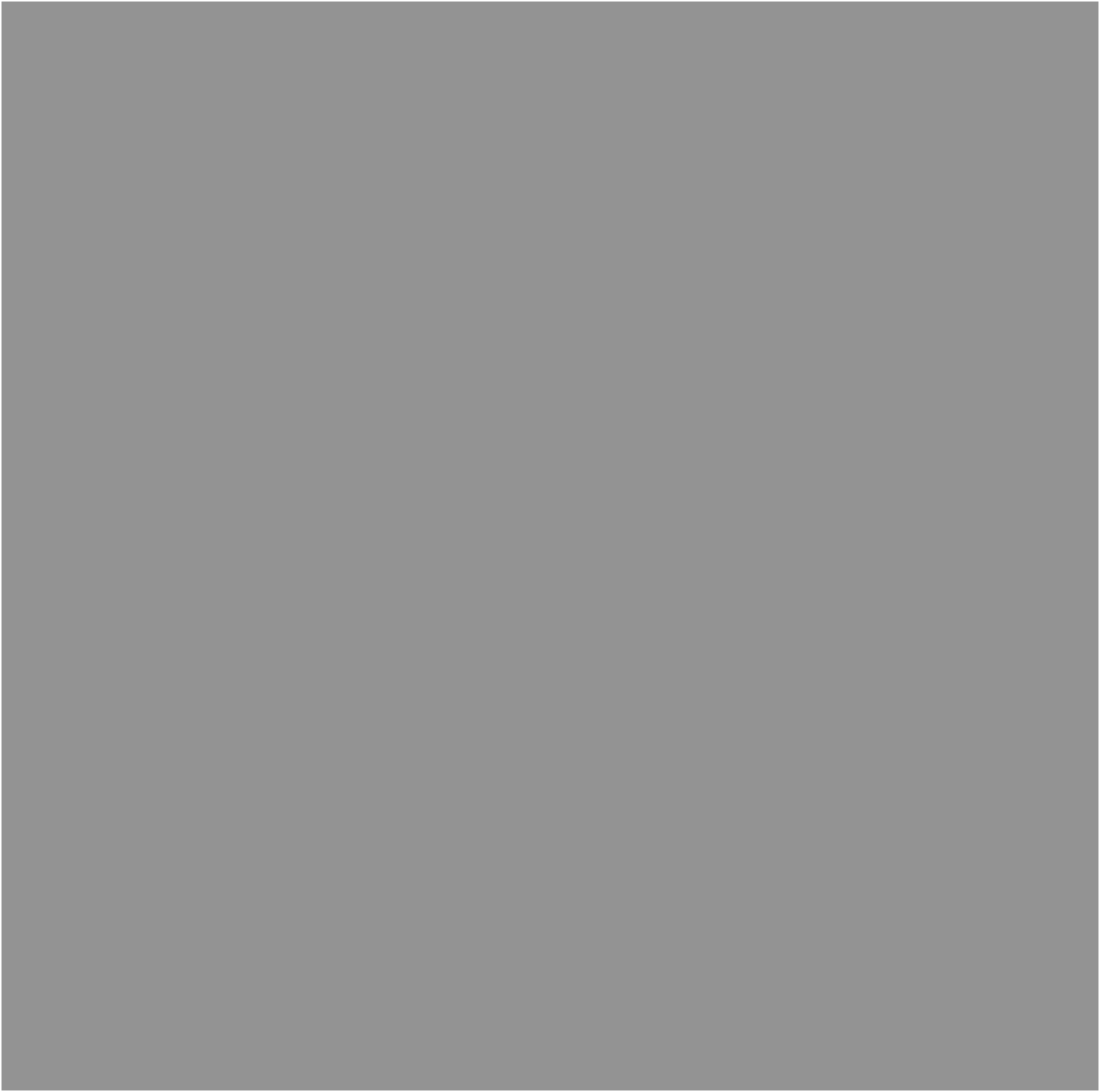} 
    & \includegraphics[width=.234\linewidth, height = 1.3cm]{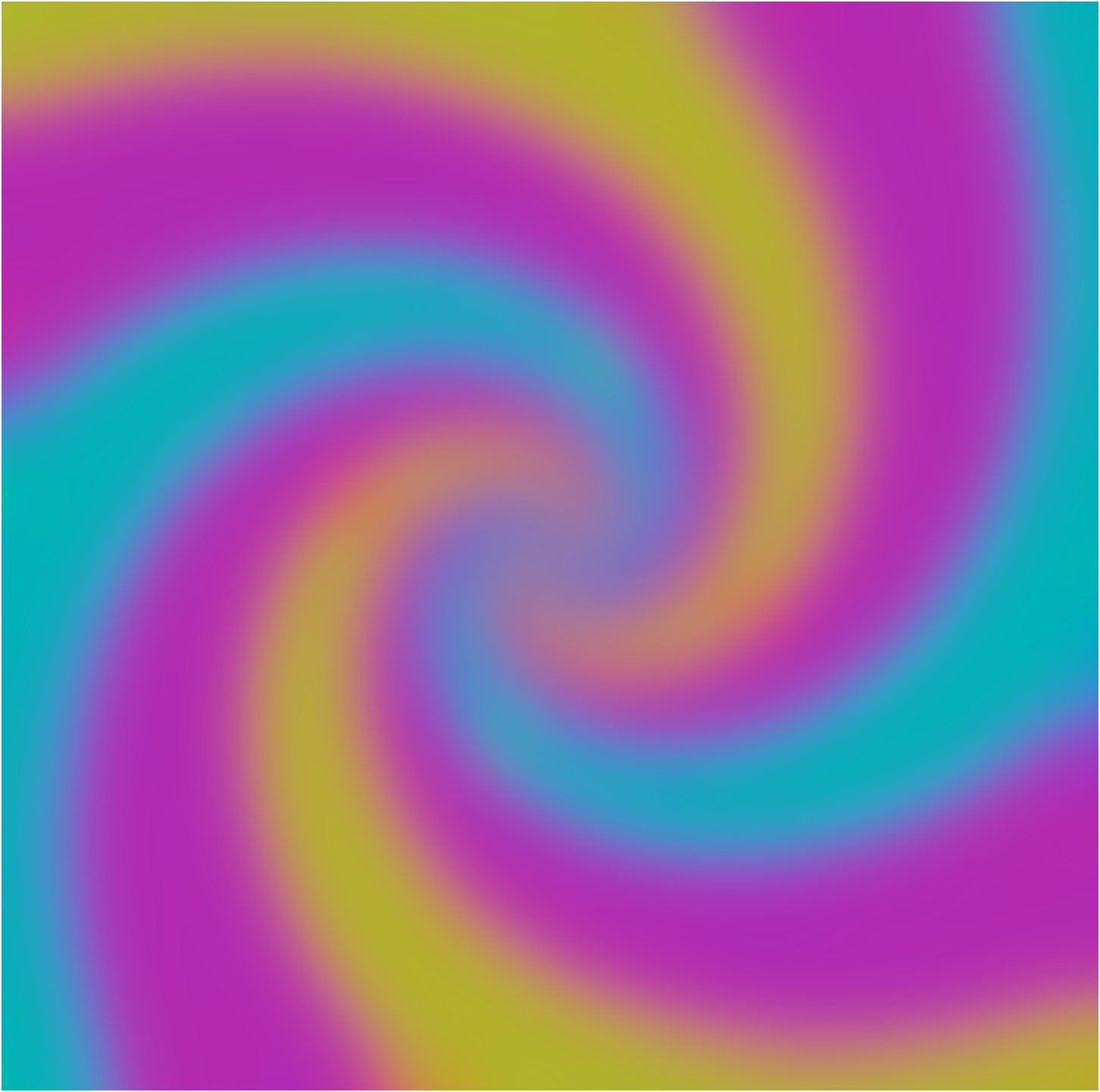} 
    & \includegraphics[width=.234\linewidth, height = 1.3cm]{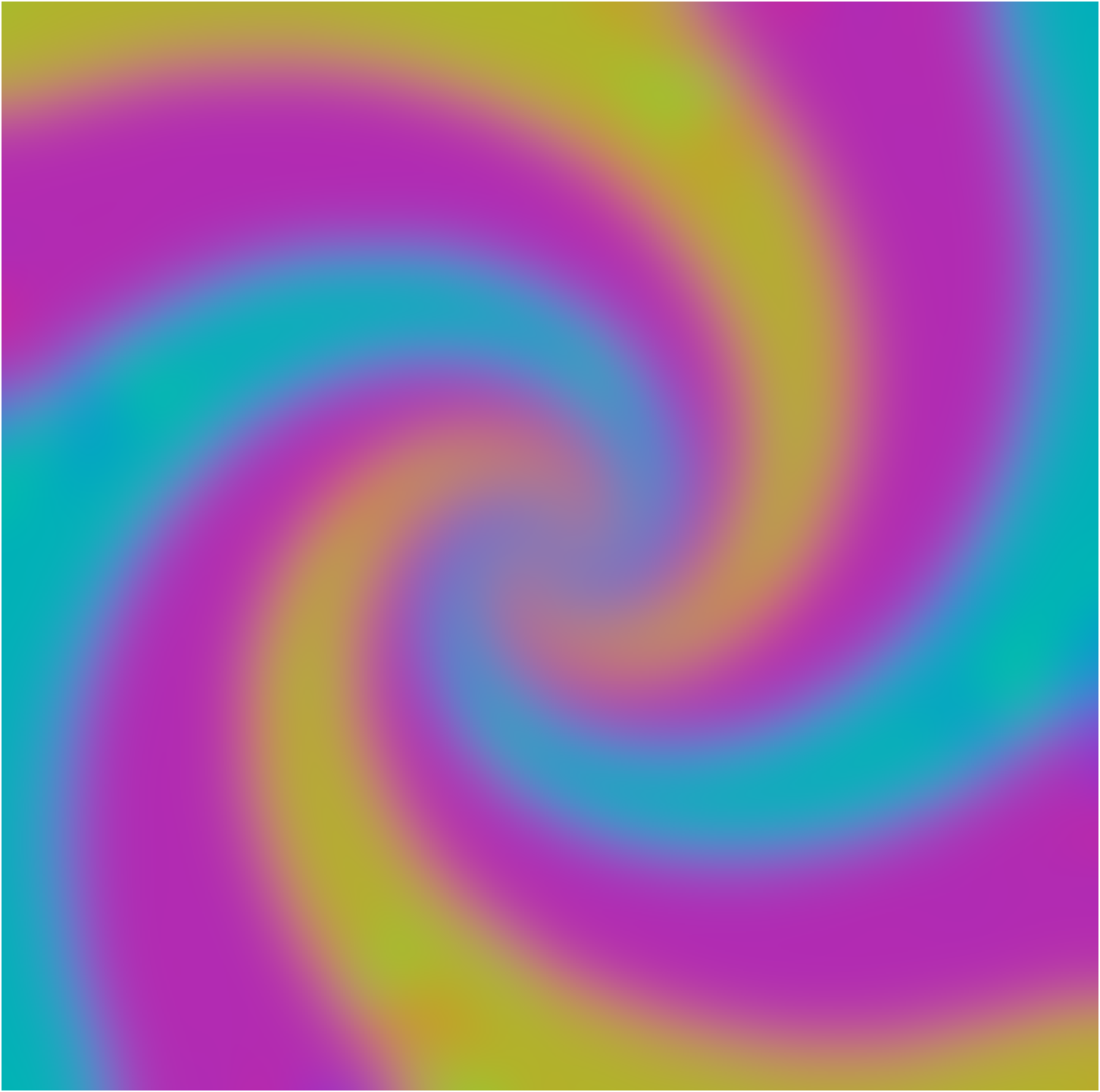}
    & \includegraphics[width=.234\linewidth, height = 1.3cm]{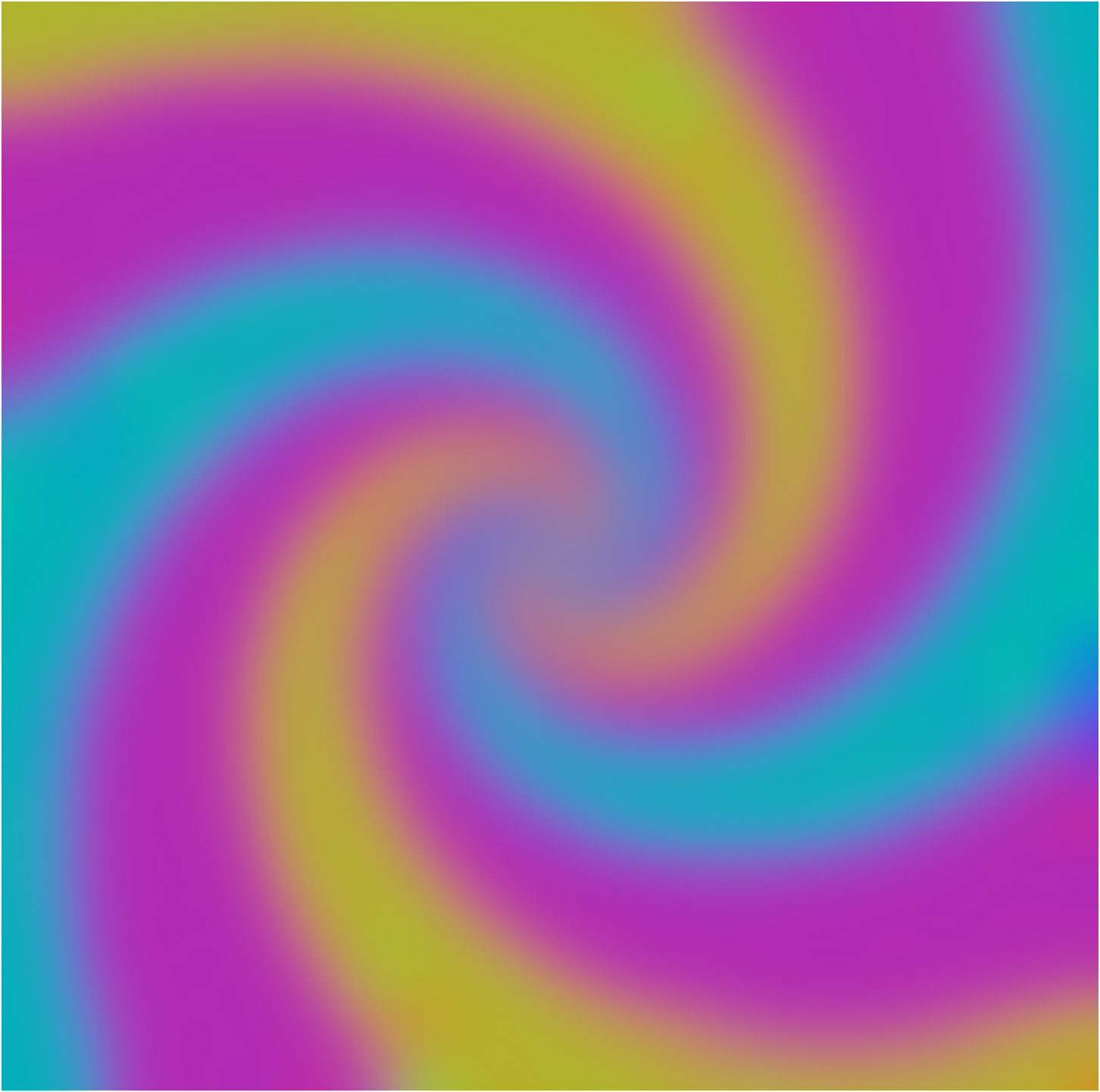} 
    \\
    \includegraphics[width=.234\linewidth, height = 1.3cm]{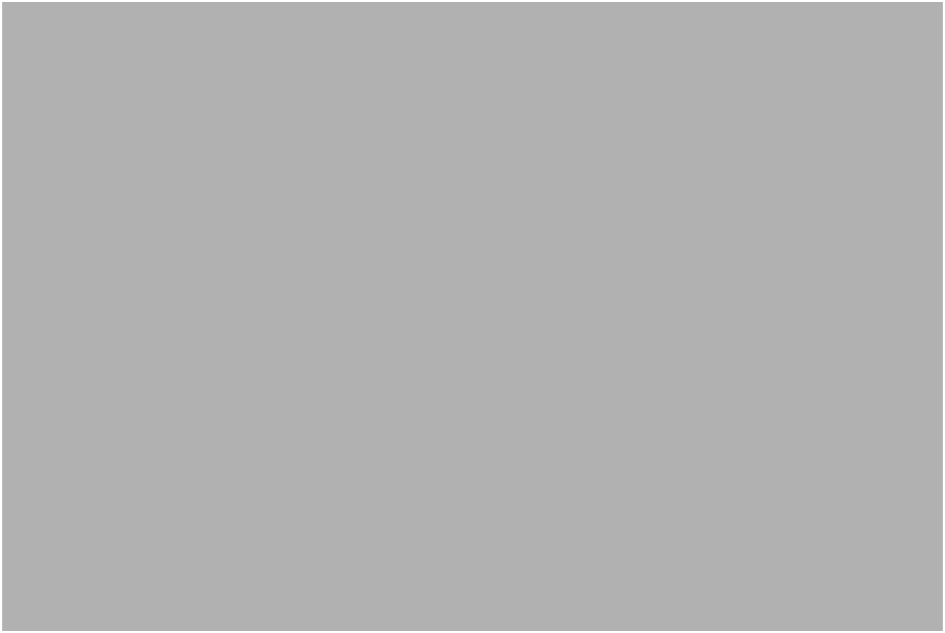} 
    & \includegraphics[width=.234\linewidth, height = 1.3cm]{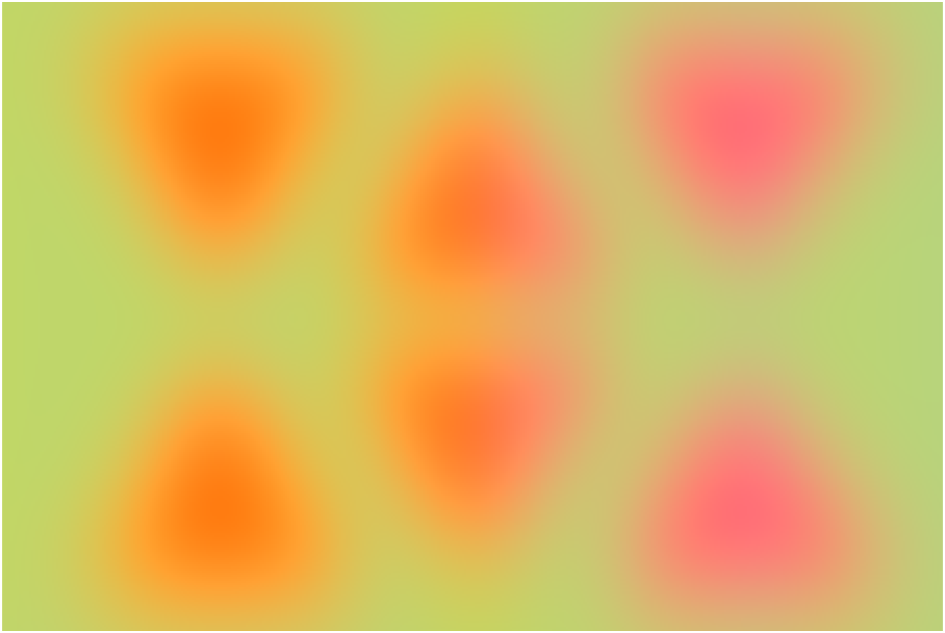} 
    & \includegraphics[width=.234\linewidth, height = 1.3cm]{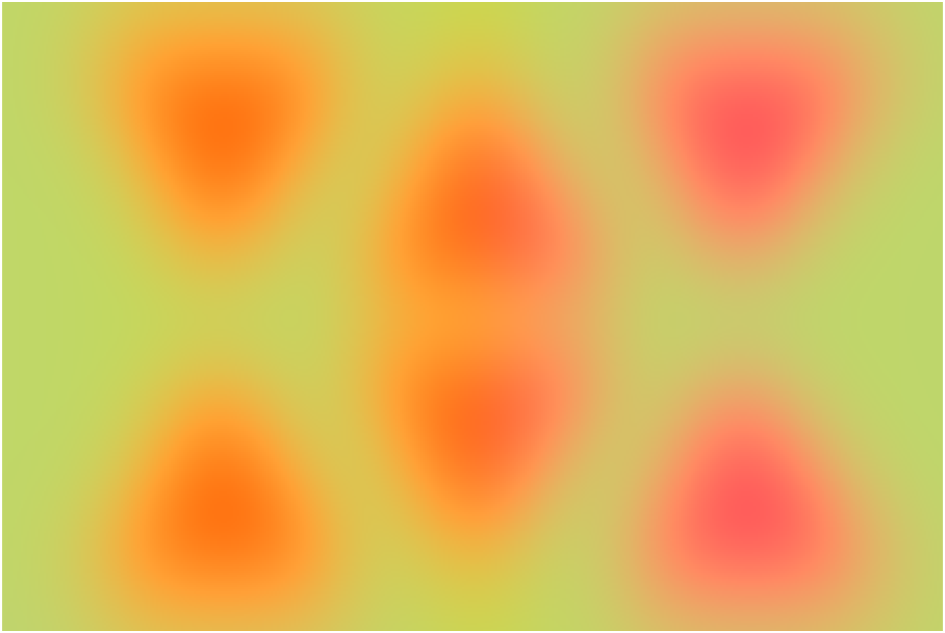}
    & \includegraphics[width=.234\linewidth, height = 1.3cm]{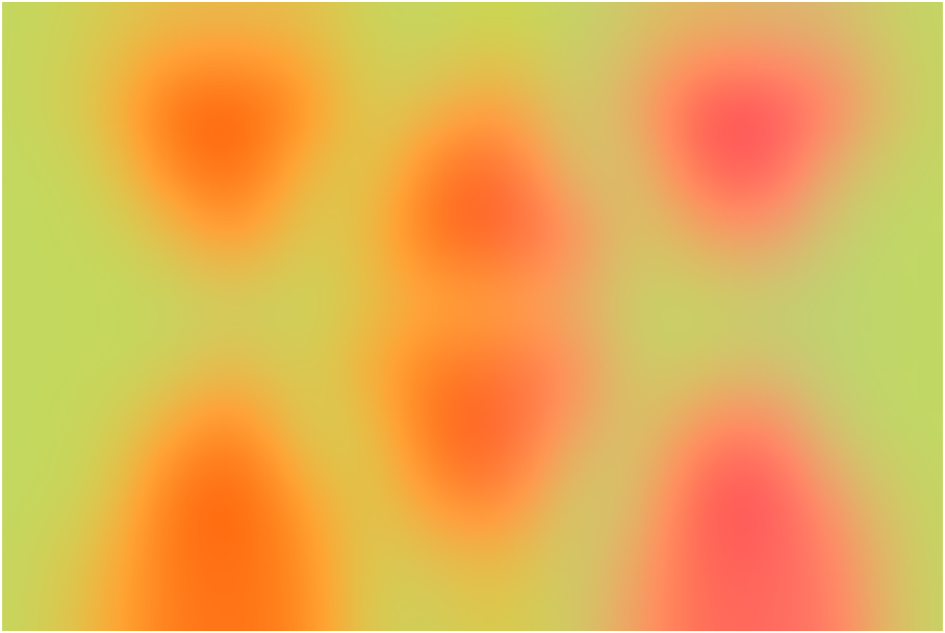} 
    \\ 
    \multicolumn{1}{c}{\footnotesize \bf{$\sigma = 4$}} 
    & \multicolumn{1}{c}{\footnotesize \bf{$\sigma = 16$}} 
    & \multicolumn{1}{c}{\footnotesize \bf{$\sigma = 24$}} 
    & \multicolumn{1}{c}{\footnotesize \bf{$\sigma = 48$}}
    \\ 
    \includegraphics[width=.234\linewidth, height = 1.3cm]{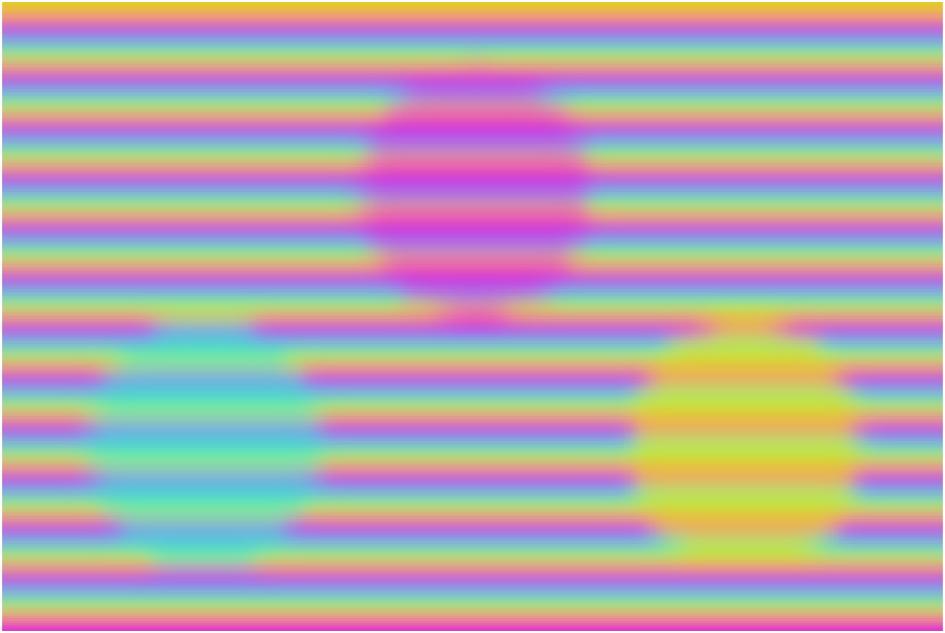} 
    & \includegraphics[width=.234\linewidth, height = 1.3cm]{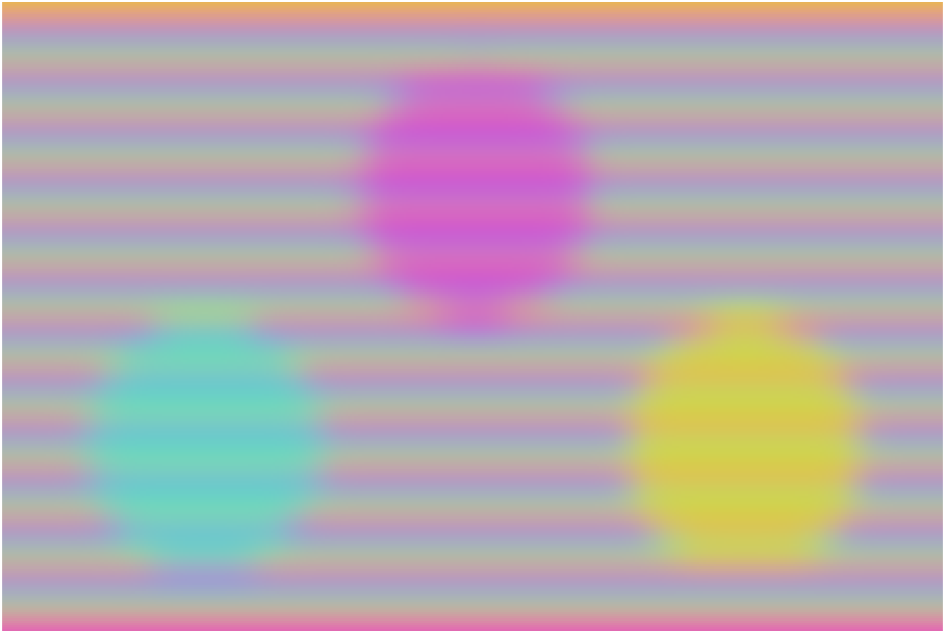}
    & \includegraphics[width=.234\linewidth, height = 1.3cm]{parameters/disk_8x8.png} 
    & \includegraphics[width=.234\linewidth, height = 1.3cm]{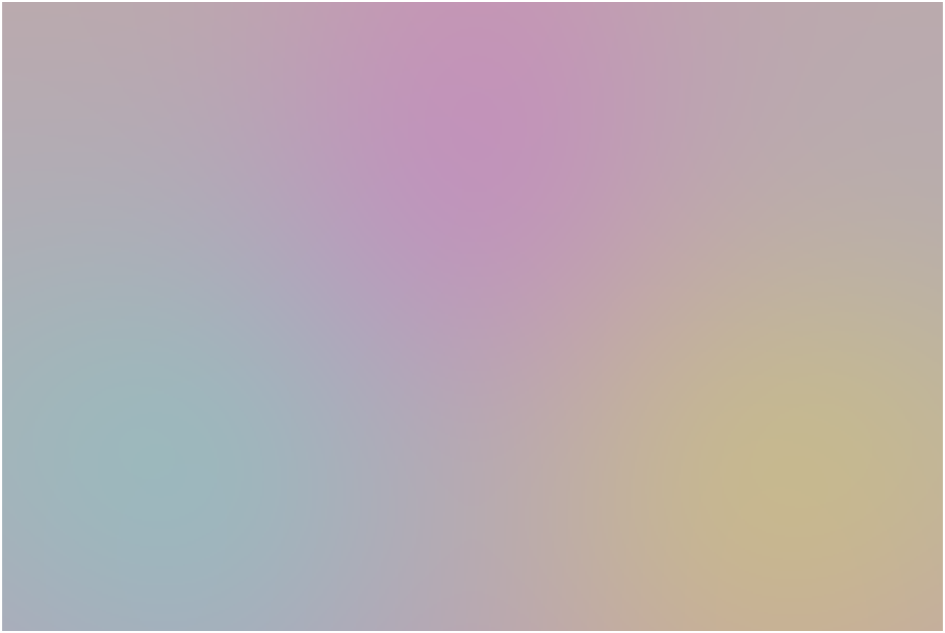}
    \\  
    \includegraphics[width=.234\linewidth, height = 1.3cm]{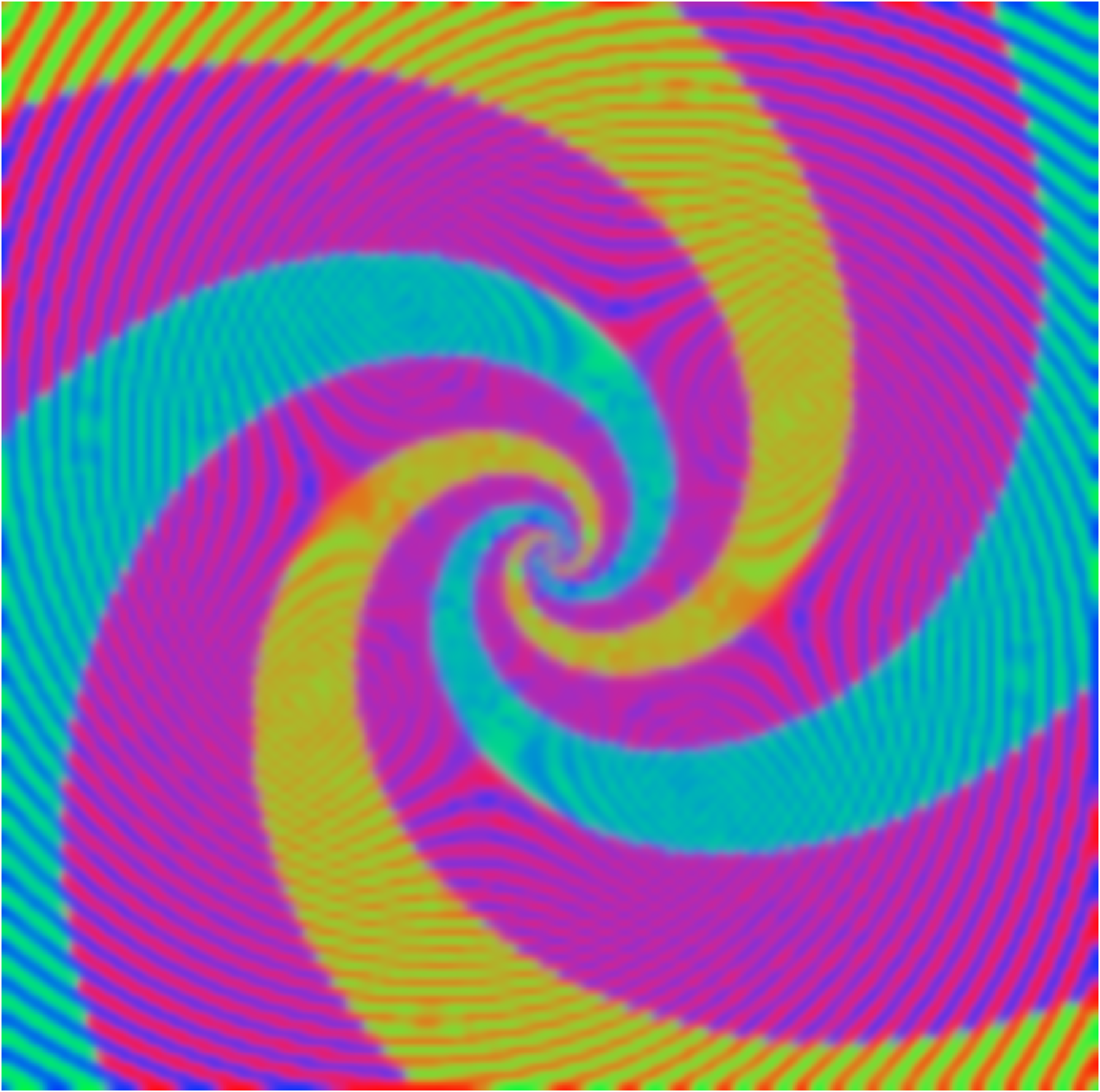} 
    & \includegraphics[width=.234\linewidth, height = 1.3cm]{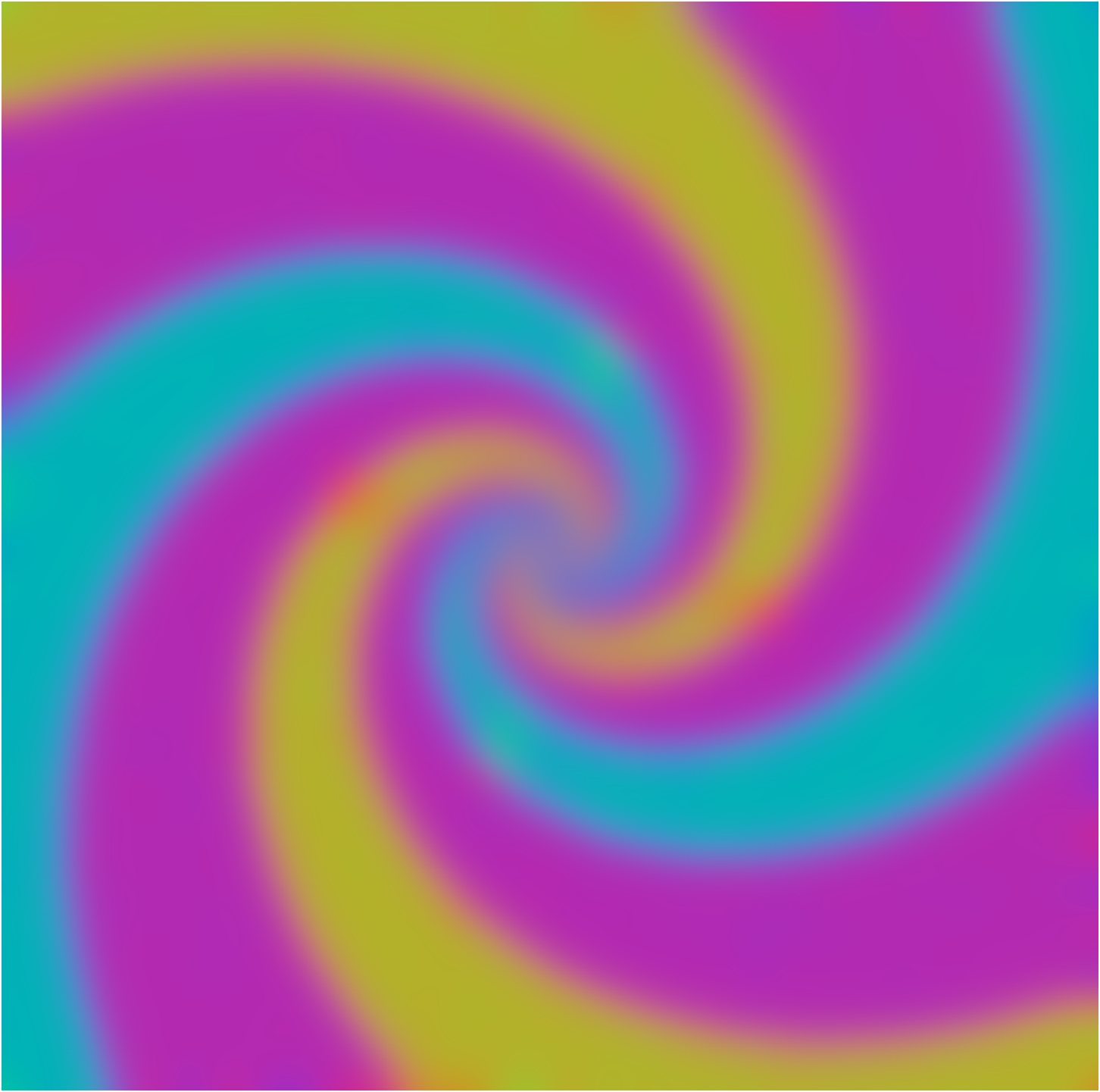}
    & \includegraphics[width=.234\linewidth, height = 1.3cm]{parameters/orochi_8x8.png} 
    & \includegraphics[width=.234\linewidth, height = 1.3cm]{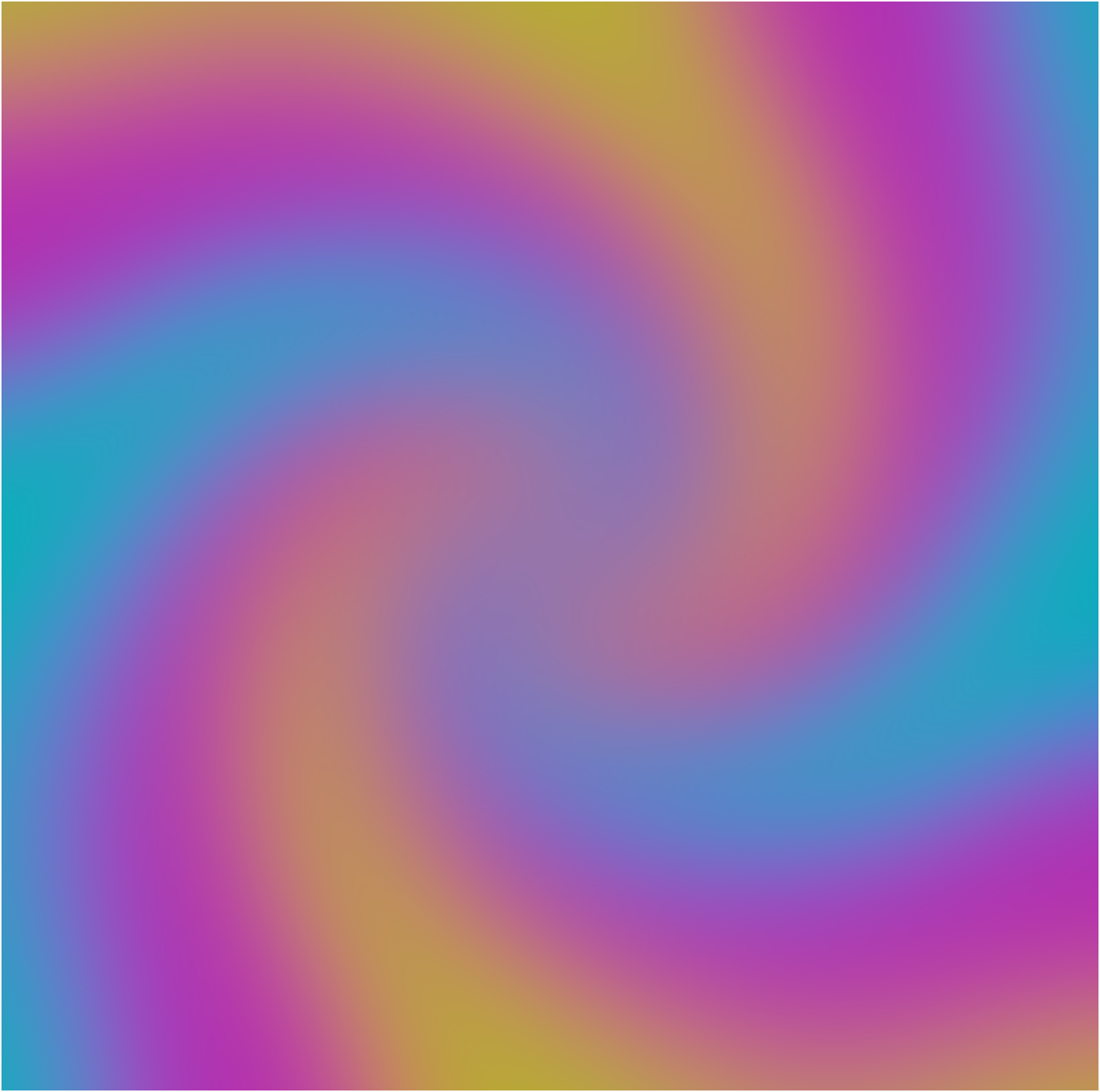}
    \\  
    \includegraphics[width=.234\linewidth, height = 1.3cm]{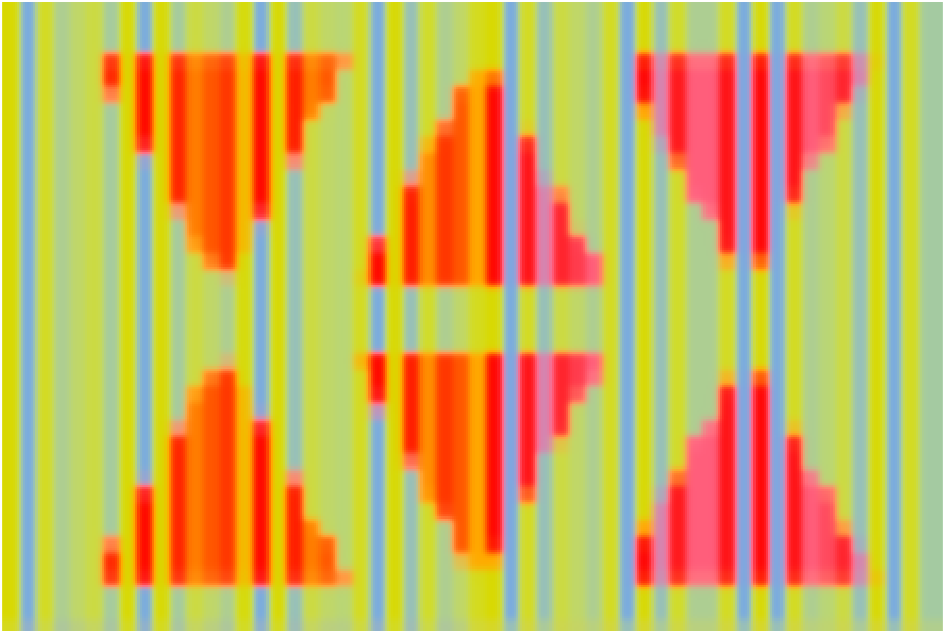} 
    & \includegraphics[width=.234\linewidth, height = 1.3cm]{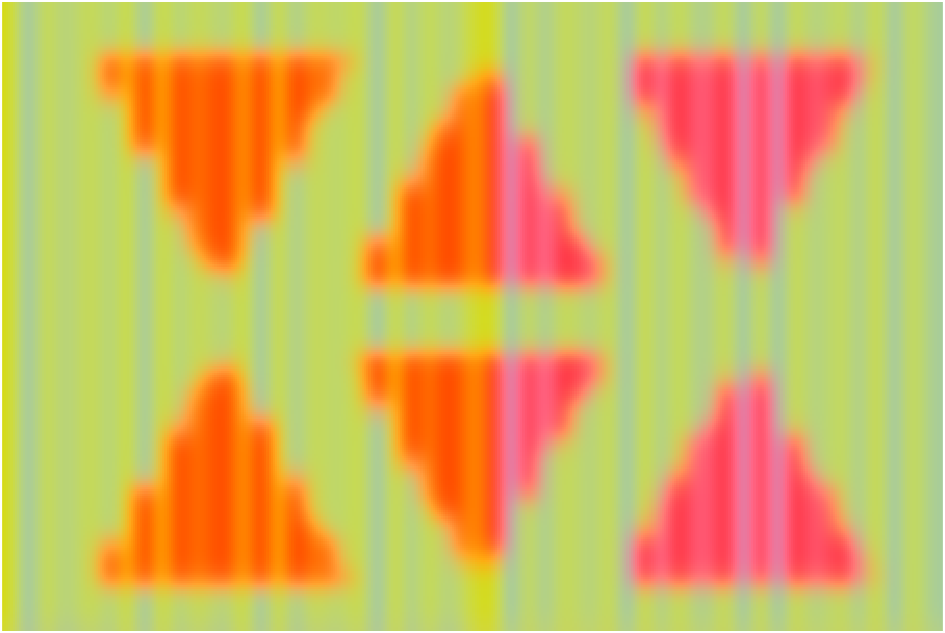} 
    & \includegraphics[width=.234\linewidth, height = 1.3cm]{parameters/tr_8x8.png}
    & \includegraphics[width=.234\linewidth, height = 1.3cm]{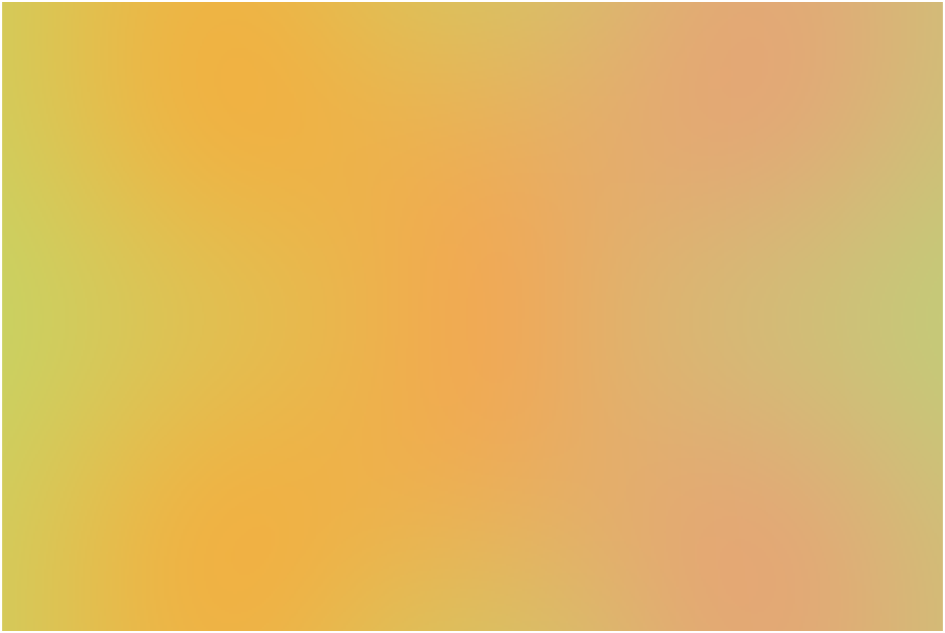} 
    \end{tabular}
    \caption{Examples of the parameter selection procedure with several combinations on different illusions. Experiments are carried out with different parameter combinations, and the parameters that work best on all illusions on average are chosen. This analysis is performed by visual analysis since there is no quantitative method for this task.}
    \label{fig:parameter_selection}
\end{figure}

\begin{figure}
    \centering
    \setlength{\tabcolsep}{1.5pt} 
    \renewcommand{\arraystretch}{0.5} 
    \begin{tabular}{c c c c c}
    \multicolumn{1}{c}{\footnotesize Input Image} 
    & \multicolumn{1}{c}{\footnotesize Input Target} 
    & \multicolumn{1}{c}{\footnotesize Estimates} 
    & \multicolumn{1}{c}{\footnotesize Output Target}
    \\ 
    \includegraphics[width=.234\linewidth, height = 1.3cm]{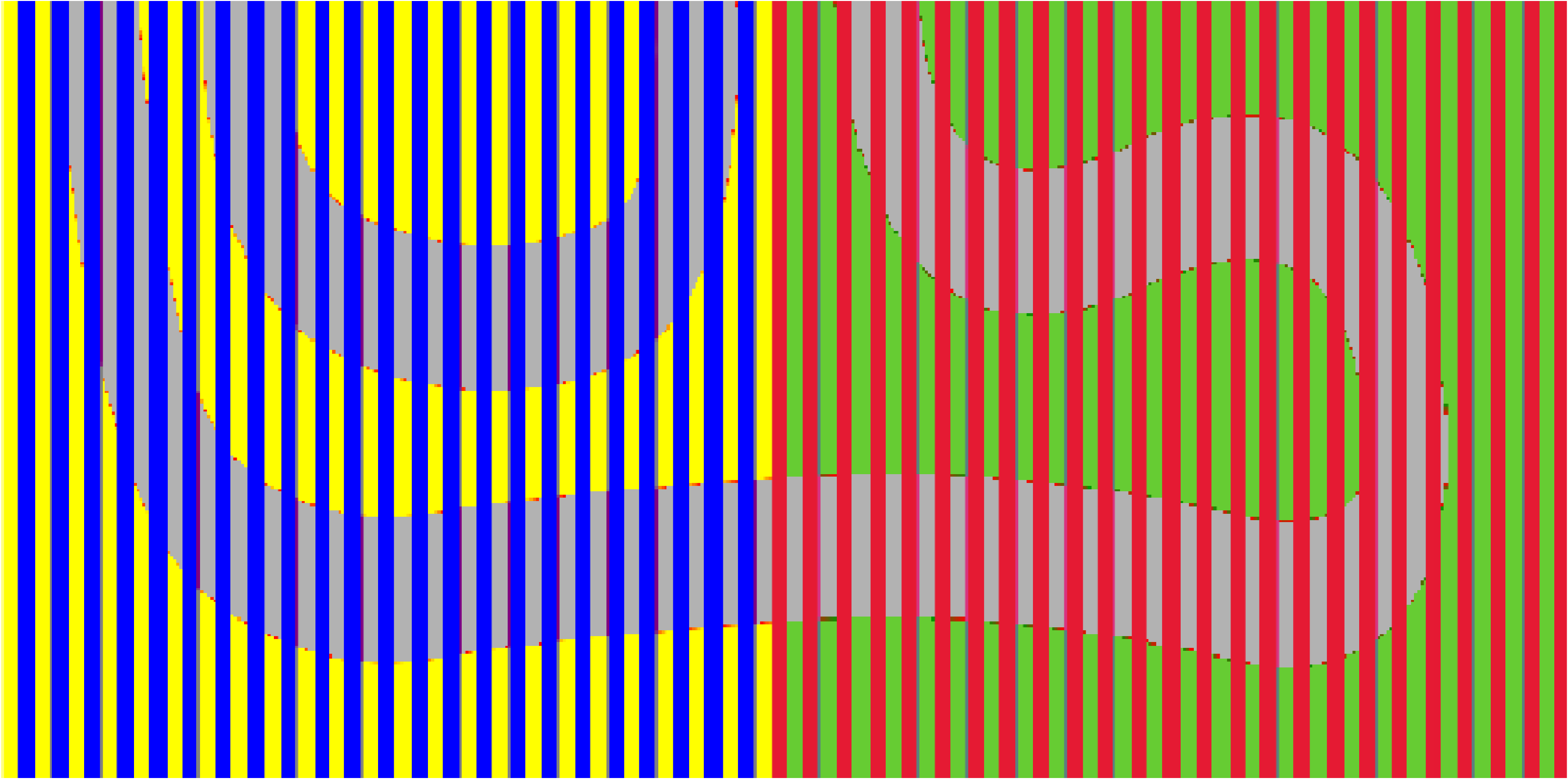} 
    &\includegraphics[width=.234\linewidth, height = 1.3cm]{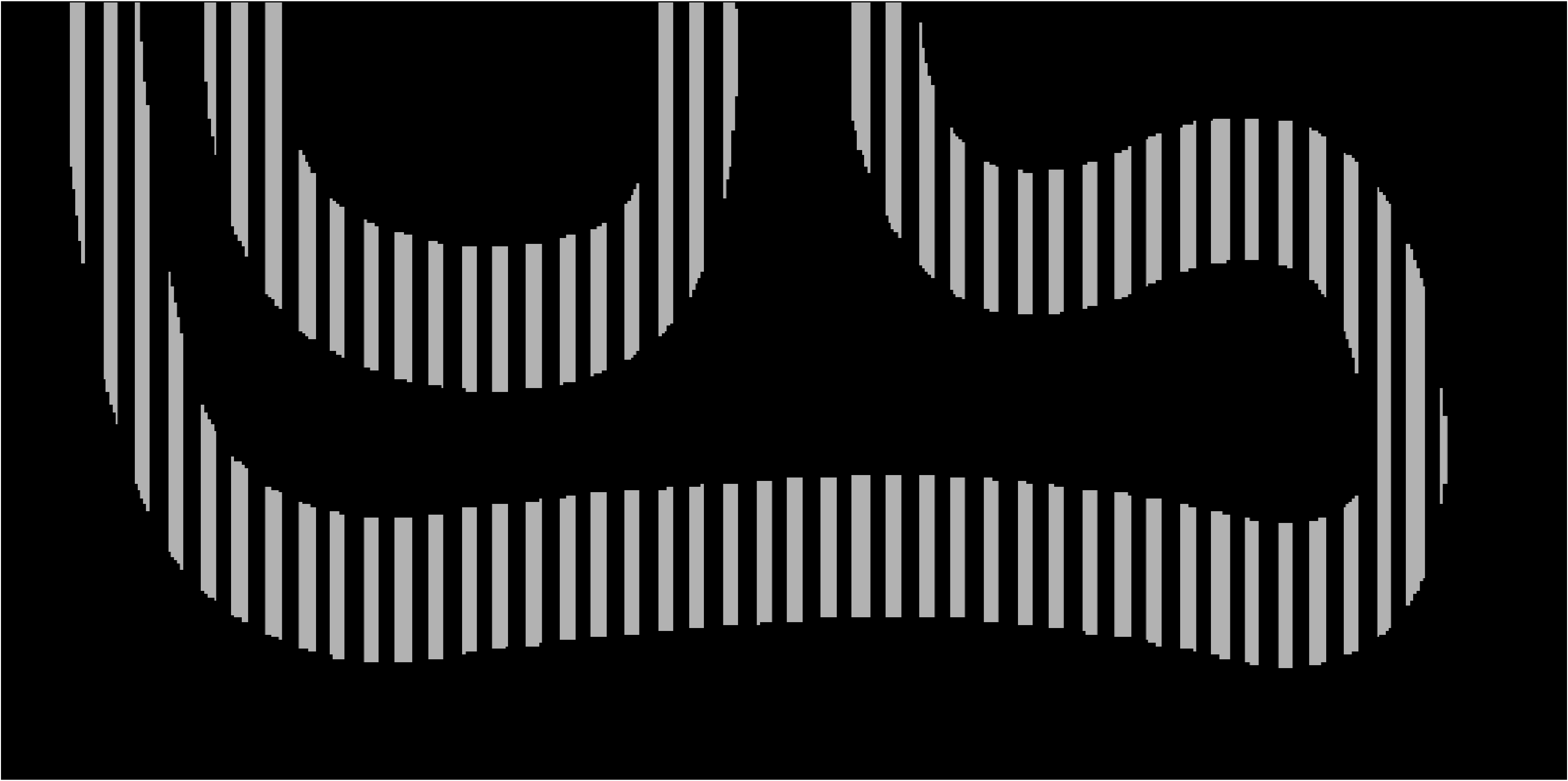}
    &\includegraphics[width=.234\linewidth, height = 1.3cm]{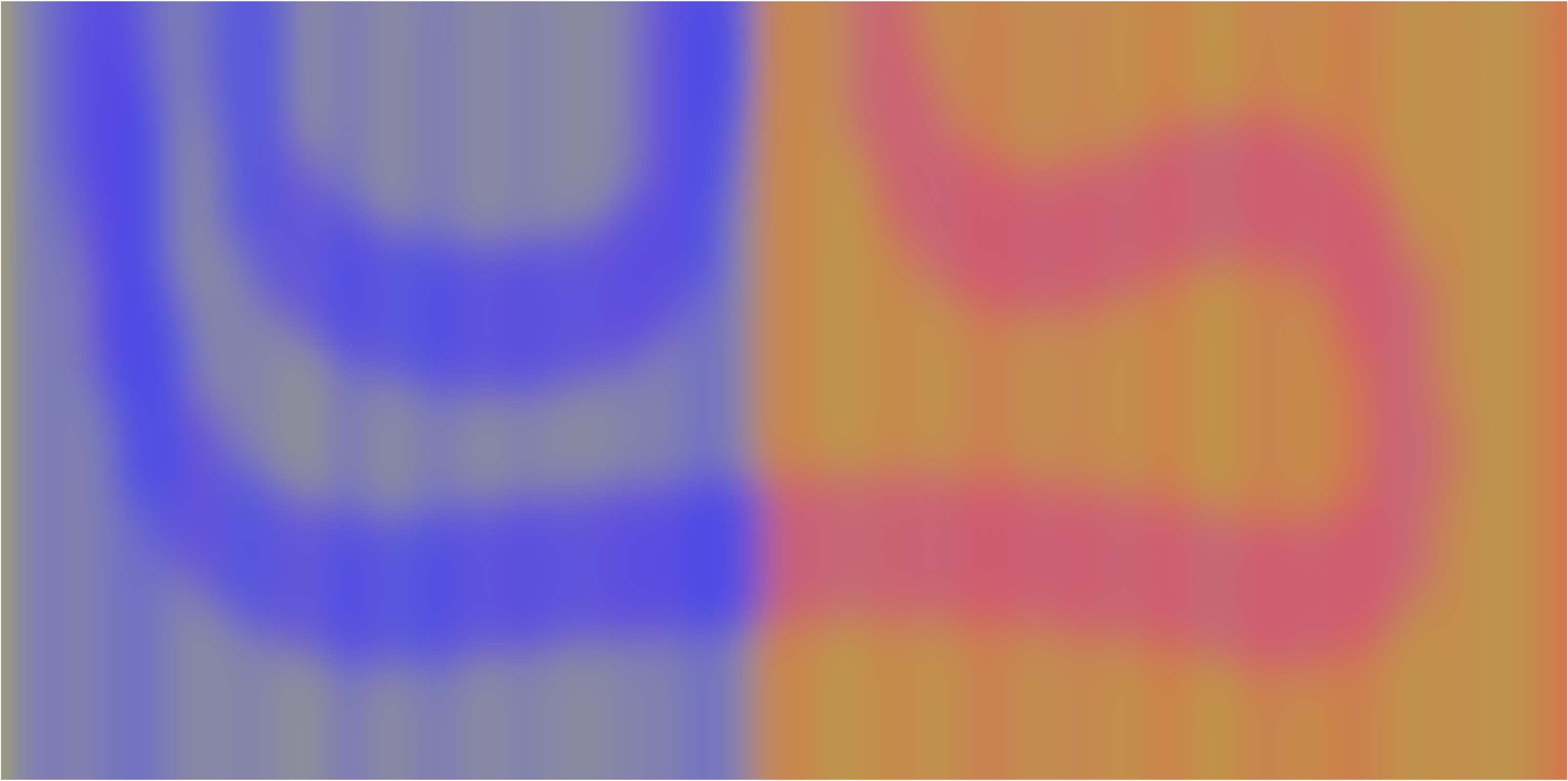}
    &\includegraphics[width=.234\linewidth, height = 1.3cm]{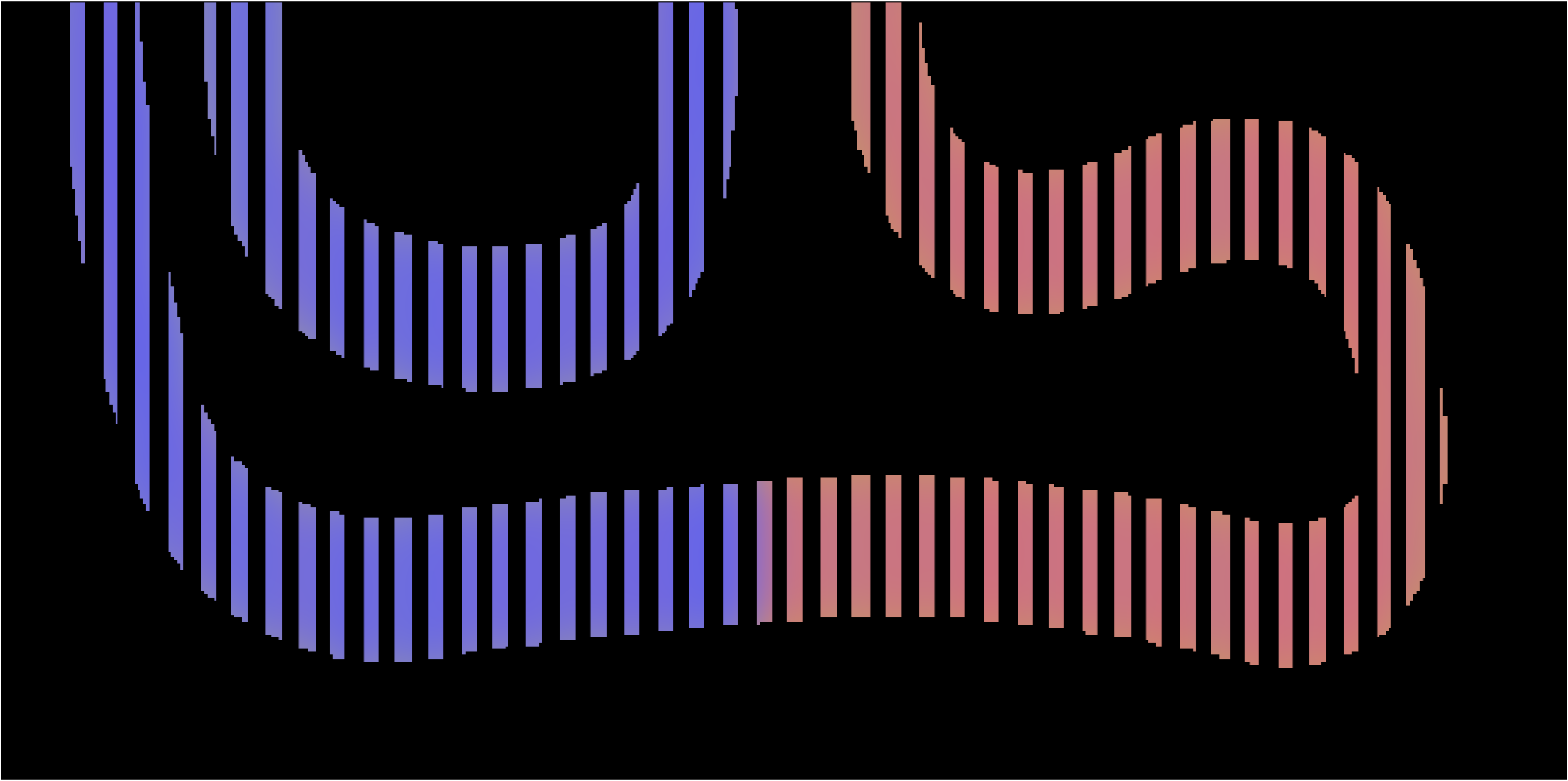}
    \\
    \includegraphics[width=.234\linewidth, height = 1.3cm]{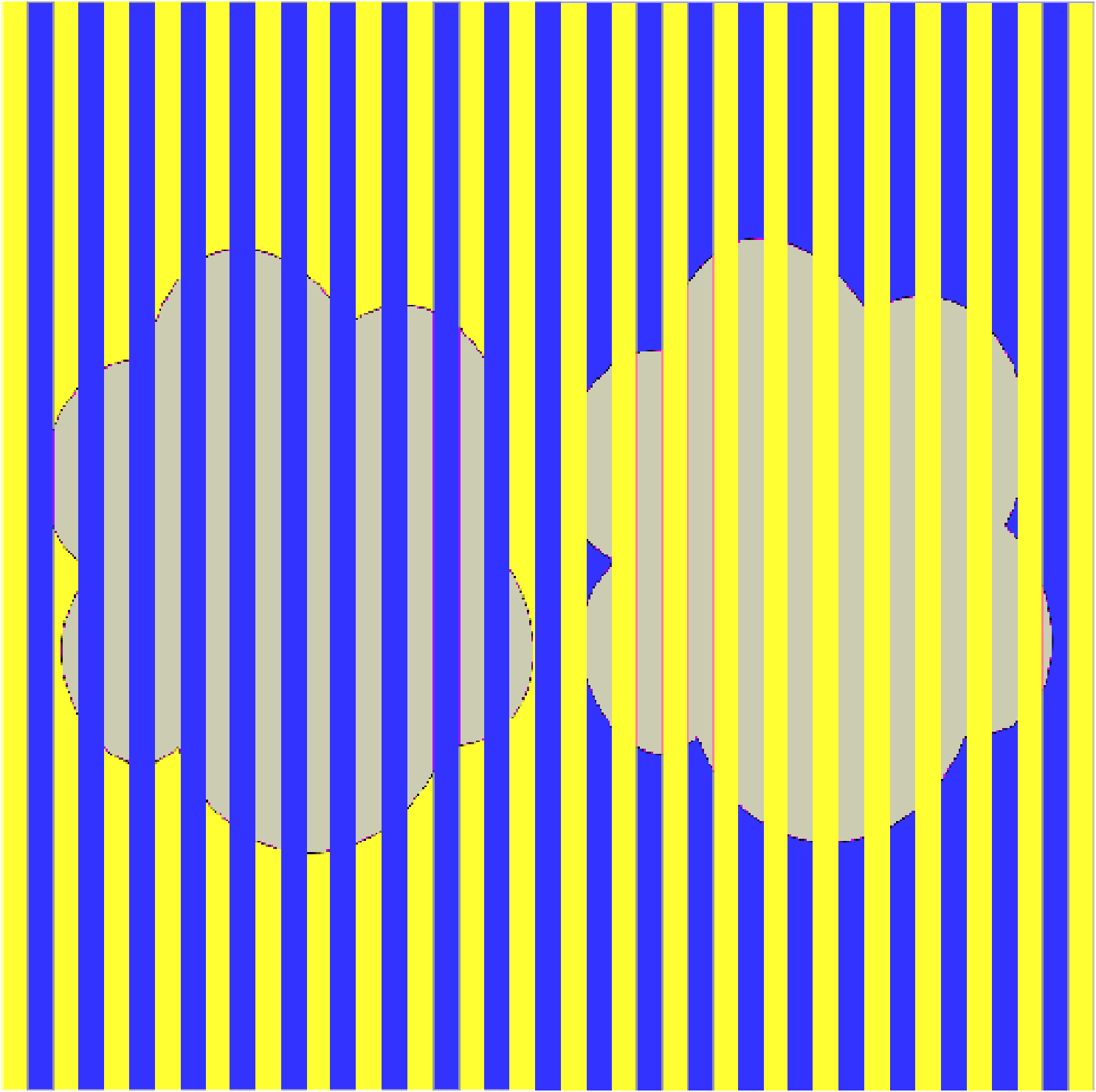} 
    &\includegraphics[width=.234\linewidth, height = 1.3cm]{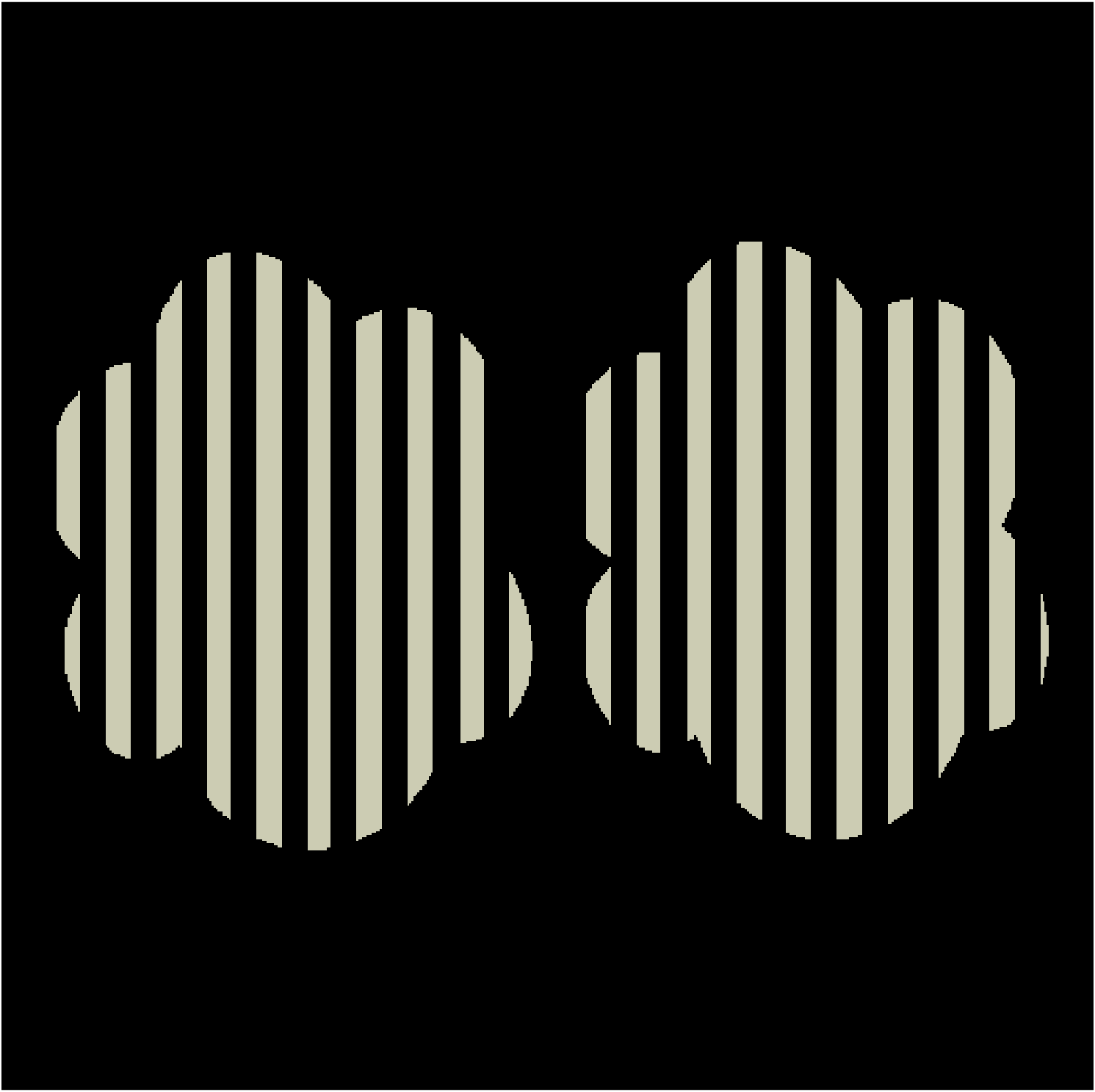}
    &\includegraphics[width=.234\linewidth, height = 1.3cm]{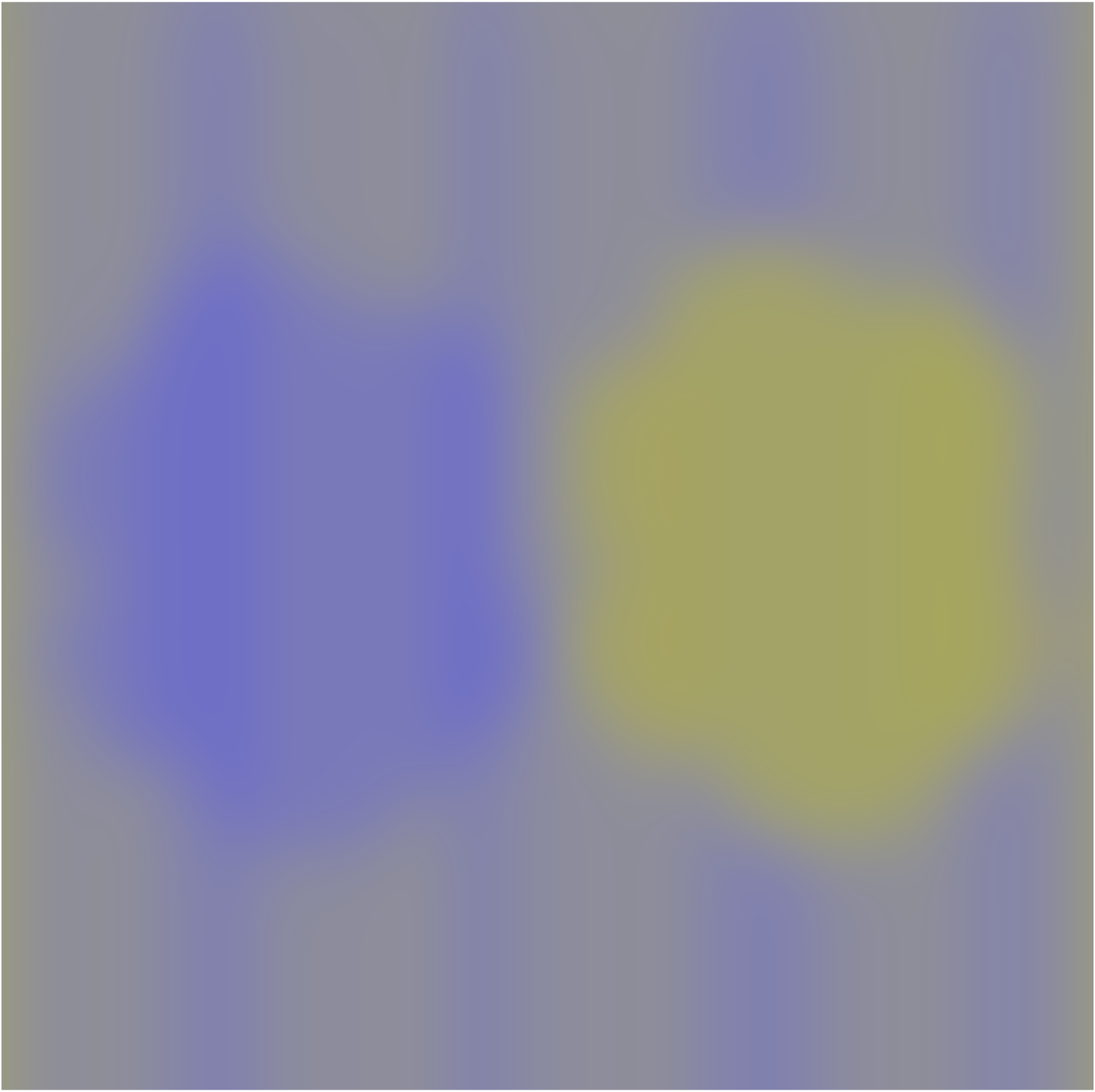}
    &\includegraphics[width=.234\linewidth, height = 1.3cm]{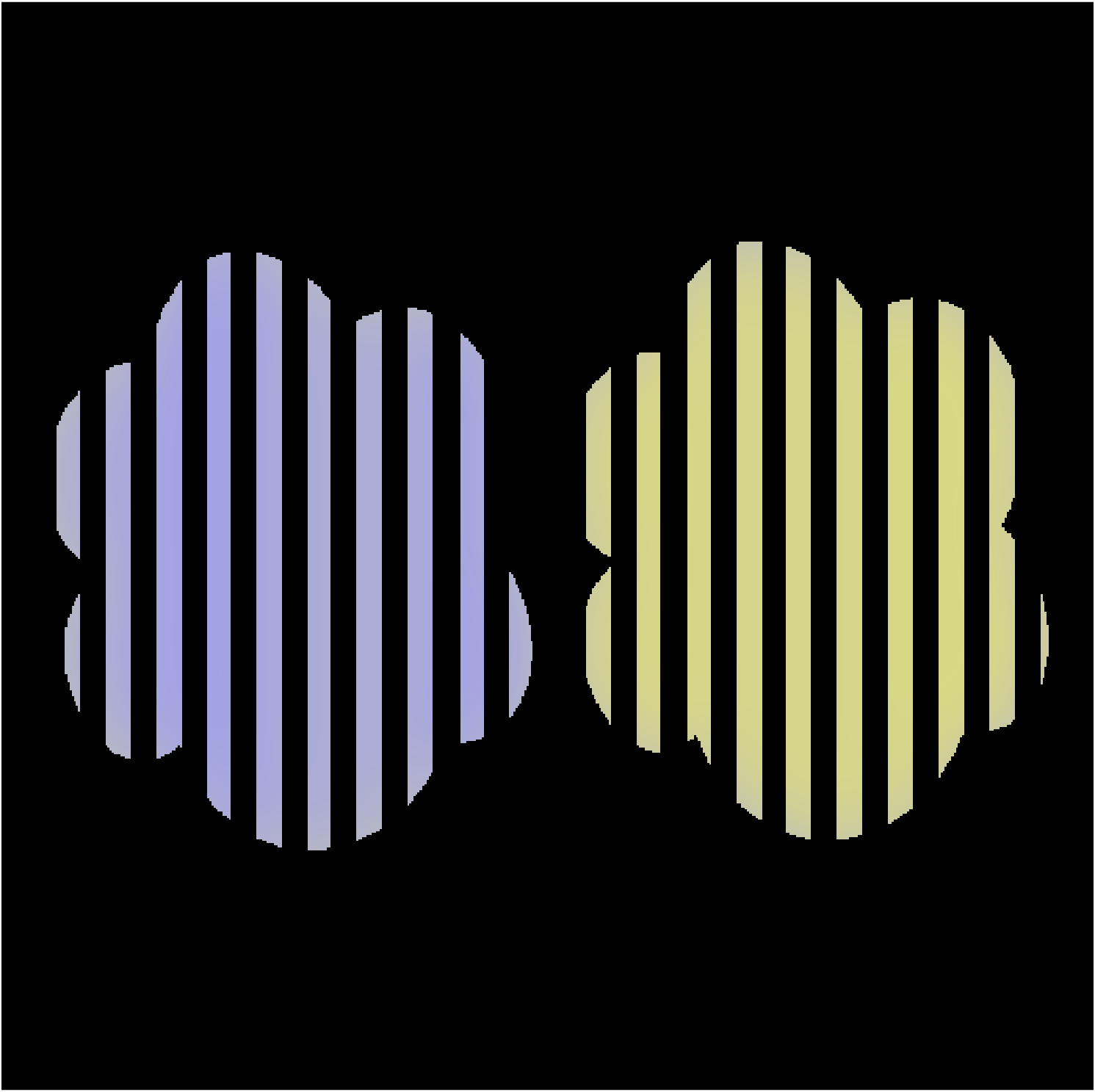}
    \\
    \includegraphics[width=.234\linewidth, height = 1.3cm]{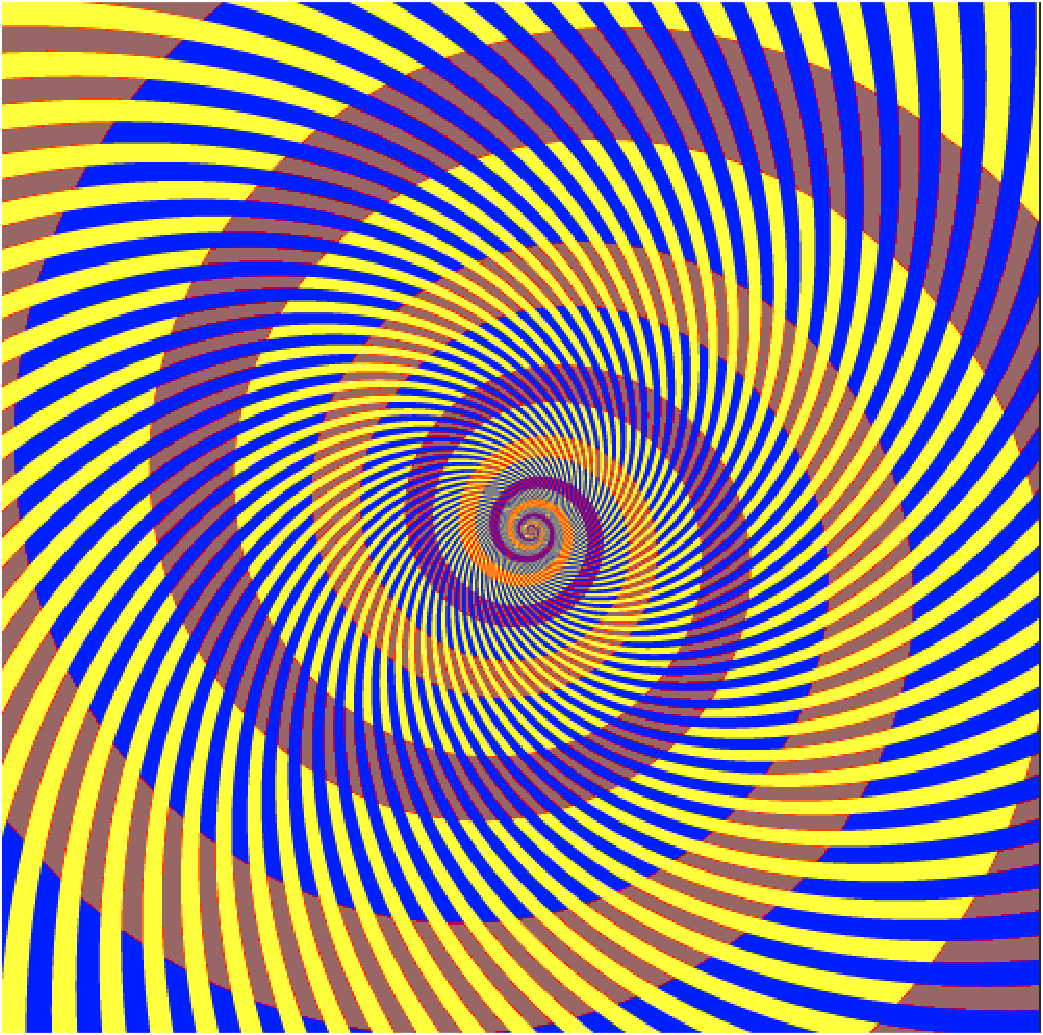} 
    &\includegraphics[width=.234\linewidth, height = 1.3cm]{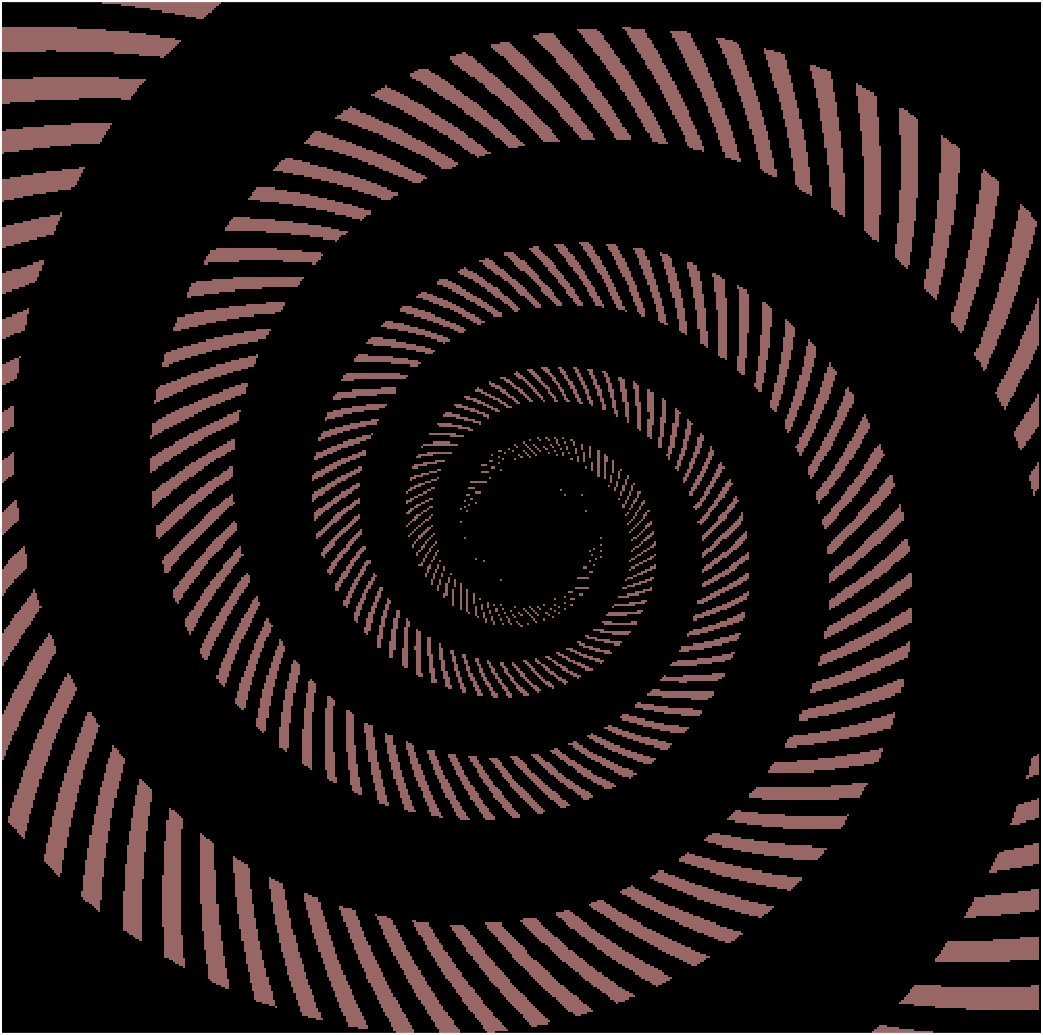}
    &\includegraphics[width=.234\linewidth, height = 1.3cm]{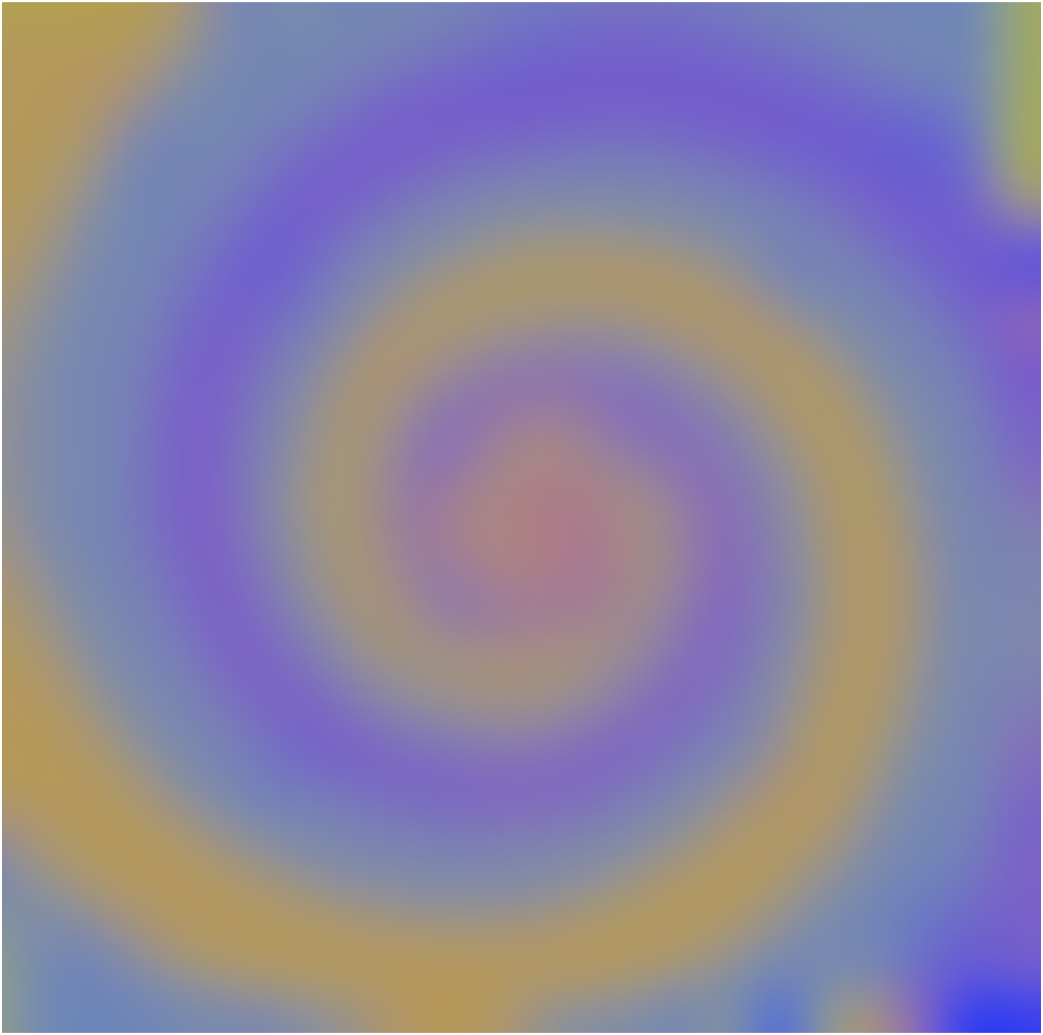} 
    &\includegraphics[width=.234\linewidth, height = 1.3cm]{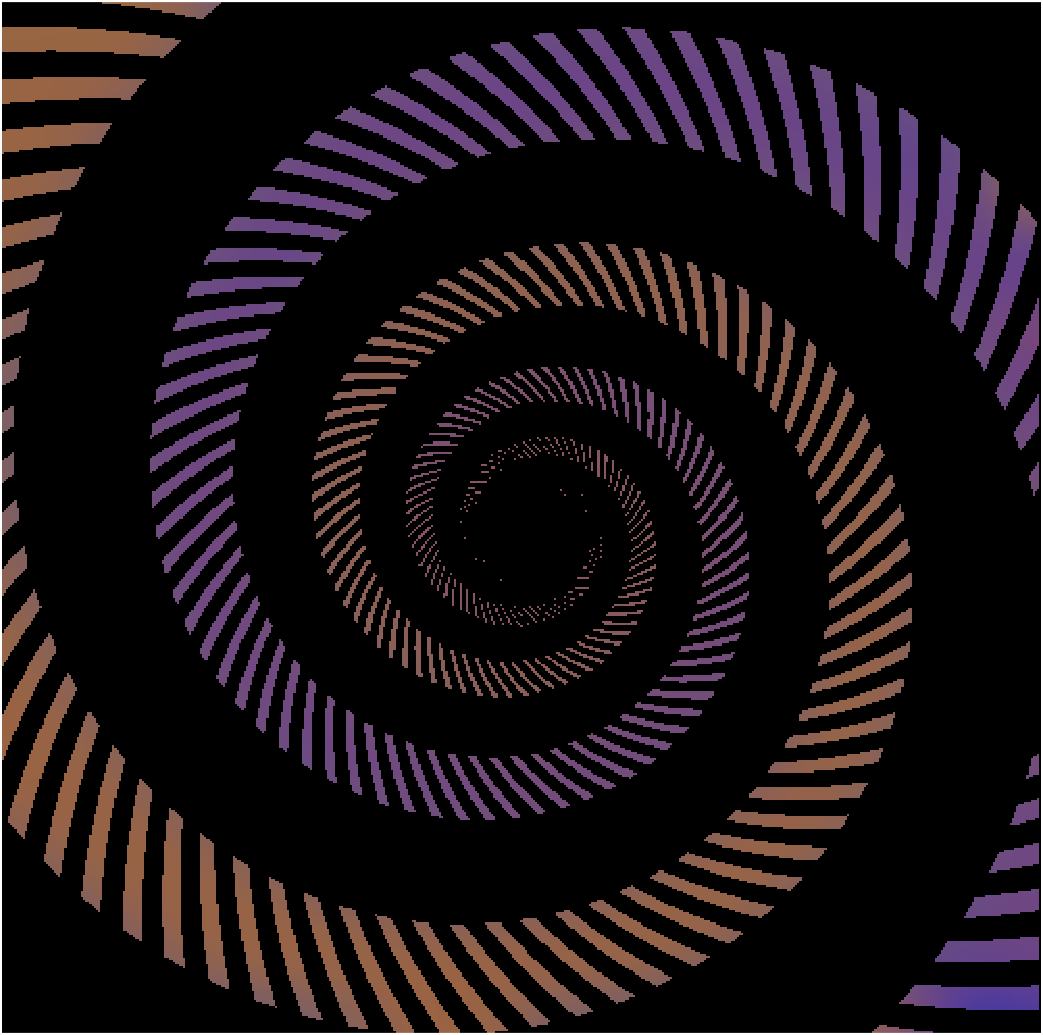}
    \\ 
    \includegraphics[width=.234\linewidth, height = 1.3cm]{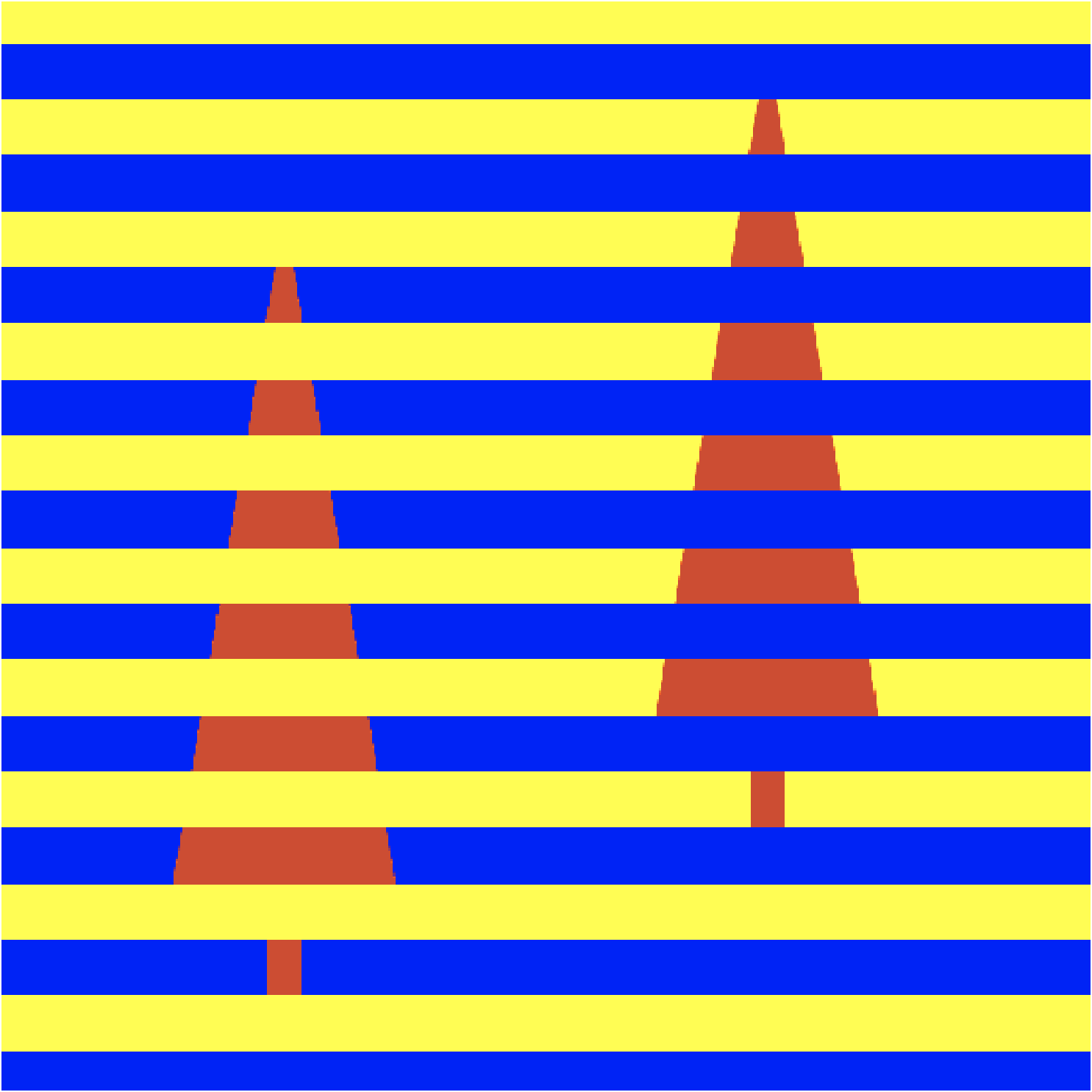} 
    &\includegraphics[width=.234\linewidth, height = 1.3cm]{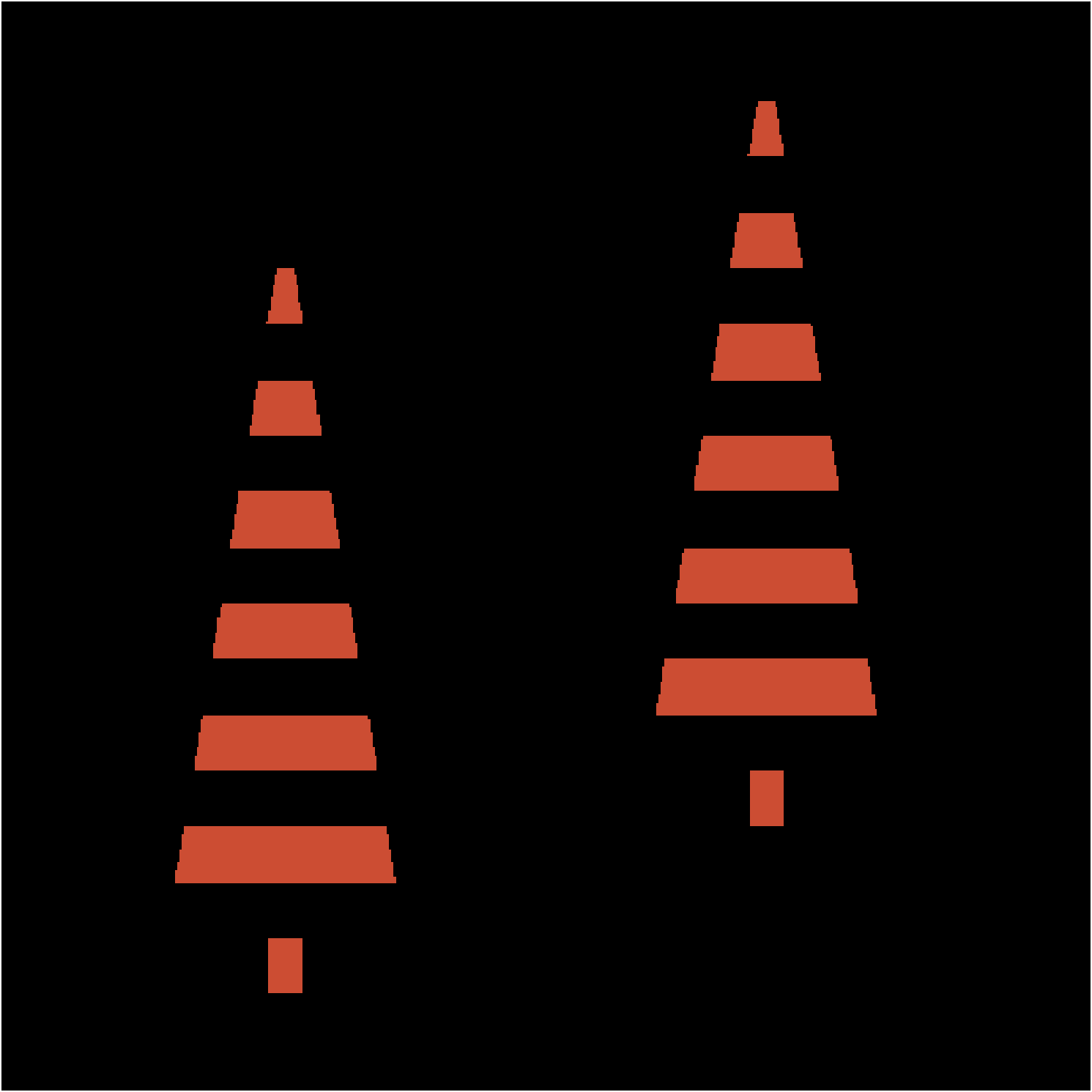}
    &\includegraphics[width=.234\linewidth, height = 1.3cm]{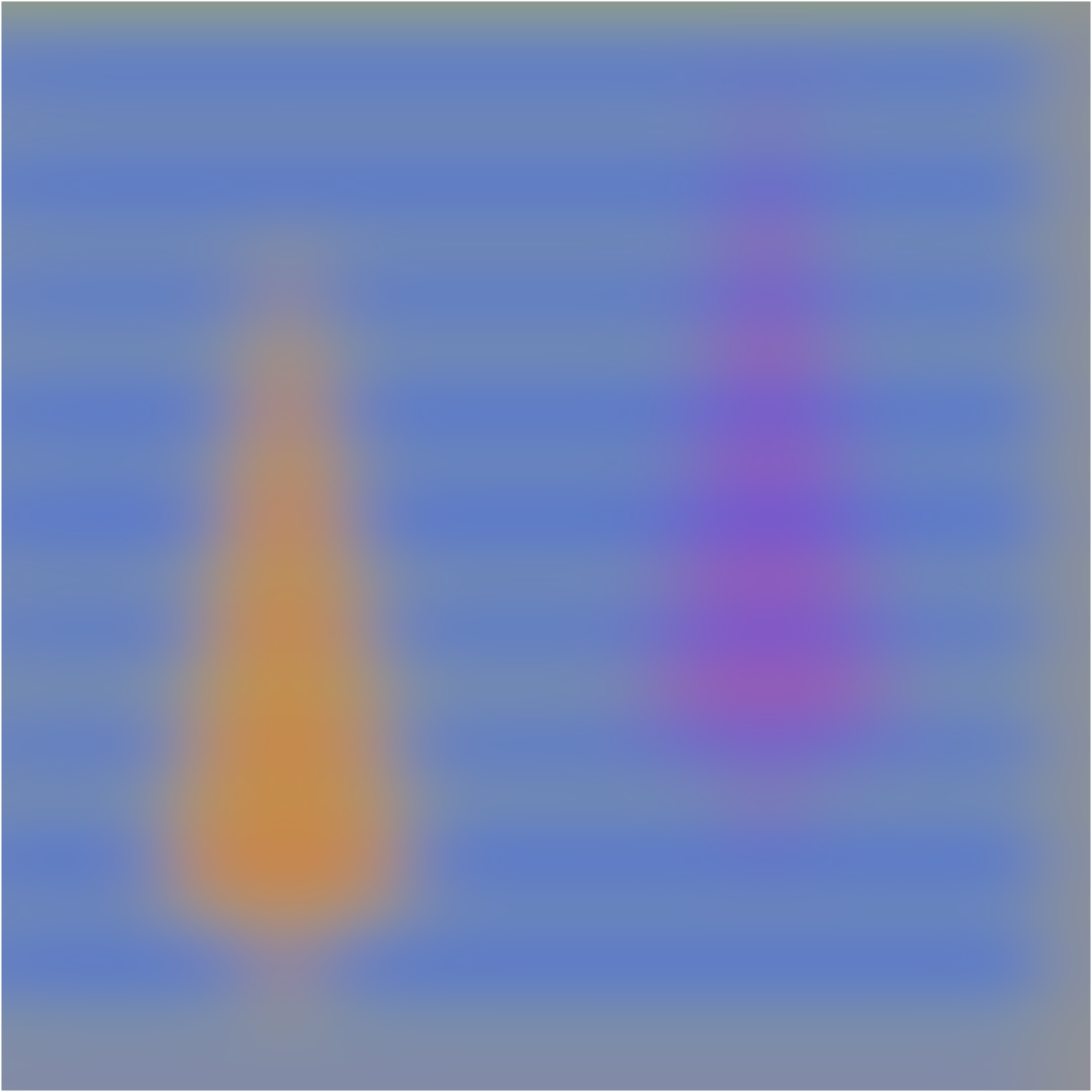}
    &\includegraphics[width=.234\linewidth, height = 1.3cm]{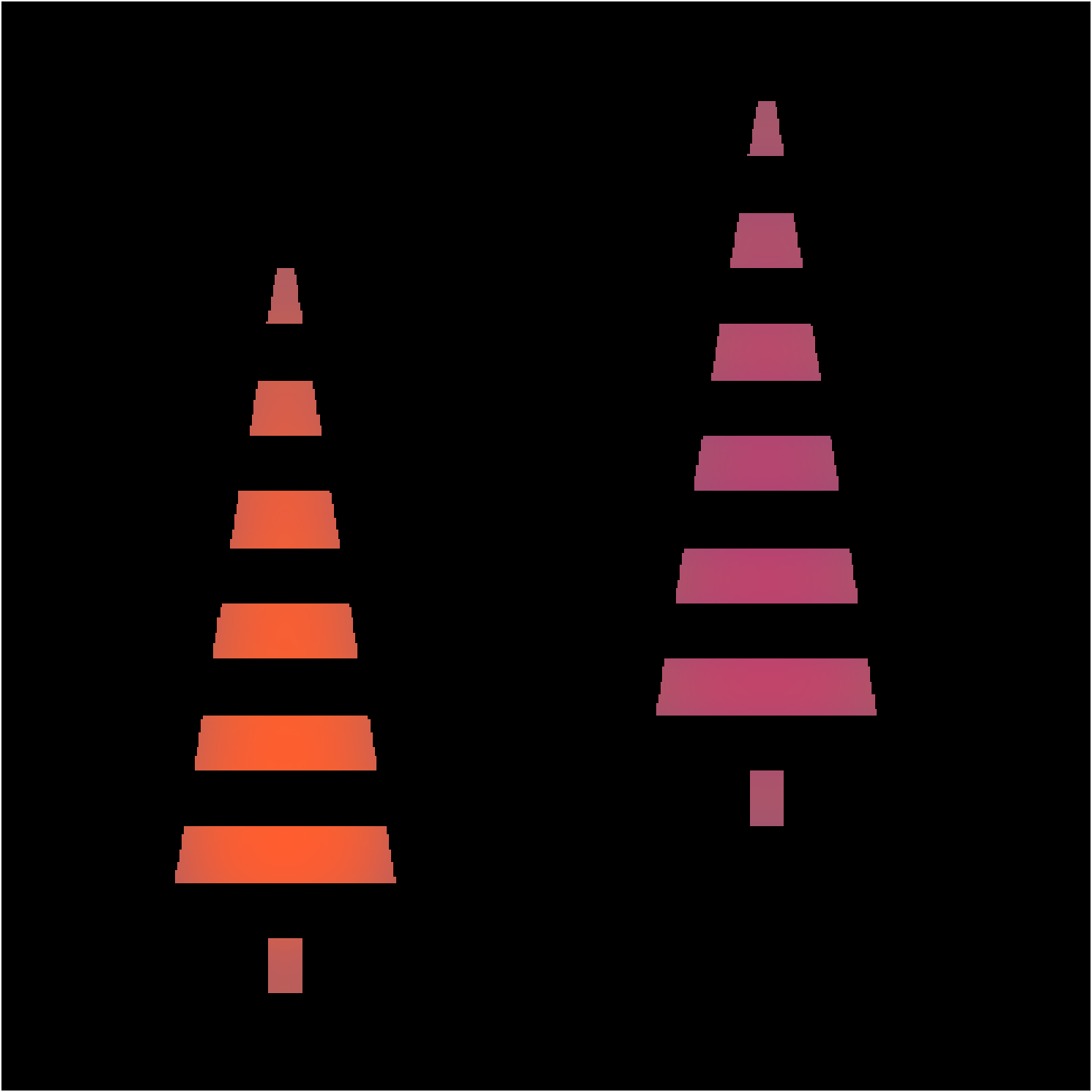}
    \\
    \includegraphics[width=.234\linewidth, height = 1.3cm]{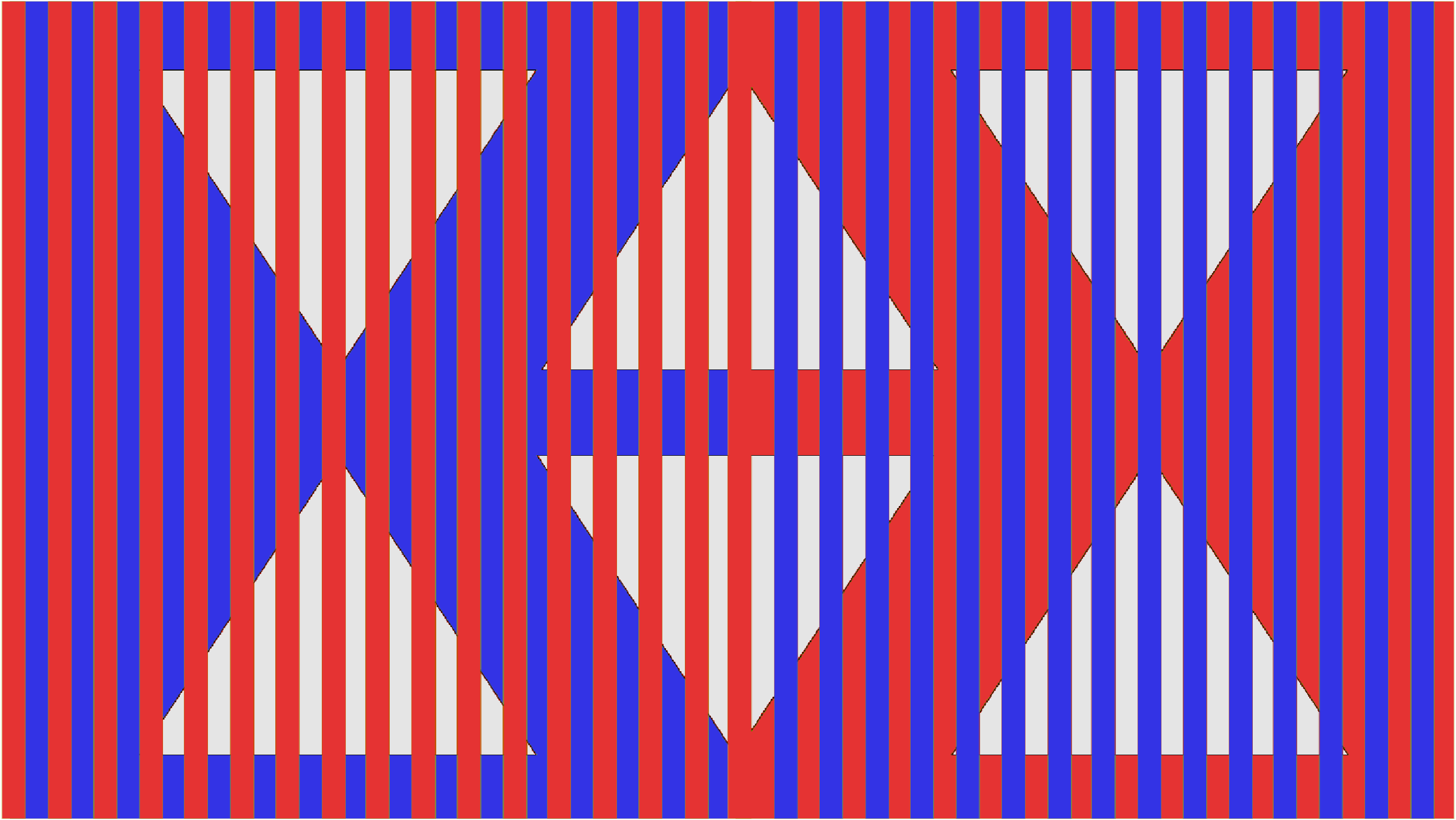} 
    &\includegraphics[width=.234\linewidth, height = 1.3cm]{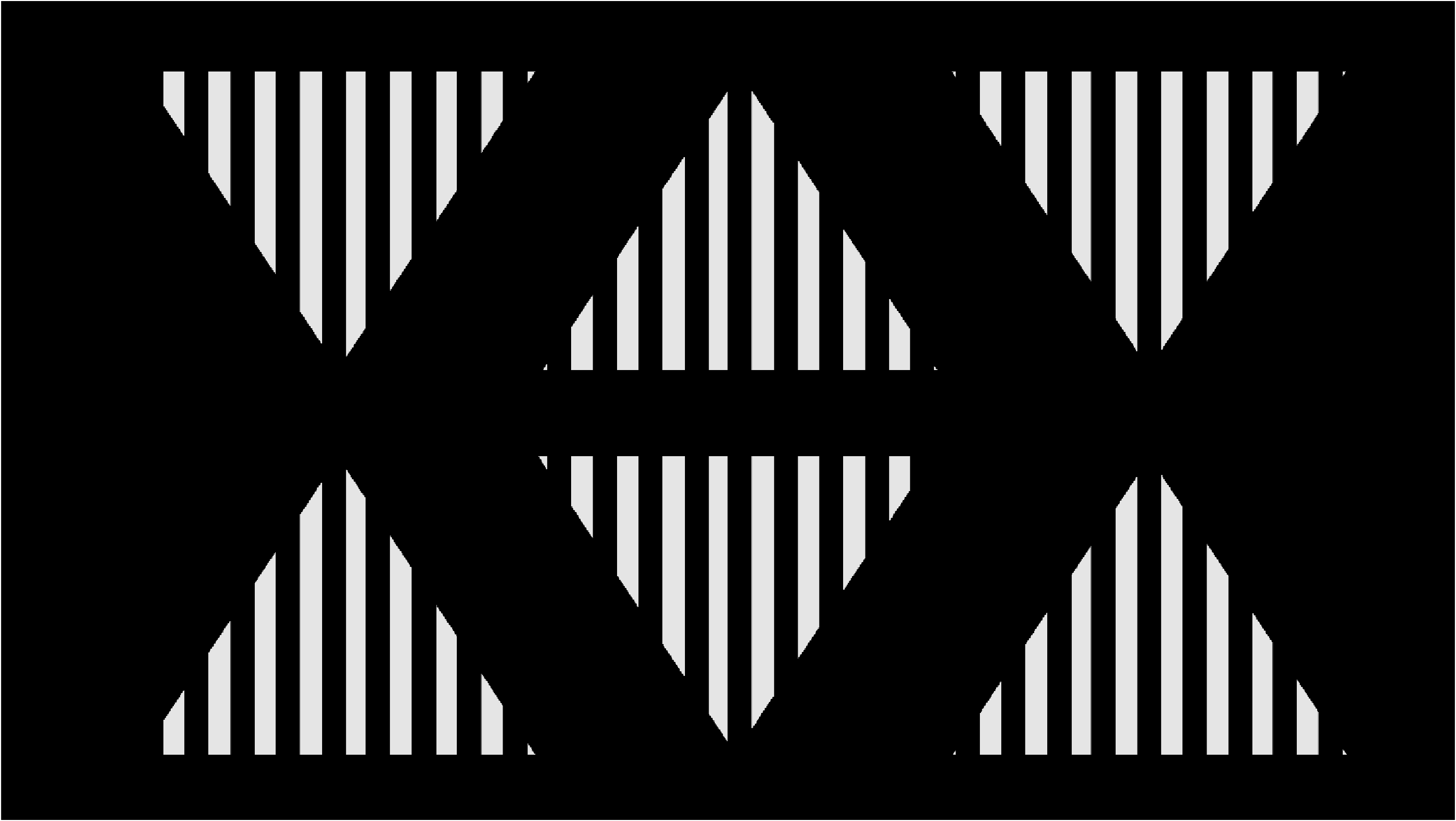}
    &\includegraphics[width=.234\linewidth, height = 1.3cm]{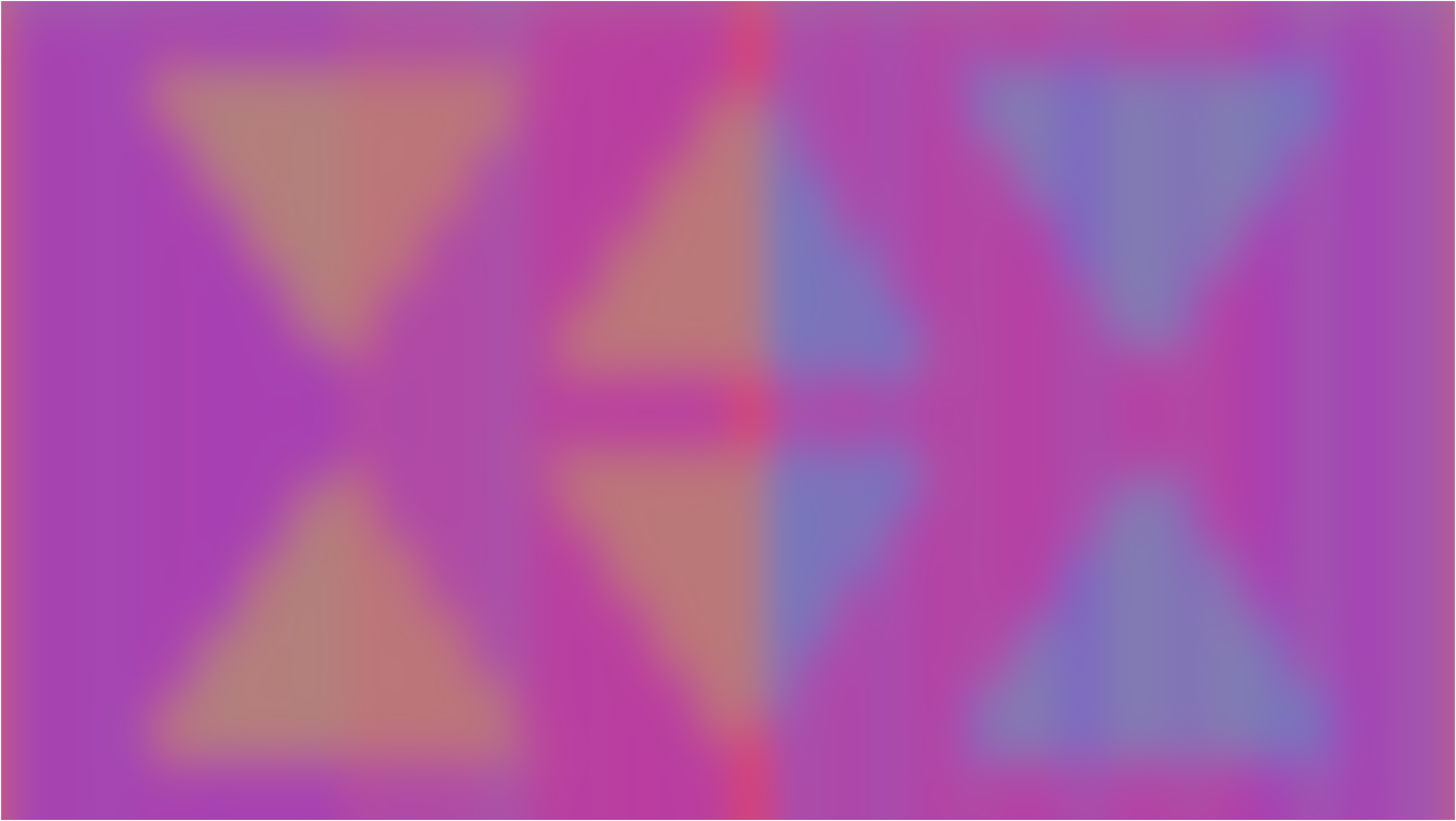}
    &\includegraphics[width=.234\linewidth, height = 1.3cm]{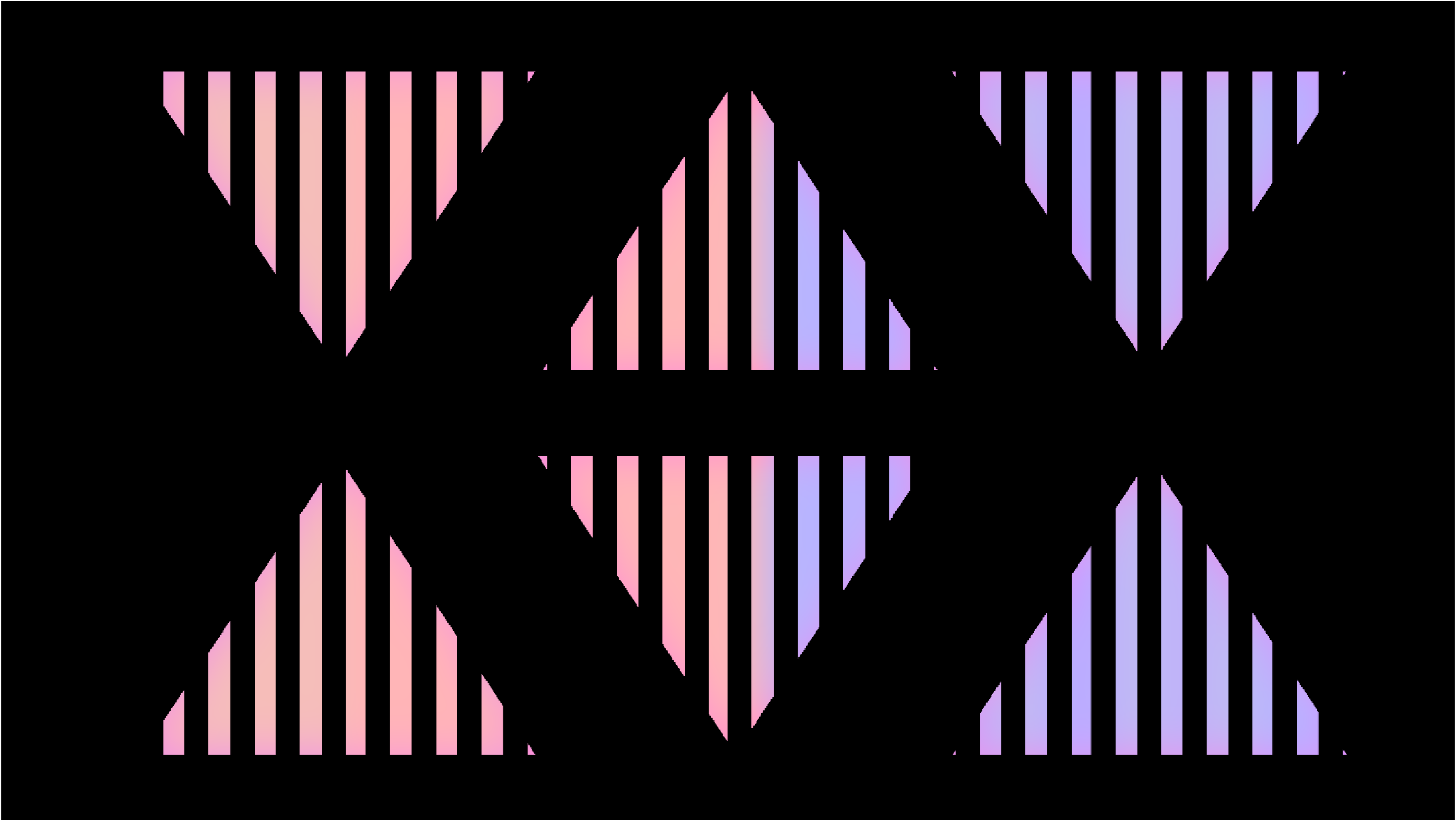}
    \end{tabular}
    \caption{Results of the proposed approach with different methods. (Top-to-bottom) The results of gray world, white-patch Retinex, shades of gray, gray edge, and double-opponent cells based color constancy.
    }
    \label{fig: excolorillusion}
\end{figure}

The key point in selecting parameters is that they have to be chosen so that they are able to mimic our sensation on images having both weak and strong illusion sensations. Thus, the inducer's frequency range is limited to the interval where we can create illusions. As presented in Fig.~\ref{fig:parameter_selection}, the best results are obtained when we take the block size as $8 \times 8$. Smaller block sizes are generally not able to produce satisfying results when our approach is applied with certain color constancy algorithms. This is not surprising since color constancy algorithms depend on image statistics, thus, the block size cannot be too small, since applying a color constancy algorithm to a small number of pixels could violate their assumptions, i.e., the gray world method is only valid when there is a sufficient number of different colors present in the scene~\cite{Ebner:2009}. On the other hand, when we consider larger block sizes, our sensation on color assimilation illusions cannot be accurately reproduced, i.e., the shapes and textures degrade, since the locality has to be preserved for color illusions~\cite {Ulucan/Ulucan/Ebner:2022a}.

Similar to the block size, $\sigma$ cannot be too large in order to preserve locality. As the block size, the controlling parameter is determined by visually analyzing the pixel-wise estimates instead of only considering the target regions, since alongside reproducing the colors we perceive, it is also important to preserve the shapes of the target regions in the illusions (Fig.~\ref{fig:parameter_selection}). Therefore, while in Fig.~\ref{fig:parameter_selection} more than one parameter may seem a proper choice, we need to select the suitable parameter combination that reproduces the colors we perceive as close as possible without damaging the shapes when all different color illusions in our set are considered. Thereupon, we determined $\sigma$ as $24$ which is $3$ times $\beta$ indicating that at least three estimations should fall inside the area of support. 

We present the results of our approach with different global color constancy methods on color assimilation illusions in Fig.~\ref{fig: excolorillusion}. From the examples, it is observable that with our modification, computational color constancy algorithms are able to respond to color illusions similar to the human visual system. Our approach transforms algorithms so that they can provide accurate pixel-wise estimations for illusions.

\section{\uppercase{Experiments for Color Constancy}} \label{exp: colorconstancy}
In this section, we present our experimental setup and results on color constancy, while we also show that the parameters selected purely from color assimilation illusions provide the best statistical outcomes on a multi-illuminant color constancy benchmark.

\subsection{Experimental Setup}
 
\subsubsection{Algorithms}
We investigate the effect of our approach on both multi-illuminant and global color constancy. We compare our outcomes with the following algorithms; white-patch Retinex~\cite{Land:1977}, gray world~\cite{Buchsbaum:1980}, shades of gray~\cite{Finlayson/Trezzi:2004}, $1^{st}$ order gray edge~\cite{Weijer/Gevers/Gijsenij:2007}, weighted gray edge~\cite{Gijsenij/Gevers/Weijer:2009}, double-opponent cells based color constancy~\cite{Gao/Yang/Li/Li:2015}, PCA based color constancy~\cite{Cheng/Prasad/Brown:2014}, color constancy with local surface reflectance estimation~\cite{Gao/Han/Yang/Li/Li:2014}, mean shifted gray pixels~\cite{Qian/Pertuz/Nikkanen/Kamarainen/Matas:2018}, gray pixels~\cite{Qian/Kamarainen/Nikkanen/Matas:2019}, block-based color constancy~\cite{Ulucan/Ulucan/Ebner:2022b}, biologically inspired color constancy~\cite{Ulucan/Ulucan/Ebner:2022a}, color constancy convolutional autoencoder~\cite{Laakom/Raitoharju/Iosifidis/Nikkanen/Gabbouj:2019}, sensor-independent illumination estimation~\cite{Afifi/Brown:2019}, cross-camera convolutional color constancy~\cite{Afifi/Barron/LeGendre/Tsai/Bleibel:2021}, local space average color~\cite{Ebner:2003}, Gijsenij~\textit{et al.}~\cite{Gijsenij/Lu/Gevers:2011}, conditional random fields~\cite{Beigpour/Riess/Weijer/Angelopoulou:2013}, N-white balancing~\cite{Akazawa/Kinoshita/Shiota/Kiya:2022}, visual mechanism based color constancy~\cite{Gao/Ren/Zhang/Li:2019}, retinal inspired color constancy~\cite{Zhang/Gao/Li/Du/Li/Li:2016}, color constancy weighting factors~\cite{Hussain/Akbari:2018}, color constancy adjustment based on texture of image~\cite{Hussain/Akbari/Halpin:2019}, GAN based color constancy~\cite{Das/Liu/Karaoglu/Gevers:2021}, and CNNs based color constancy~\cite{Bianco/Cusano/Schettini:2017}. While all traditional methods, whose codes are available are utilized without any modification or optimization, the results of the remaining algorithms are reported based on the publications of these works. As a final note, we performed the experiments on an Intel i7 CPU @ 2.7 GHz Quad-Core 16GB RAM machine using MATLAB R2021a. 

\subsubsection{Datasets}
\begin{figure}
    \centering
    \setlength{\tabcolsep}{1.5pt} 
    \renewcommand{\arraystretch}{1} 
    \begin{tabular}{c c c c}
    \includegraphics[width=.234\linewidth,  height = 1.2cm]{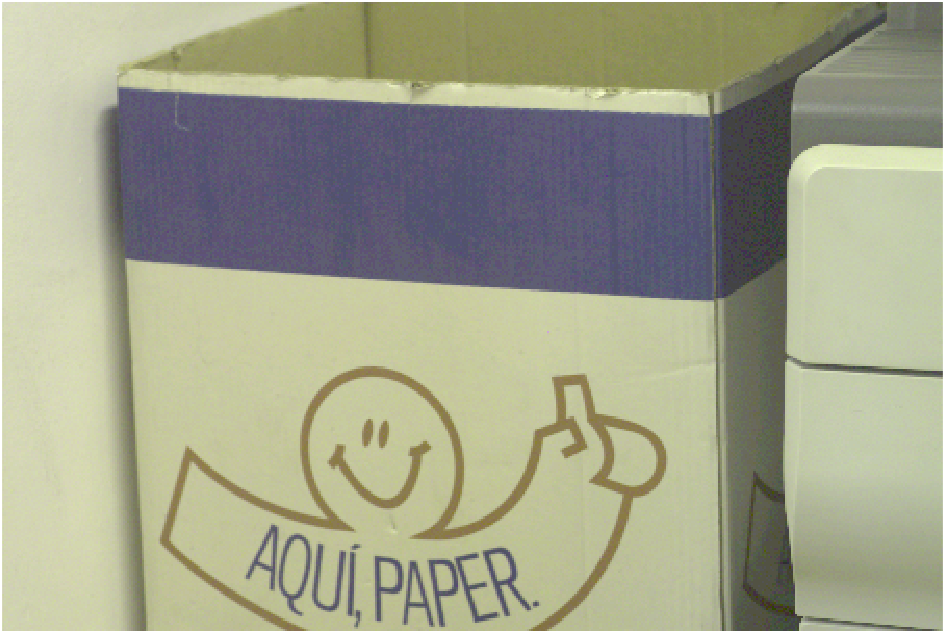}
    & \includegraphics[width=.234\linewidth,  height = 1.2cm]{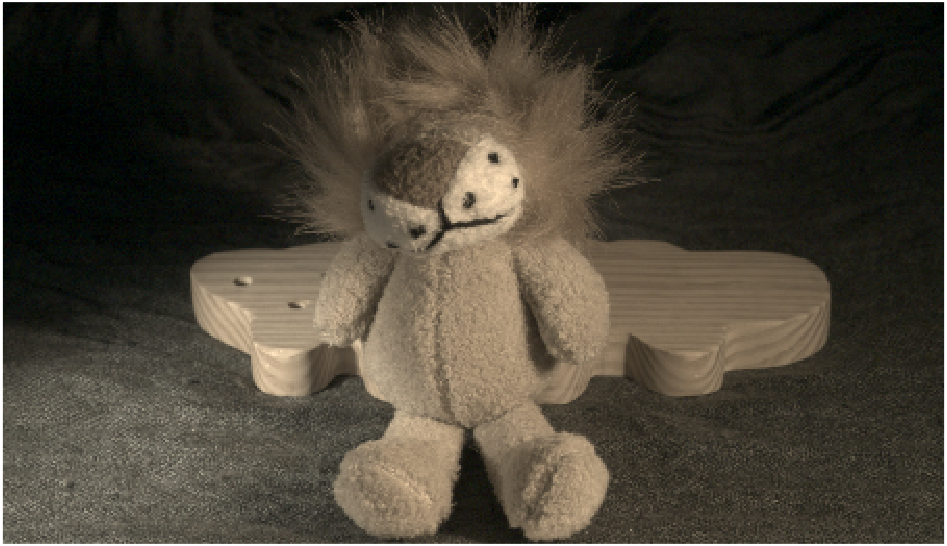}
    & \includegraphics[width=.234\linewidth,  height = 1.2cm]{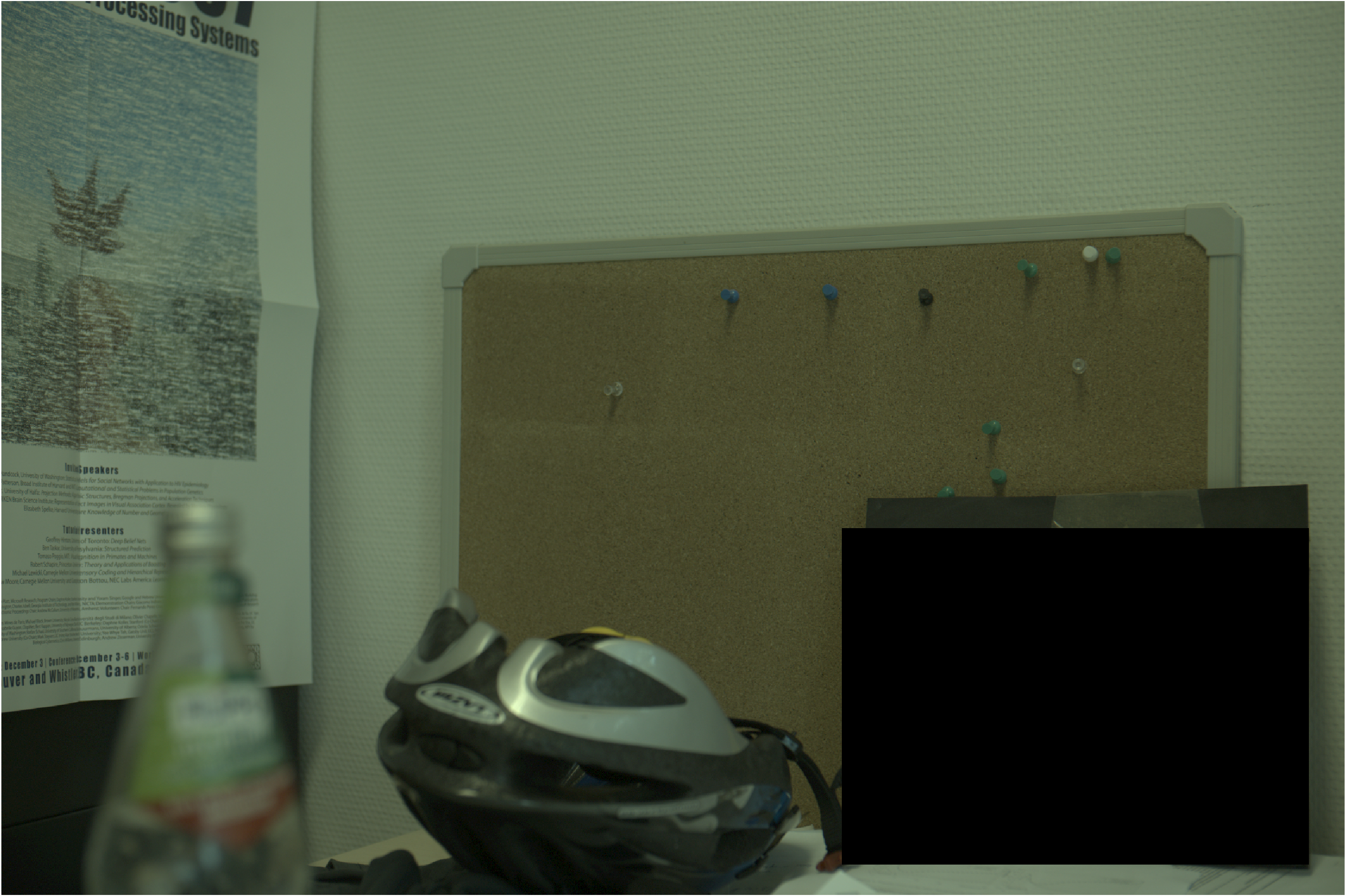}
    & \includegraphics[width=.234\linewidth,  height = 1.2cm]{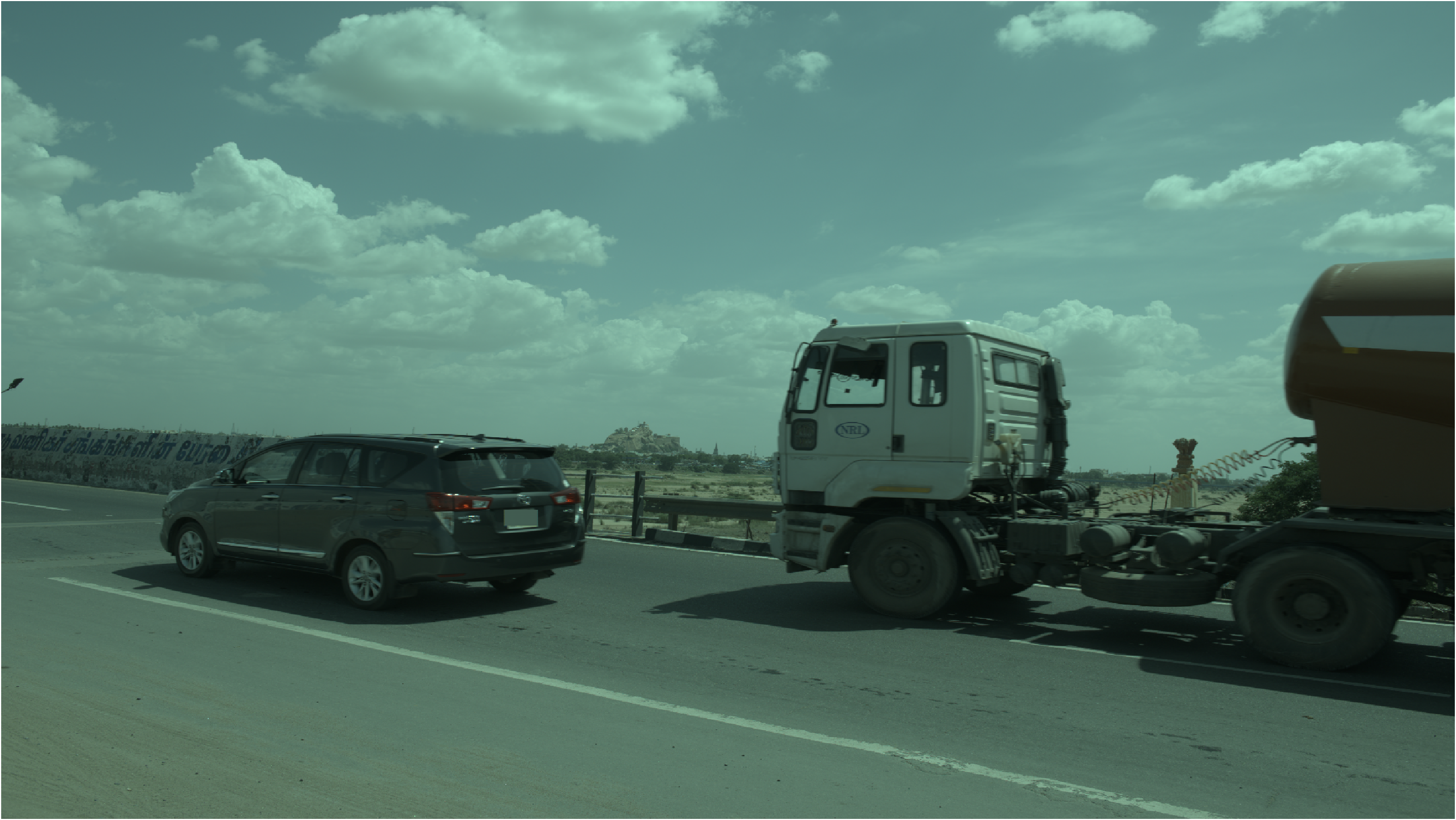}
    \end{tabular}
    \caption{Examples from color constancy benchmarks. (Left-to-right) The real-set and the laboratory-set of the MIMO dataset, the RECommended ColorChecker, and the INTEL-TAU datasets, respectively.}
\label{fig:cc_benchmark}
\end{figure}

To benchmark our technique on images captured under multiple light sources, we choose the Multiple Illuminant and Multiple Object (MIMO) dataset~\cite{Beigpour/Riess/Weijer/Angelopoulou:2013}, one of the most notable multi-illuminant datasets in the field of computational color constancy, since many algorithms have already been benchmarked on this dataset~\cite{Buzzelli/Zini/Bianco/Ciocca/Schettini/Tchobanou:2023}. The MIMO dataset contains scenes, which are taken under controlled illumination conditions, and their pixel-wise ground truths. To test our approach on images captured under global illumination conditions, we use two color constancy datasets; Recommended ColorChecker and INTEL-TAU datasets~\cite{Hemrit/Finlayson/Gijsenij/Gehler/Bianco/Funt/Drew/Shi:2018,Laakom/Raitoharju/Nikkanen/Iosifidis/Gabbouj:2021}. The Recommended ColorChecker dataset is the modified version of the Gehler-Shi dataset~\cite{Gehler/Rother/Blake/Minka/Sharp:2008}. It contains indoor and outdoor scenes, which are captured with two different capturing devices. The INTEL-TAU dataset is one of the recent color constancy benchmarks containing both indoor and outdoor images, which are captured with three different cameras. During the evaluation, images captured with different cameras are combined to form a single set since it is crucial to test the algorithms with images, whose spectral distribution is unknown. Examples from the utilized datasets are given in Fig.~\ref{fig:cc_benchmark}.

\subsubsection{Error Metric}
To quantitatively evaluate the modified version of the color constancy algorithms, we use the angular error which is a standardized error metric in the field of color constancy. The angular error ($\varepsilon$) between the ground truth light source ($L_{gt}$) and the estimated illuminant ($L_{est}$) can be calculated as follows;

\begin{equation}
    \varepsilon (L_{gt}, L_{est}) = cos^{-1} \begin{pmatrix} \frac{L_{gt} L_{est}}{\left\| L_{gt} \right\| \left\| L_{est} \right\|} 
\end{pmatrix}.
\label{eqn: angerr}
\end{equation}

We report the mean and the median of the errors for the multi-illuminant cases, while we report the mean, the median, the best $25\%$, the worst $25\%$, and the maximum of the angular error for the global cases.

\subsection{Discussion on Color Constancy}
First of all, we provide an analysis of the parameters selected during the investigation of color illusions to show that these parameters also give the best outcomes for multi-illuminant color constancy (Table~\ref{tab:para_select}). The best results on the MIMO dataset are obtained when $\beta$ is chosen as $8$, and $\sigma$ is set to $24$ as in the reproduction of color illusions. The fact that the parameters obtained purely by analyzing color assimilation illusions provide also the best results on a multi-illuminant color constancy benchmark underlines the link between color illusions and multi-illuminant color constancy which implies that these two phenomena should be further investigated together as we argue. 

\begin{table}
\centering
\caption{Analysis of the parameters on the MIMO dataset. The mean of the pixel-wise angular error between the estimates and the ground truth is provided. The best combination is highlighted. The parameter combinations that are selected from color illusions work better for color constancy.}
\resizebox{\linewidth}{!}{%
\setlength{\tabcolsep}{5.5pt} 
\renewcommand{\arraystretch}{1.3} 
\begin{tabular}{lccccccccccc}
\toprule \toprule
 &  & \multicolumn{10}{c}{Scaling Factor $\sigma$}\\
 &  &  $2$ & $4$ & $8$ & $12$ & $16$ & $20$ & $24$ & $28$ & $32$ & $48$ \\ \cline{2-12} \cline{2-12}
\parbox[t]{2mm}{\multirow{10}{*}{\rotatebox[origin=c]{90}{Block Size $\beta$}}}
 & \multicolumn{1}{c |}{$4$} & $4.828$ & $4.374$ & $4.058$ & $3.929$ & $3.875$ & $3.859$ & $3.859$ & $3.865$ & $3.874$ & $3.918$ \\ 
 & \multicolumn{1}{c |}{$8$} & $4.807$ & $4.281$ & $3.883$ & $3.735$ & $3.677$ & $3.659$ & $\mathbf{3.658}$ & $3.666$ & $3.675$ & $3.721$ \\
 & \multicolumn{1}{c |}{$12$} & $4.897$ & $4.450$ & $4.004$ & $3.821$ & $3.737$ & $3.706$ & $3.701$ & $3.705$ & $3.713$ & $3.746$ \\
 & \multicolumn{1}{c |}{$16$} & $4.932$ & $4.591$ & $4.100$ & $3.894$ & $3.804$ & $3.763$ & $3.748$ & $3.749$ & $3.758$ & $3.819$ \\
 & \multicolumn{1}{c |}{$20$} & $4.918$ & $4.692$ & $4.255$ & $4.052$ & $3.954$ & $3.905$ & $3.882$ & $3.871$ & $3.867$ & $3.867$ \\
 & \multicolumn{1}{c |}{$24$} & $4.964$ & $4.770$ & $4.335$ & $4.104$ & $3.998$ & $3.948$ & $3.931$ & $3.928$ & $3.931$ & $3.968$ \\
 & \multicolumn{1}{c |}{$28$} & $4.973$ & $4.927$ & $4.458$ & $4.173$ & $4.014$ & $3.928$ & $3.884$ & $3.859$ & $3.845$ & $3.842$ \\
 & \multicolumn{1}{c |}{$32$} & $5.083$ & $5.042$ & $4.599$ & $4.322$ & $4.155$ & $4.054$ & $3.989$ & $3.949$ & $3.924$ & $3.895$ \\
 & \multicolumn{1}{c |}{$48$} & $4.969$ & $5.481$ & $4.759$ & $4.583$ & $4.424$ & $4.303$ & $4.212$ & $4.144$ & $4.091$ & $3.954$ \\ \hline \hline
\end{tabular}}
\label{tab:para_select}
\end{table}

\begin{table}
\centering
\caption{Statistical results on the MIMO dataset. For each metric, the top three results are highlighted.}
\resizebox{\linewidth}{!}{%
\renewcommand{\arraystretch}{1.3} 
\begin{tabular}{c l | c c | c c }
\toprule \toprule
\multicolumn{2}{l}{}      
& \multicolumn{2}{c}{\textbf{Real-World}} & \multicolumn{2}{c}{\textbf{Laboratory}}  
\\ 
\multicolumn{2}{l}{Algorithms}      
& \multicolumn{1}{c}{Mean} & \multicolumn{1}{c}{Median}
& \multicolumn{1}{c}{Mean} & \multicolumn{1}{c}{Median} 
\\ \hline
\parbox[t]{2mm}{\multirow{14}{*}{\rotatebox[origin=c]{90}{\textbf{Single-Illuminant Algorithms}}}} 
%
%
& White-Patch Retinex    & $6.8$  & $5.7$   & $7.8$  & $7.6$ 
\\ 
& Gray World  & $5.3$  & $4.3$   & $3.5$  & $2.9$  
\\ 
& Shades of Gray & $6.2$  & $3.7$   & $4.9$  & $4.6$ 
\\ 
& $1^{st}$ order Gray Edge  & $8.0$  & $4.7$  & $4.3$  & $4.1$ 
\\
& Weighted Gray Edge     & $7.9$  & $4.1$  & $4.4$  & $4.0$ 
\\ 
& Double-Opponent Cells based Color Constancy         & $7.9$  & $5.0$   & $4.6$  & $4.4$ 
\\
& PCA based Color Constancy  & $7.7$  & $3.5$ & $4.1$  & $3.8$ 
\\ 
& Local Surface Reflectance Estimation       & $4.9$  & $3.8$ & $3.9$  & $3.5$ 
\\ 
& Mean Shifted Gray Pixels     & $5.8$  & $5.0$ & $13.3$  & $12.6$ 
\\
& Block-based Color Constancy    & $4.8$  & $3.6$  & $3.1$  & $2.8$ 
\\
& Biologically Inspired Color Constancy    & $5.0$  & $4.3$  & $4.2$  & $4.1$
\\
& Color Constancy Convolutional Autoencoder    & $12.4$  & $12.3$  & $13.9$  & $14.1$ 
\\
& Sensor-Independent Color Constancy    & $5.9$  & $5.1$   & $9.0$  & $9.0$ 
\\
& Cross-Camera Convolutional Color Constancy    & $11.9$  & $13.0$   & $7.0$  & $7.1$ 
\\
\hdashline
\parbox[t]{2mm}{\multirow{21}{*}{\rotatebox[origin=c]{90}{\textbf{Multi-Illuminant Algorithms}}}} 
& Local Space Average Color      & $4.9$  & $4.2$ & $2.7$  & $2.5$ 
\\
& Gijsenij~\textit{et al.} with White-Patch Retinex      & $4.2$  & $3.8$ & $5.1$  & $4.2$ 
\\
& Gijsenij~\textit{et al.} with Gray-World    & $4.4$  & $4.3$ & $6.4$  & $5.9$
\\
& Gijsenij~\textit{et al.} with $1^{st}$ order Gray-Edge     & $9.1$  & $9.2$ & $4.8$  & $4.2$
\\
& Conditional Random Fields with White-Patch Retinex     & $4.1$  & $3.3$ & $3.0$  & $2.8$ 
\\
& Conditional Random Fields with Gray-World    & $3.7$  & $3.4$ & $3.1$  & $2.8$ 
\\
& Conditional Random Fields with $1^{st}$ order Gray-Edge      & $4.0$  & $3.4$ & $2.7$  & $2.6$ 
\\
& N-White Balancing with White-Patch Retinex     & $4.1$  & $3.4$ & $2.6$ & $\mathbf{2.2}$ 
\\ 
& N-White Balancing with Gray World     & $4.6$  & $4.5$ & $3.7$ & $3.1$ 
\\ 
& N-White Balancing with Shades of Gray     & $4.2$  & $3.8$ & $2.8$ & $2.3$ 
\\ 
& N-White Balancing with $1^{st}$ order Gray Edge     & $4.7$  & $3.6$ & $\mathbf{2.5}$ & $\mathbf{2.2}$ 
\\ 
& Visual Mechanism based Color Constancy with Bottom-Up     & $5.0$  & $4.0$ & $3.7$ & $3.4$ 
\\ 
& Visual Mechanism based Color Constancy with Top-Down     & $3.8$  & $\mathbf{2.9}$ & $2.8$ & $2.8$ 
\\ 
& Retinal Inspired Color Constancy     & $5.2$  & $4.3$ & $3.2$ & $2.7$ 
\\ 
& Color Constancy Weighting Factors    & $3.8$  & $3.8$ & $\mathbf{1.6}$ & $\mathbf{1.5}$
\\
& Color Constancy Adjustment based on Texture of Image     & $3.8$  & $3.8$ & $2.6$ & $2.6$ 
\\ 
& Gray Pixels with 2 clusters       & $3.7$  & $3.3$ & $3.0$  & $2.5$ 
\\ 
& Gray Pixels with 4 clusters       & $3.9$  & $3.4$ & $2.7$  & $\mathbf{2.2}$ 
\\ 
& Gray Pixels with 6 clusters       & $3.9$  & $3.4$ & $2.6$  & $\mathbf{2.1}$ 
\\ 
& CNNs-based Color Constancy     & $\mathbf{3.3}$  & $\mathbf{3.1}$ & $\mathbf{2.3}$ & $\mathbf{2.2}$ 
\\ 
& GAN-based Color Constancy       & $\mathbf{3.5}$  & $\mathbf{2.9}$ & - & -
\\ 
\hdashline
& Proposed w/ White-Patch Retinex  & $\mathbf{3.6}$ & $\mathbf{2.9}$ & $2.8$  & $2.5$ 
\\ 
& Proposed w/ Gray World       & $3.9$ & $3.5$ & $2.9$  & $2.7$ 
\\ 
& Proposed w/ Shades of Gray       & $3.7$ & $\mathbf{2.9}$ & $2.9$  & $2.6$ 
\\ 
& Proposed w/ $1^{st}$ order Gray Edge       & $3.9$ & $3.3$ & $2.8$  & $2.5$ 
\\ 
& Proposed w/ Weighted Gray Edge       & $3.8$ & $3.3$ & $2.8$  & $2.5$ 
\\ 
& Proposed w/ Double-Opponent Cells based Color Constancy    & $3.8$ & $\mathbf{3.2}$ & $2.9$  & $2.7$ \\ \hline \hline
\end{tabular}}
\label{tab:table_mimo}
\end{table}

\begin{figure}
    \centering
    \setlength{\tabcolsep}{1.5pt} 
    \renewcommand{\arraystretch}{0.5} 
    \begin{tabular}{c c c}
        \includegraphics[width=.312\linewidth]{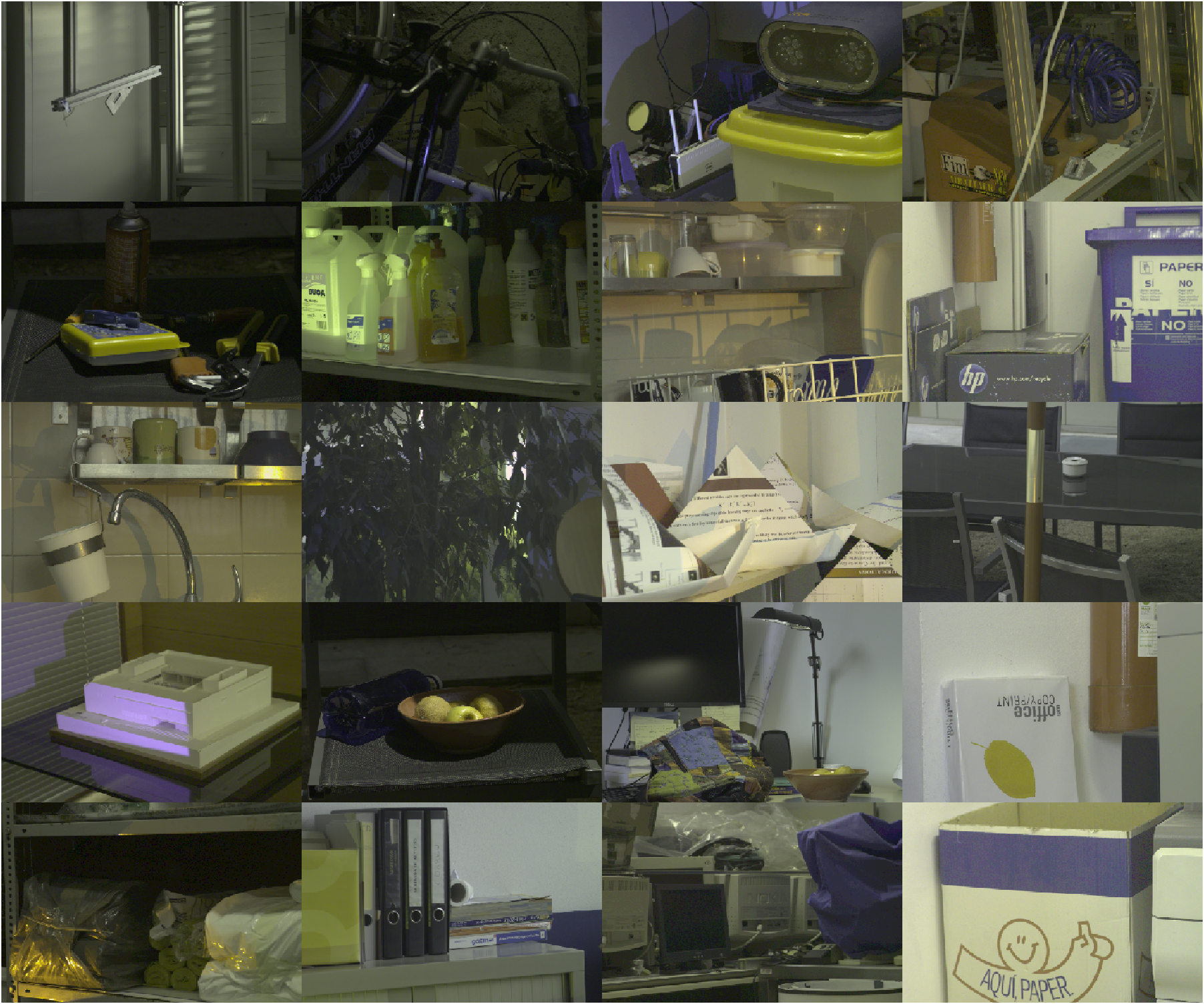} 
        & \includegraphics[width=.312\linewidth]{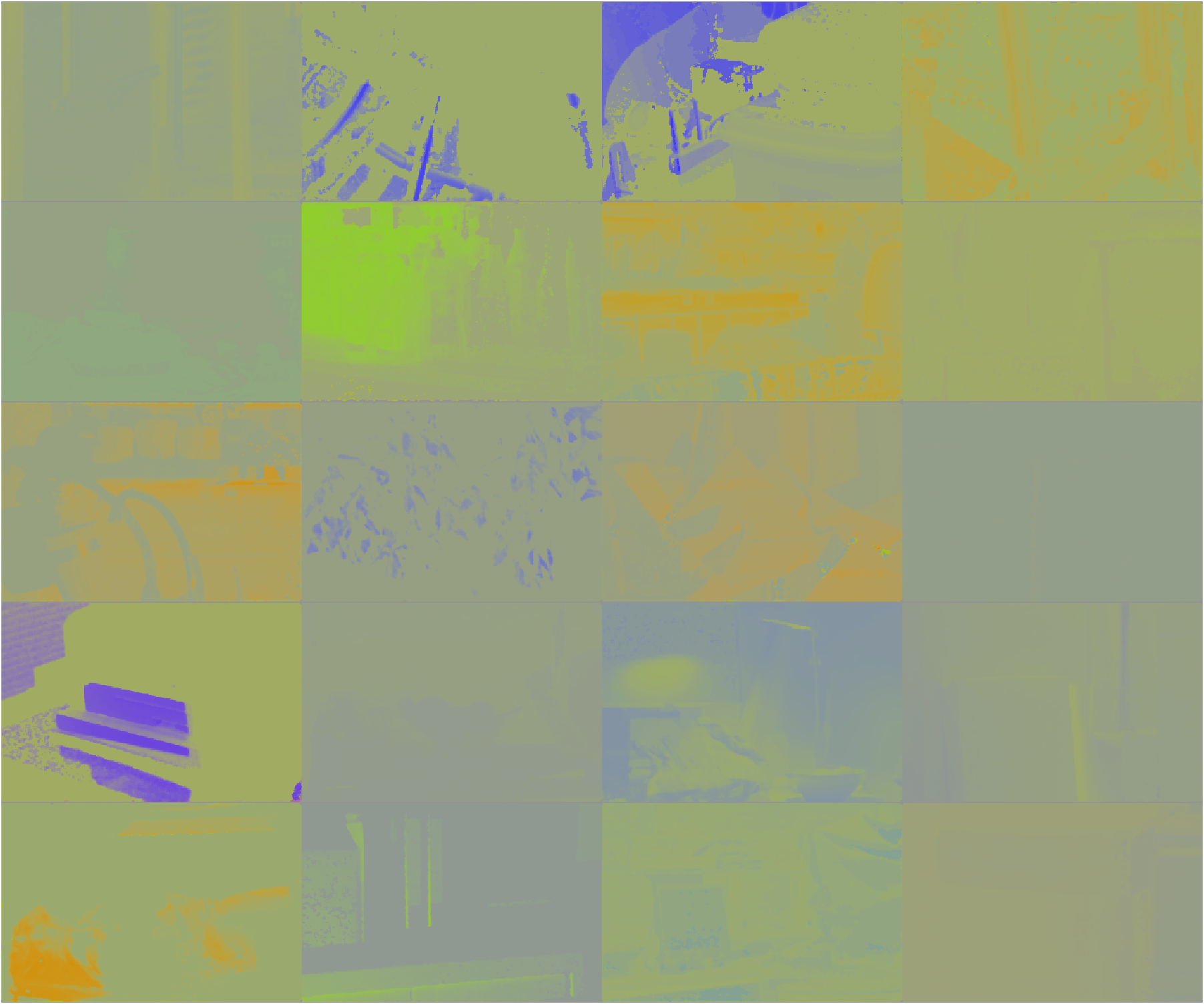}
        & \includegraphics[width=.312\linewidth]{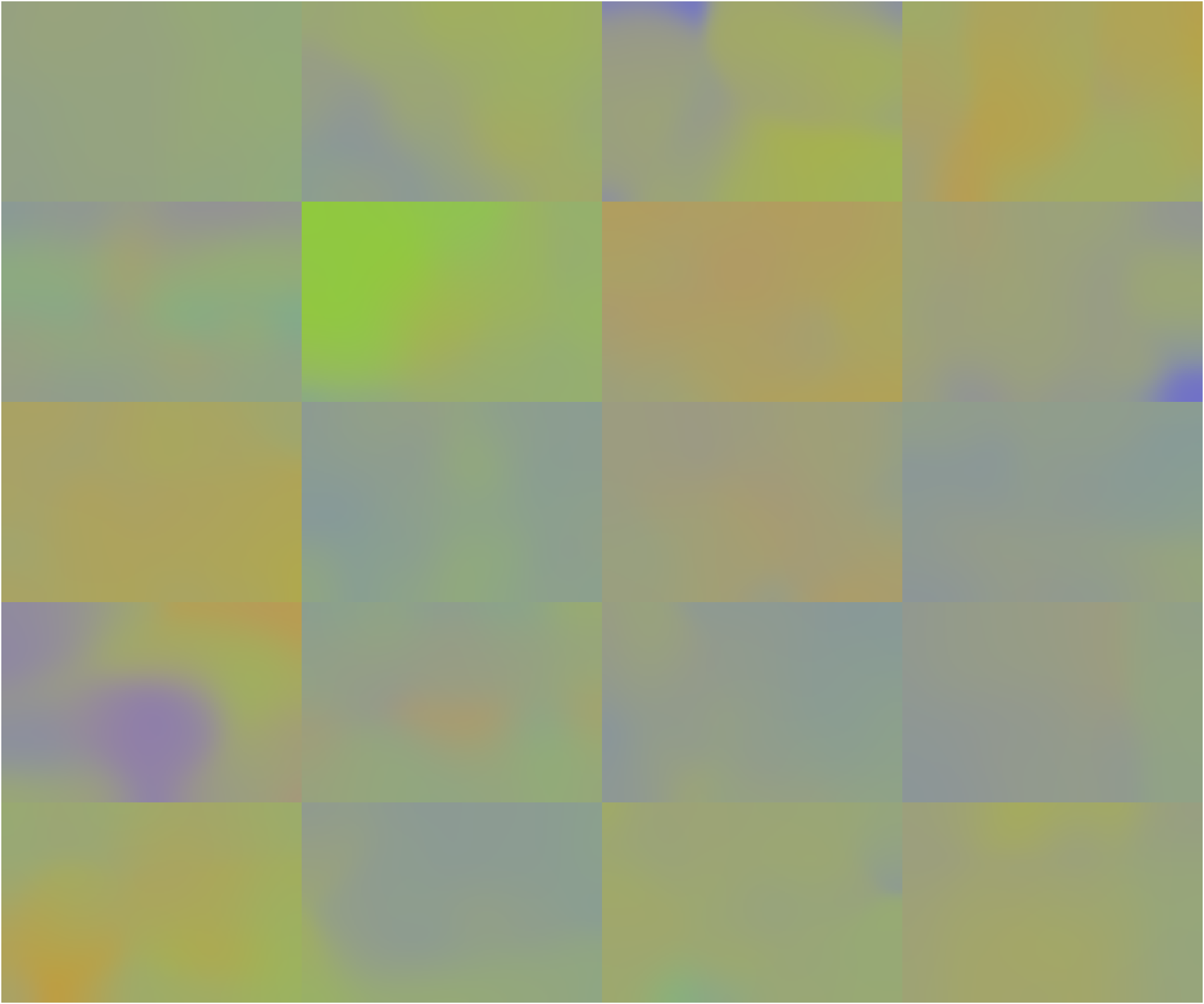}
    \\
       \includegraphics[width=.312\linewidth]{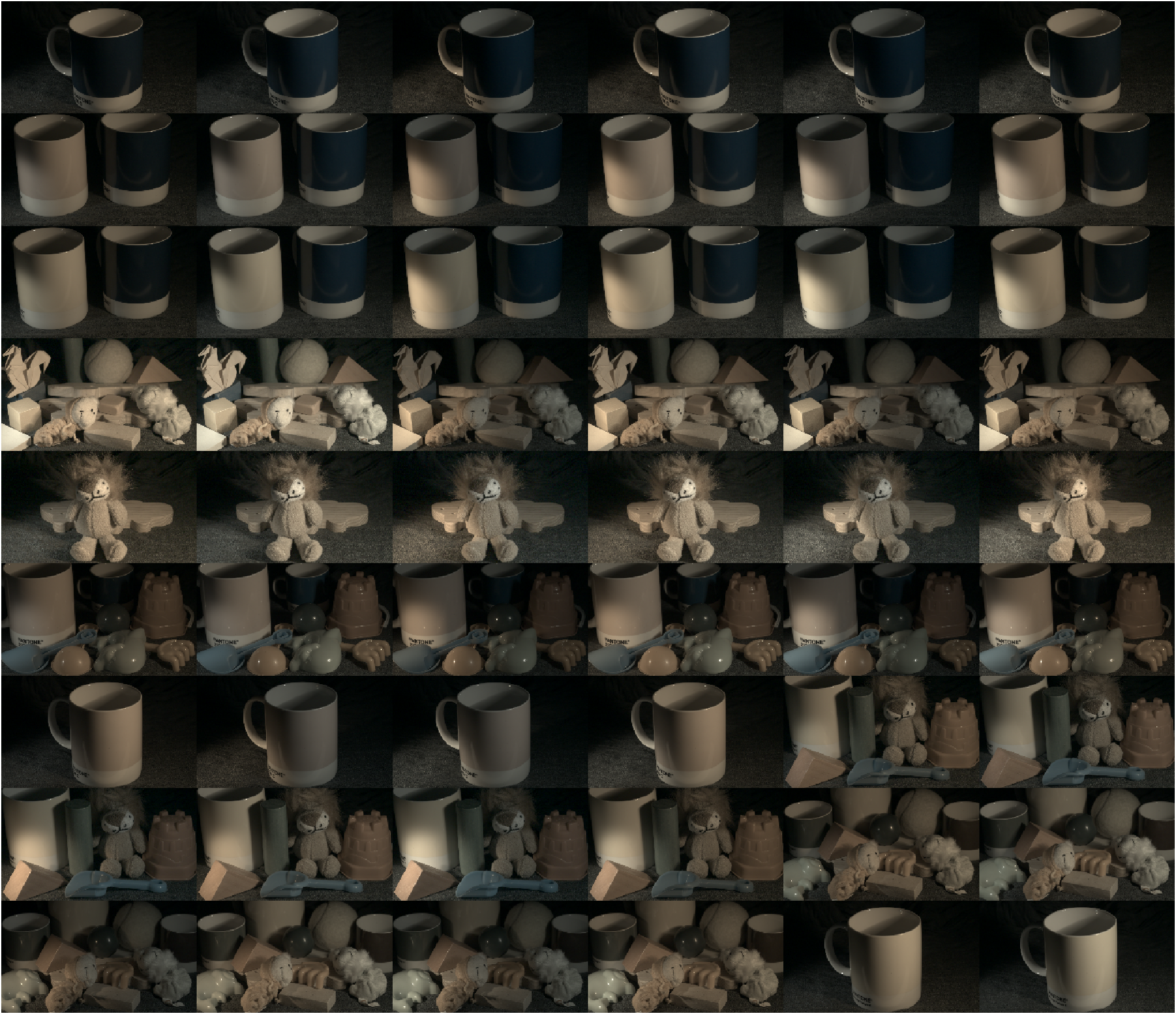}  
       & \includegraphics[width=.312\linewidth]{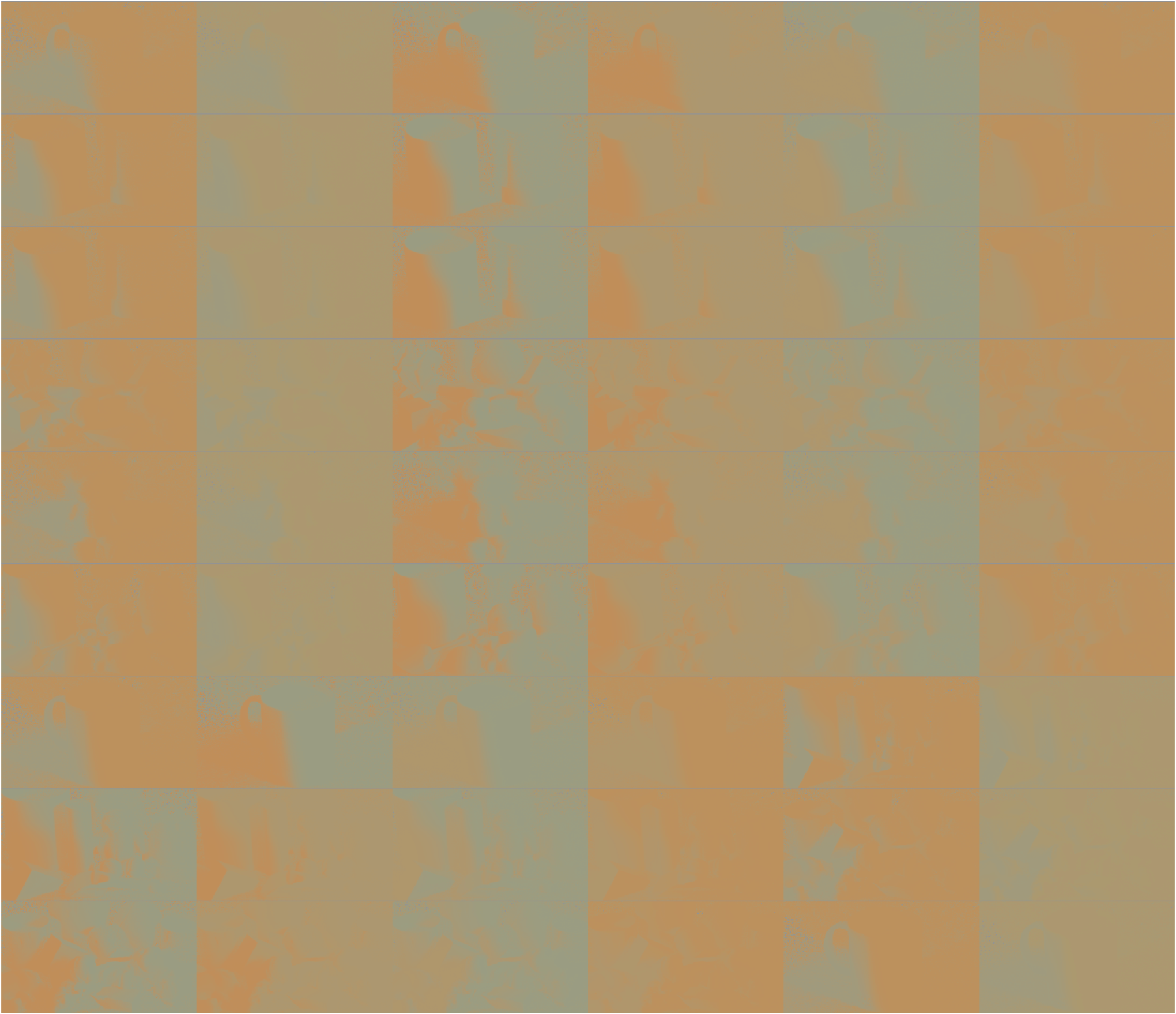} 
       & \includegraphics[width=.312\linewidth]{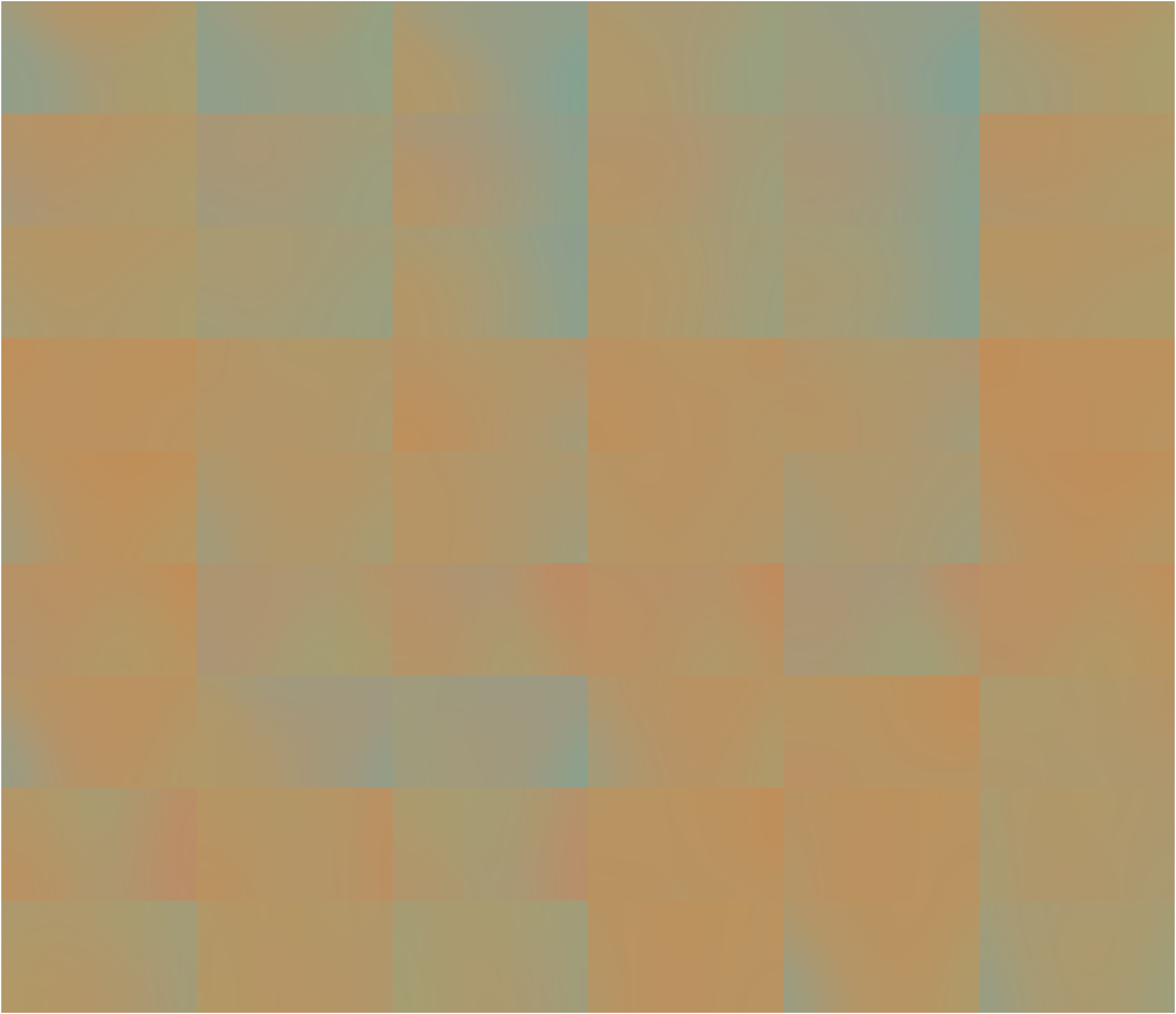} 
    \end{tabular}
    \caption{Results on MIMO dataset, the top row contains scenes from the real-world set, and the bottom row shows images from the laboratory set. (Left-to-right) The input scenes, ground truths, and pixel-wise estimations via the proposed approach with white-patch Retinex.}
    \label{fig:all_mimo_set}
\end{figure}

\begin{figure}
    \centering
    \setlength{\tabcolsep}{1.5pt} 
    \renewcommand{\arraystretch}{0.5} 
    \begin{tabular}{c c c c}
        \includegraphics[width=.234\linewidth, height = 1.2cm]{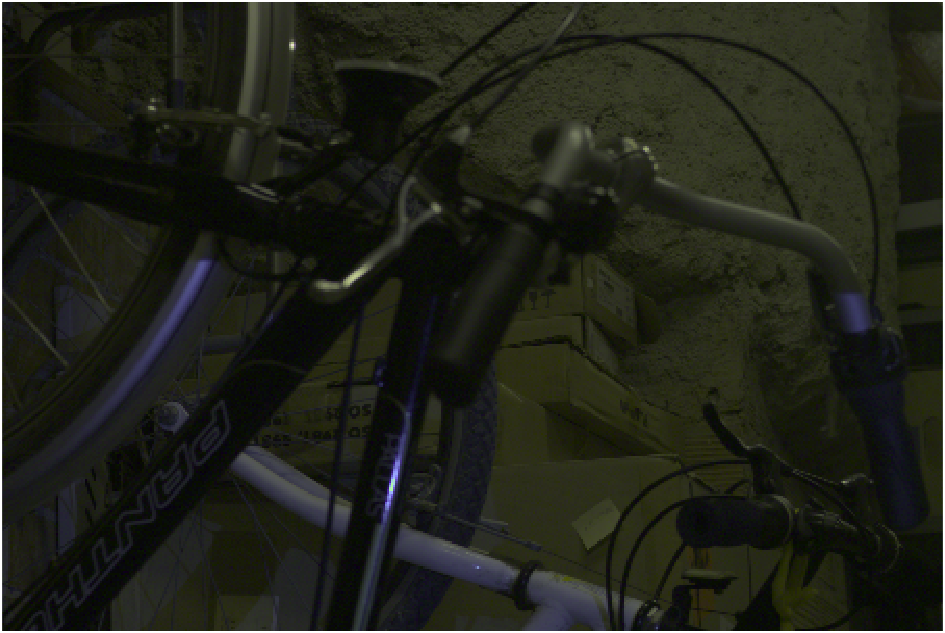} 
        & \includegraphics[width=.234\linewidth, height = 1.2cm]{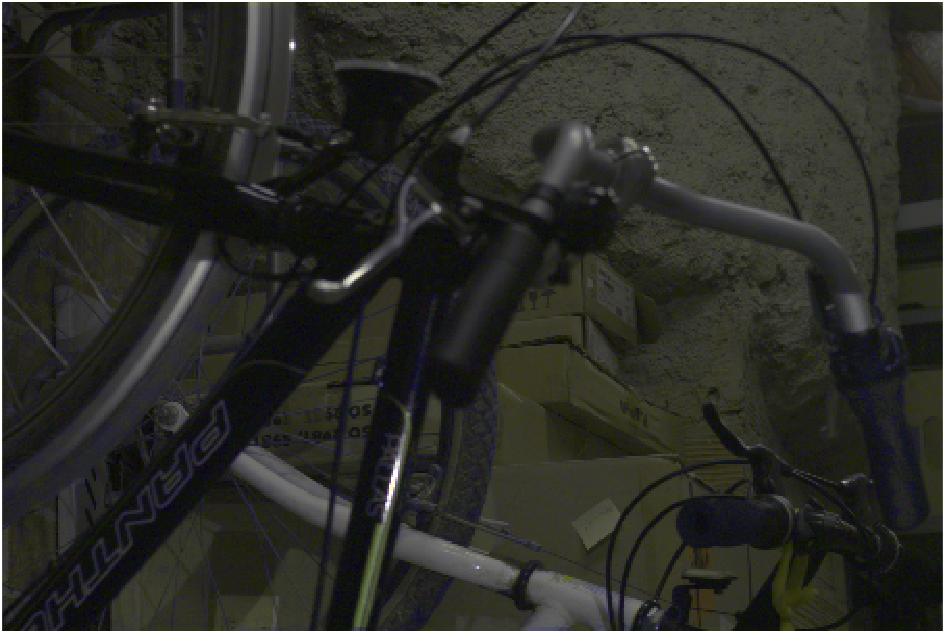}
        &\includegraphics[width=.234\linewidth, height = 1.2cm]{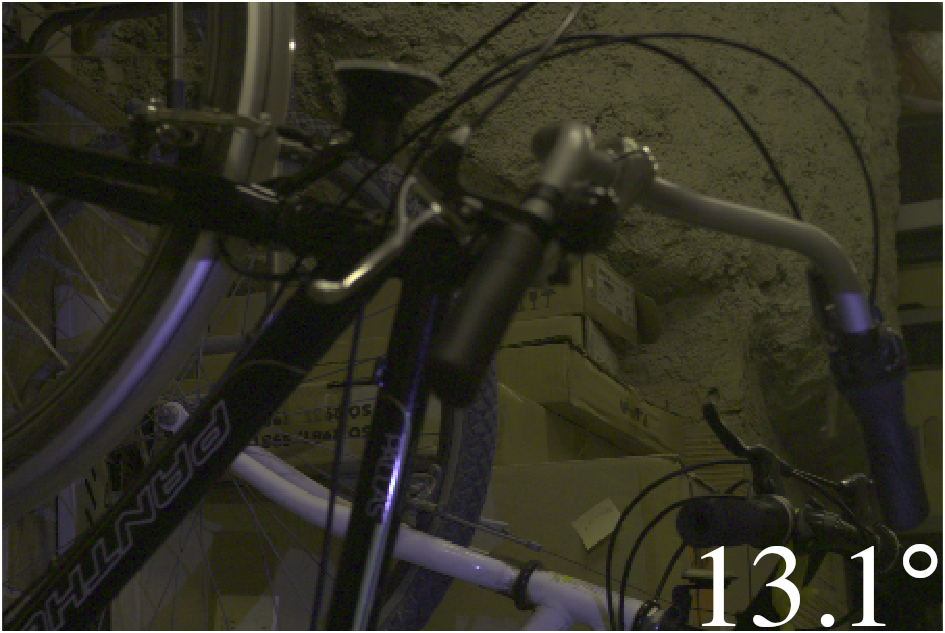} 
        & \includegraphics[width=.234\linewidth, height = 1.2cm]{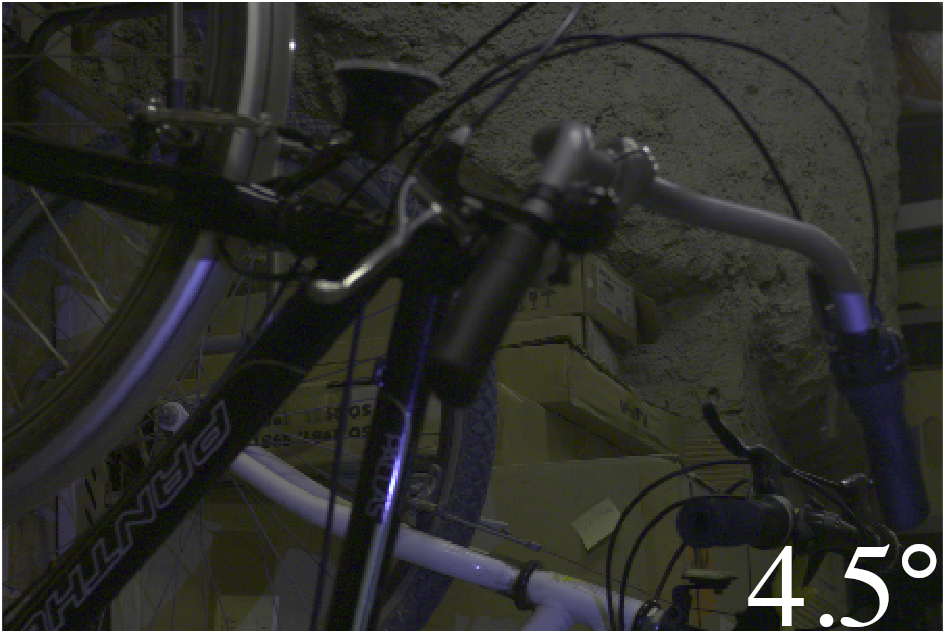}
         \\ 
        \includegraphics[width=.234\linewidth, height = 1.2cm]{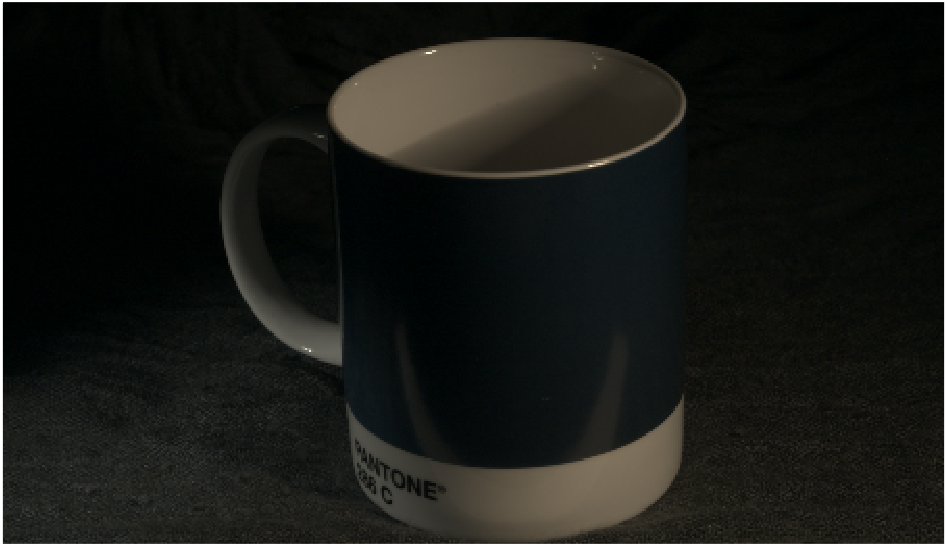} 
        & \includegraphics[width=.234\linewidth, height = 1.2cm]{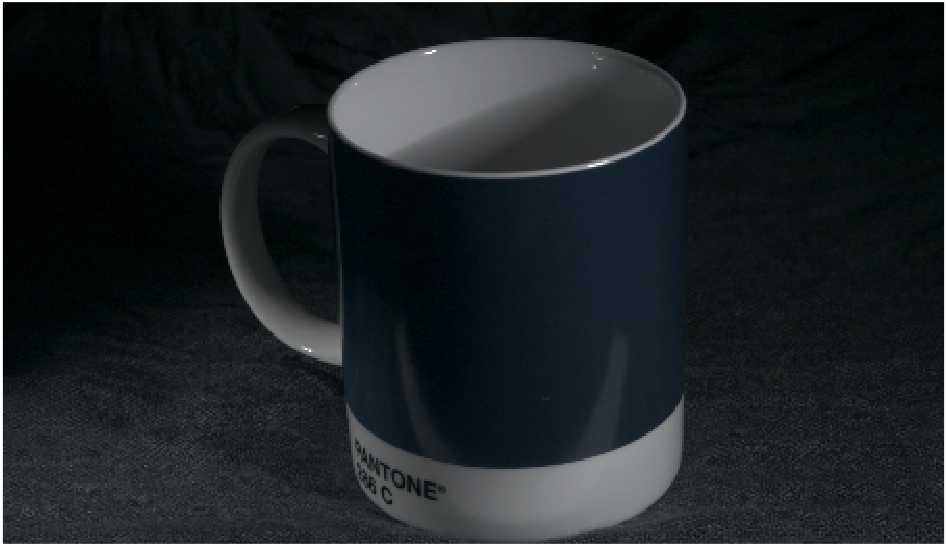}
        &\includegraphics[width=.234\linewidth, height = 1.2cm]{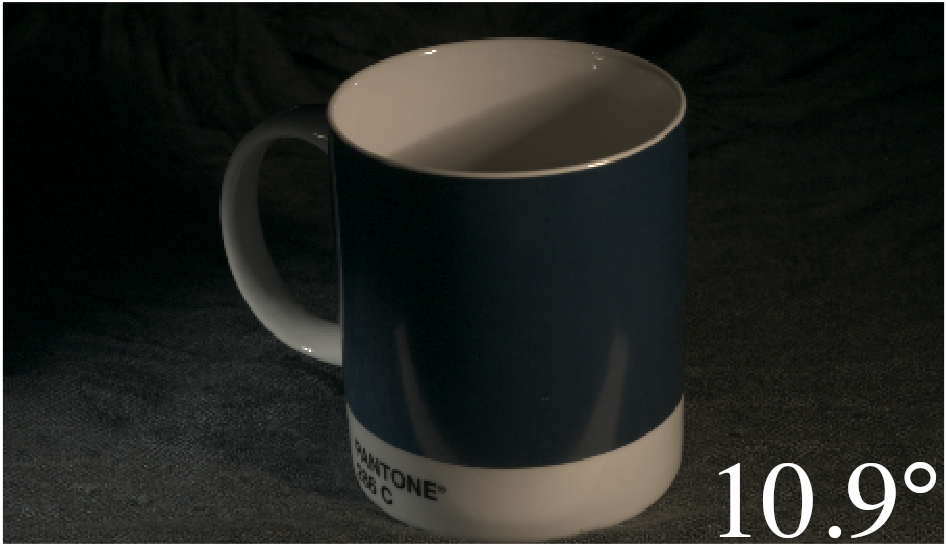}
        & \includegraphics[width=.234\linewidth, height = 1.2cm]{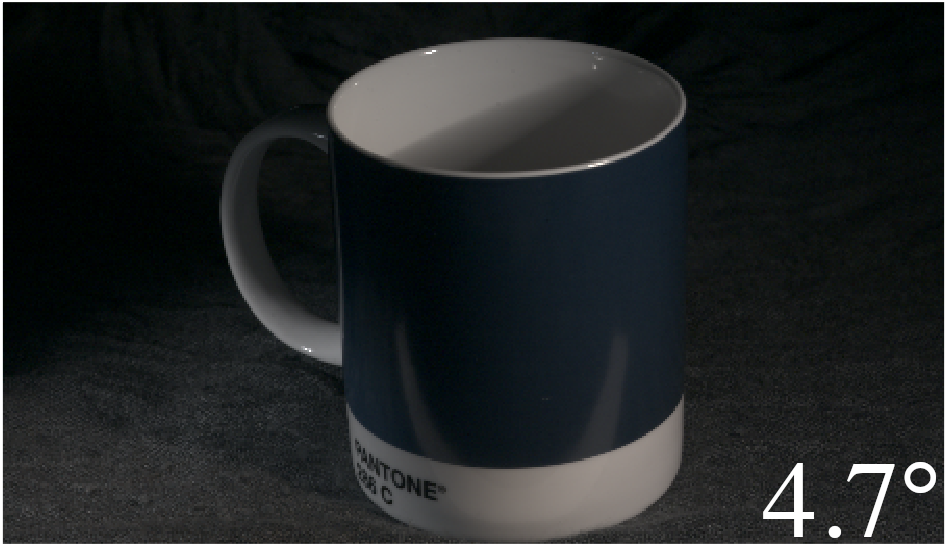}
         \\  
         \includegraphics[width=.234\linewidth, height = 1.2cm]{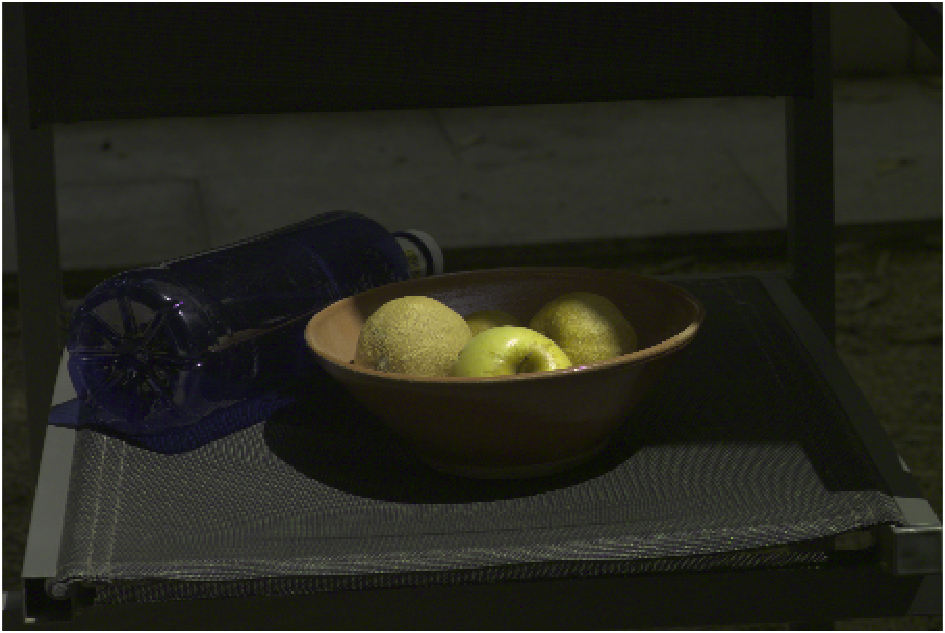} 
        & \includegraphics[width=.234\linewidth, height = 1.2cm]{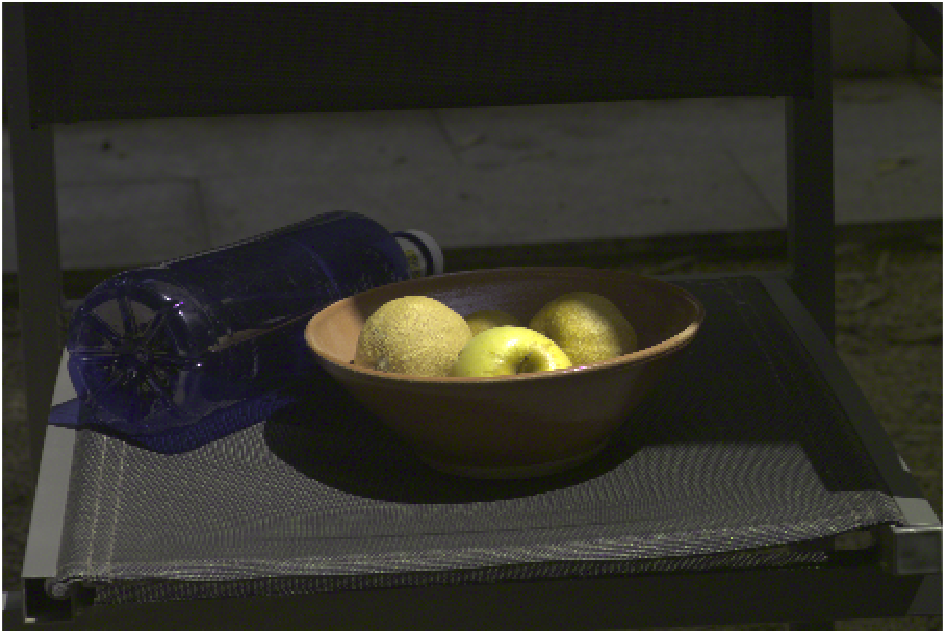}
        &\includegraphics[width=.234\linewidth, height = 1.2cm]{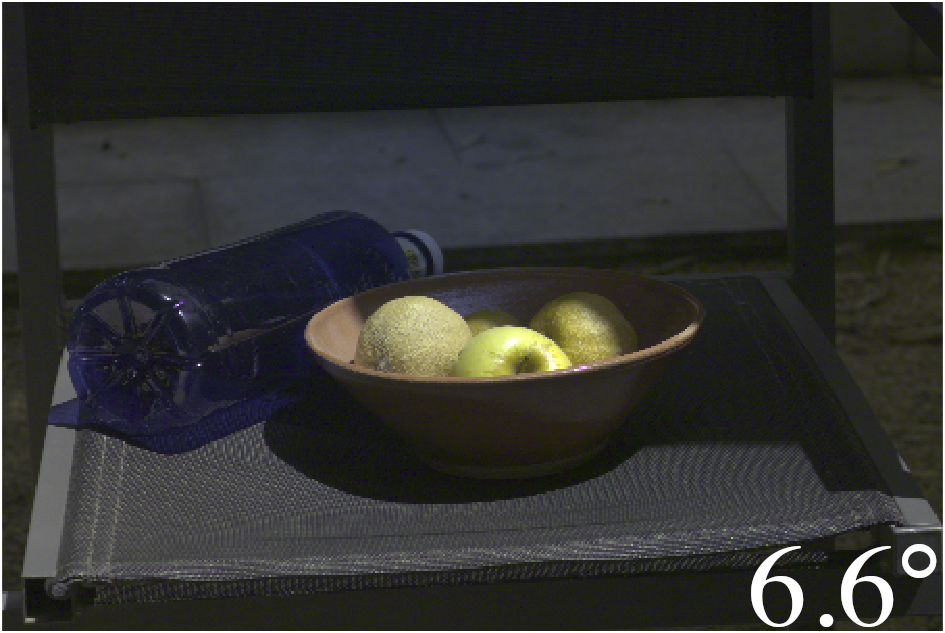}
        & \includegraphics[width=.234\linewidth, height = 1.2cm]{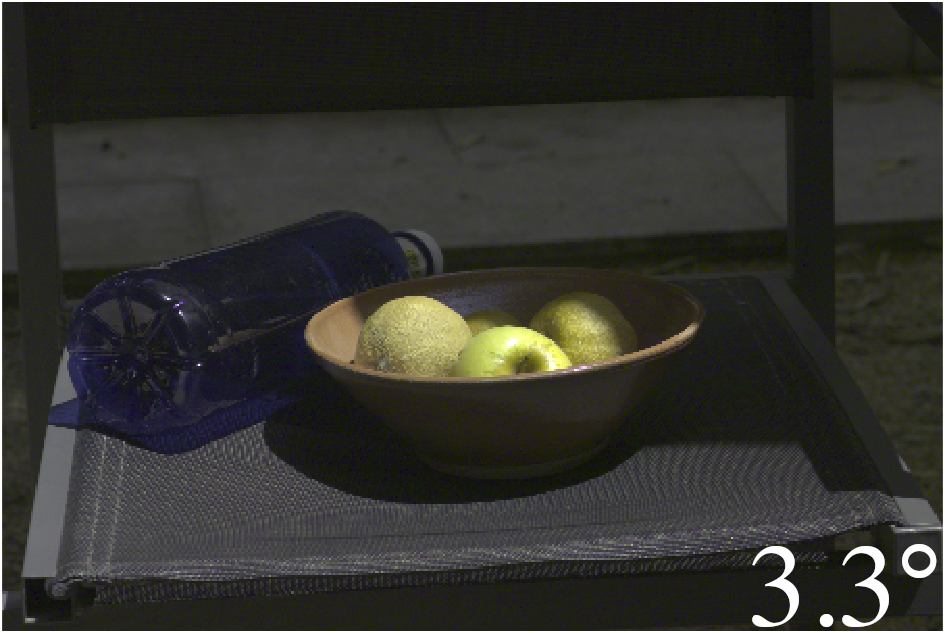}
        \\ 
        \includegraphics[width=.234\linewidth, height = 1.2cm]{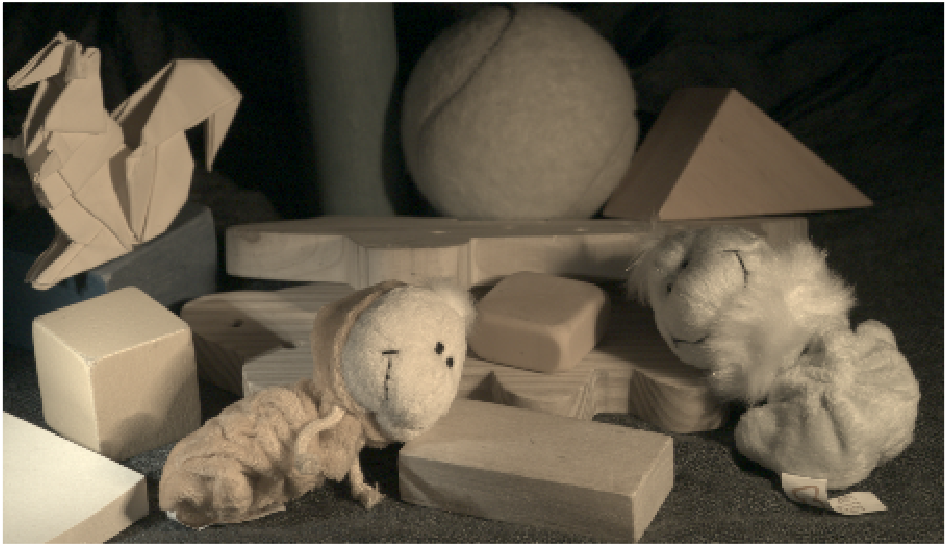} 
        & \includegraphics[width=.234\linewidth, height = 1.2cm]{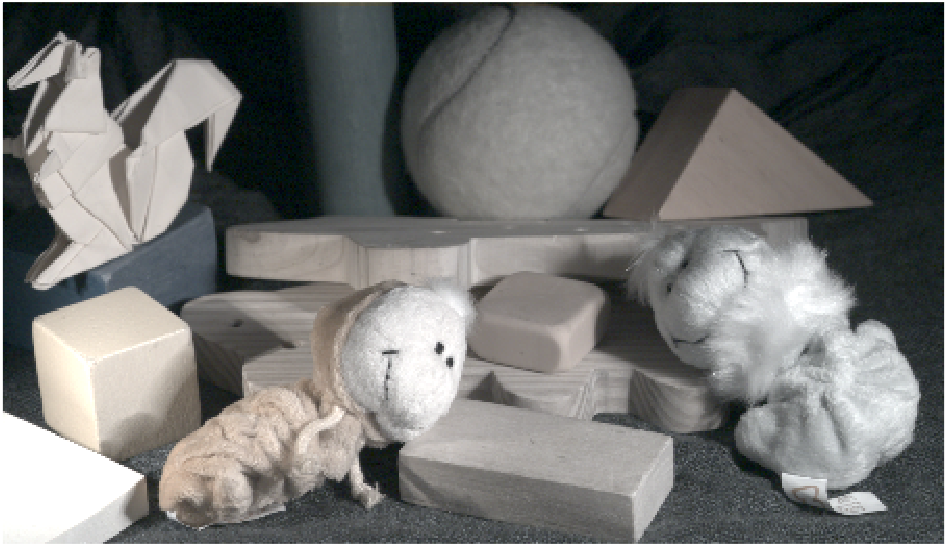}
        &\includegraphics[width=.234\linewidth, height = 1.2cm]{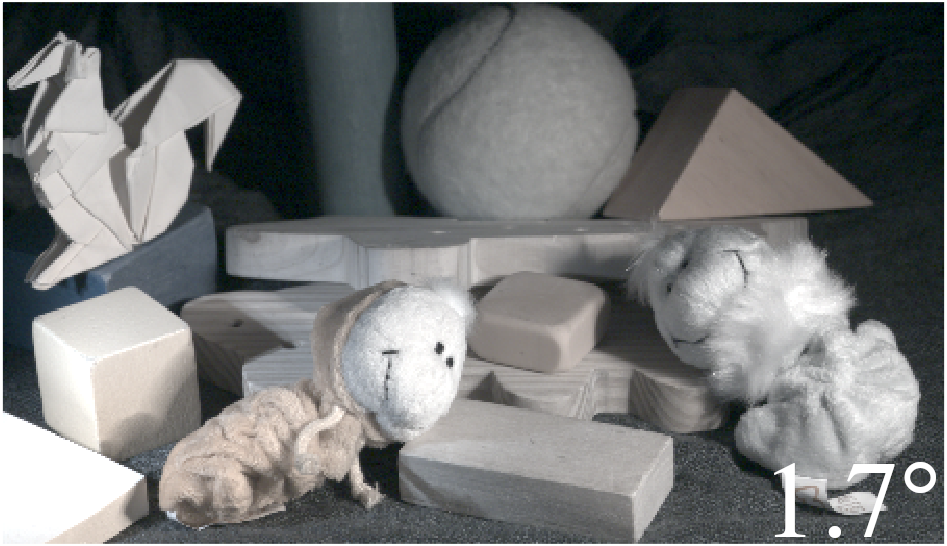}
        & \includegraphics[width=.234\linewidth, height = 1.2cm]{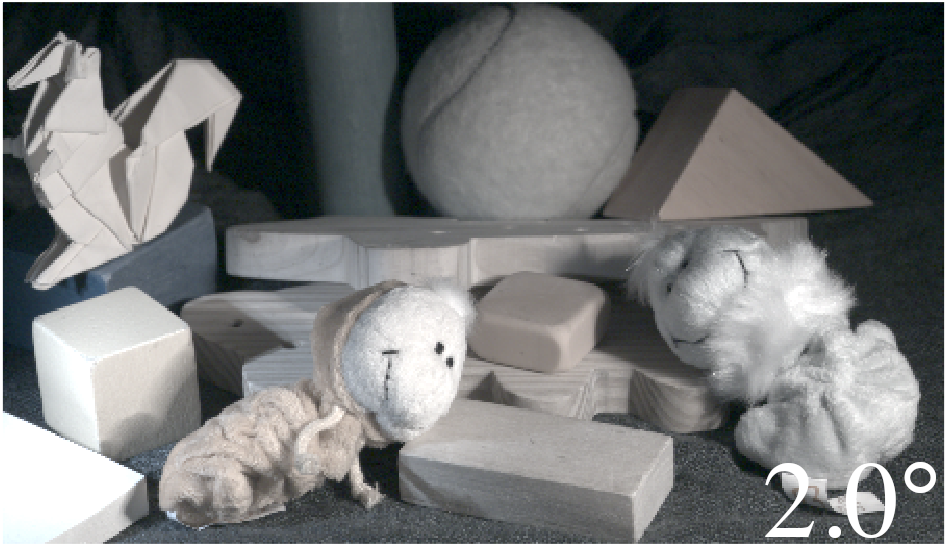}
    \end{tabular}
    \caption{Random examples from the MIMO dataset. (First two rows: Left-to-right) Input, ground truth, the result of white-patch Retinex, and proposed method with white-patch Retinex. (Last two rows: Left-to-right) Input, ground truth, the result of gray pixels, and proposed method with shades of gray. Errors are given at the bottom right corner of the images.}
    \label{fig:multi_comp_vis}
\end{figure}

After we investigate the parameters obtained from illusions, we analyze our method's effectiveness on color constancy benchmarks. The method we developed based on our observations on color assimilation illusions is able to transform global color constancy methods into algorithms, which can provide pixel-wise illumination estimates (Table~\ref{tab:table_mimo}). As we mentioned, our method does not rely on prior information or utilize any learning-based techniques, i.e., clustering or segmentation methods. We see the independence of prior information and having no requirement for a high amount of data as an advantage of our method. Since as pointed out in several color constancy studies we usually cannot have information about the number of illuminants illuminating the scene, and for multi-illuminant color constancy, it is very challenging to form a large-scale dataset with ground truth illumination~\cite{Gijsenij/Lu/Gevers:2011,Das/Liu/Karaoglu/Gevers:2021}. 

The results of our experiments coincide with our discussion that the relationship between color assimilation illusions and color constancy should be taken into account. A simple approach that can reproduce our sensation on color assimilation illusions is able to output pixel-wise estimates for multi-illuminant color constancy (Fig.~\ref{fig:all_mimo_set}). The modified algorithms utilizing the parameters obtained from color illusions present very competitive results when compared with methods particularly designed for multi-illuminant color constancy (Fig.~\ref{fig:multi_comp_vis}), while they clearly outperform several of them, i.e., the proposed approach with white-patch Retinex is able to outperform most of the algorithms specifically designed for multi-illuminant purposes. Another noteworthy outcome is that our proposed approach enhances existing algorithms with more efficiency on the real-world set on average compared to other similar strategies such as Gijsenij~\textit{et al.}, conditional random fields, and N-white balancing (Table~\ref{tab:table_mimo}). For the laboratory set, our outcomes do not produce the best results but they are competitive. The reason behind this can be explained by the block size and controlling parameter of the Gaussian kernel of our approach. These parameters are determined by investigating color assimilation illusions that contain frequent color changes as the real-world set of the MIMO dataset. On the other hand laboratory images do not contain as much complexity as real-world scenes (Fig.~\ref{fig:all_mimo_set}). We would like to emphasize that our learning-free approach is able to output competitive results when compared to learning-based algorithms. Also, it is important to note here that the GAN-based model which has the second-best results on average on the MIMO dataset, uses the illuminants from the MIMO dataset during training.  

One might think that color assimilation illusions and global color constancy are not directly related since global color constancy algorithms provide only a single RGB triplet rather than pixel-wise estimates. Yet, as we argued, our approach for reproducing the behavior of the human visual system on color illusions is able to improve the outcomes of algorithms as presented in Table~\ref{tab:table_global}. The mean angular errors of the algorithms we modified are able to compete with the state-of-the-art. Furthermore, the errors of the worst cases decrease substantially, especially for the algorithms using the maximum intensities as illumination estimates of the scene (Fig.~\ref{fig:single_comp_vis}). Additionally, the lowest errors for the worst cases are obtained via our approach on both benchmarks. As pointed out in color constancy studies it is important to enhance the outcomes for the most difficult cases.

\begin{table}
\centering
\caption{Statistical results on global color constancy datasets. The top three results are highlighted.}
\resizebox{\linewidth}{!}{%
\setlength{\tabcolsep}{2pt} 
\renewcommand{\arraystretch}{1.2} 
\begin{tabular}{c l | c c c c c || c c c c c}
\toprule \toprule
\multicolumn{2}{l}{}      
& \multicolumn{5}{c ||}{RECommended ColorChecker} & \multicolumn{5}{c}{INTEL-TAU}
\\  
\multicolumn{2}{l}{Algorithms}      
& \multicolumn{1}{c}{Mean} & \multicolumn{1}{c}{Median} & \multicolumn{1}{c}{B.25\%}  & \multicolumn{1}{c}{W.25\%} & \multicolumn{1}{c ||}{Max.} 
& \multicolumn{1}{c}{Mean} & \multicolumn{1}{c}{Median} & \multicolumn{1}{c}{B.25\%}  & \multicolumn{1}{c}{W.25\%} & \multicolumn{1}{c}{Max.} 
\\ \hline 
& White-Patch Retinex   
& $10.2$ & $9.1$ & $1.6$ & $20.4$ 
& $50.5$  & $11.0$ & $13.2$ & $1.8$ & $19.4$ & $43.2$
\\ 
& Gray World  
& $4.7$ & $3.6$ & $0.9$ & $10.4$ & $24.8$  
& $4.9$ & $3.9$ & $1.0$ & $10.6$  & $31.0$
\\ 
& Shades of Gray  
& $5.8$ & $4.2$ & $\mathbf{0.7}$ & $13.7$  & $32.4$ 
& $5.5$ & $4.2$ & $1.0$ & $12.3$ & $35.6$
\\ 
& $1^{st}$ order Gray Edge  
& $6.4$ & $3.8$ & $0.9$ & $15.8$ & $37.2$ 
& $6.1$ & $4.2$ & $1.0$ & $14.3$ & $50.7$
\\
& Weighted Gray Edge     
& $6.1$ & $3.3$ & $\mathbf{0.7}$ & $15.5$  & $33.2$  
& $6.0$ & $3.6$ & $0.8$ & $14.9$ & $38.9$
\\ 
& Double-Opponent Cells based Color Constancy   
& $7.2$ & $4.2$ & $\mathbf{0.7}$ & $18.0$ & $48.5$ 
& $7.2$ & $4.7$ & $0.8$ & $17.0$ & $49.4$
\\
& PCA based Color Constancy  
& $4.1$ & $\mathbf{2.5}$ & $\mathbf{0.5}$ & $10.1$ & $28.7$  
& $4.5$ & $3.0$ & $\mathbf{0.7}$ & $10.6$ & $36.8$
\\ 
& Local Surface Reflectance Estimation  
& $4.7$ & $3.7$ & $1.5$ & $9.7$ & $23.5$  
& $4.2$ & $3.4$ & $1.0$ & $8.6$ & $\mathbf{31.4}$
\\ 
& Mean Shifted Gray Pixels  
& $3.8$ & $\mathbf{2.9}$ & $\mathbf{0.7}$ & $8.3$  & $19.4$ 
& $3.6$ & $2.6$ & $\mathbf{0.6}$ & $8.2$ & $37.6$
\\
& Gray Pixels       
& $3.1$  & $\mathbf{1.9}$  & $\mathbf{0.4}$  & $8.0$  & $28.7$ 
& $\mathbf{3.3}$  & $\mathbf{2.2}$   & $\mathbf{0.5}$  & $8.0$ & $31.5$
\\
& Biologically Inspired Color Constancy    
& $4.4$  & $3.3$   & $0.8$  & $9.8$  & $23.4$  
& $4.1$  & $3.1$   & $\mathbf{0.7}$  & $9.4$ & $31.5$
\\
& Block-based Color Constancy    
& $3.8$  & $3.1$   & $1.4$  & $7.3$ & $20.5$ 
& $4.2$  & $3.6$   & $1.2$  & $8.5$ & $\mathbf{23.9}$
\\
& Color Constancy Convolutional Autoencoder    
& $\mathbf{2.1}$  & $\mathbf{1.9}$  & $0.8$  & $\mathbf{4.0}$ & - 
& $\mathbf{3.4}$  & $2.7$  & $0.9$  & $\mathbf{7.0}$ & - 
\\
& Sensor-Independent Color Constancy    
& $\mathbf{2.7}$  & $\mathbf{1.9}$   & $\mathbf{0.5}$  & $\mathbf{6.5}$ & - 
& $\mathbf{3.4}$  & $\mathbf{2.4}$   & $\mathbf{0.7}$  & $\mathbf{7.8}$ & -
\\
& Cross-Camera Convolutional Color Constancy    
& $\mathbf{2.5}$  & $\mathbf{1.9}$   & $\mathbf{0.5}$  & $\mathbf{5.4}$ & -  
& $\mathbf{2.5}$  & $\mathbf{1.7}$   & $\mathbf{0.5}$  & $\mathbf{5.9}$ & - 
\\ 
\hdashline
& Proposed w/ White-Patch Retinex       
& $3.7$ & $3.2$ & $1.1$  & $7.5$ & $\mathbf{17.9}$  
& $3.9$ & $3.0$ & $0.9$  & $8.2$ & $\mathbf{31.2}$ 
\\ 
& Proposed w/ Gray World       
& $4.2$ & $3.5$ & $0.9$  & $8.7$ & $19.5$ 
& $4.2$ & $3.2$ & $0.8$  & $9.2$ & $32.0$
\\ 
& Proposed w/ Shades of Gray       
& $4.0$ & $3.4$ & $0.9$  & $8.3$ & $\mathbf{19.0}$ 
& $4.0$ & $3.1$ & $\mathbf{0.7}$  & $8.9$ & $32.0$
\\ 
& Proposed w/ $1^{st}$ order Gray Edge  
& $4.1$ & $3.4$ & $0.9$  & $8.4$ & $\mathbf{19.2}$  
& $4.1$ & $3.1$ & $\mathbf{0.7}$  & $9.0$ & $31.9$
\\ 
& Proposed w/ Weighted Gray Edge  
& $4.0$ & $3.4$ & $0.9$  & $8.4$ & $\mathbf{19.2}$ 
& $4.1$ & $3.1$ & $\mathbf{0.7}$  & $9.0$ & $31.9$
\\ 
& Proposed w/ Double-Opponent Cells based Color Constancy  
& $4.0$ & $3.4$ & $1.0$  & $8.3$ & $\mathbf{19.0}$ 
& $4.0$ & $3.0$ & $\mathbf{0.7}$  & $8.8$ & $31.6$ \\ \hline \hline
\end{tabular}}
\label{tab:table_global}
\end{table}

\begin{figure}
    \centering
    \setlength{\tabcolsep}{1.5pt} 
    \renewcommand{\arraystretch}{0.5} 
    \begin{tabular}{c c c c}
         \includegraphics[width=.234\linewidth, height = 1.25cm]{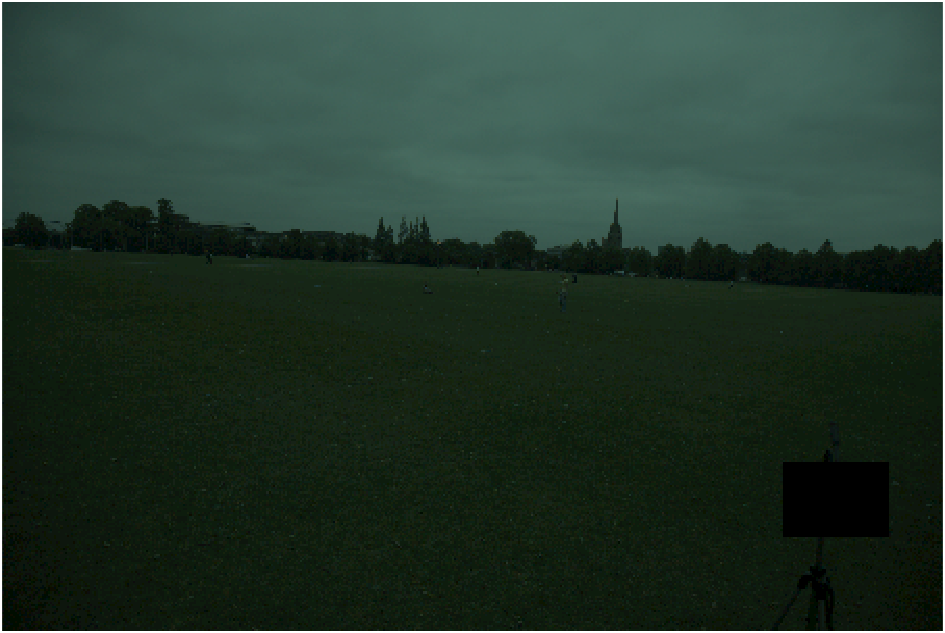} 
        &\includegraphics[width=.234\linewidth, height = 1.25cm]{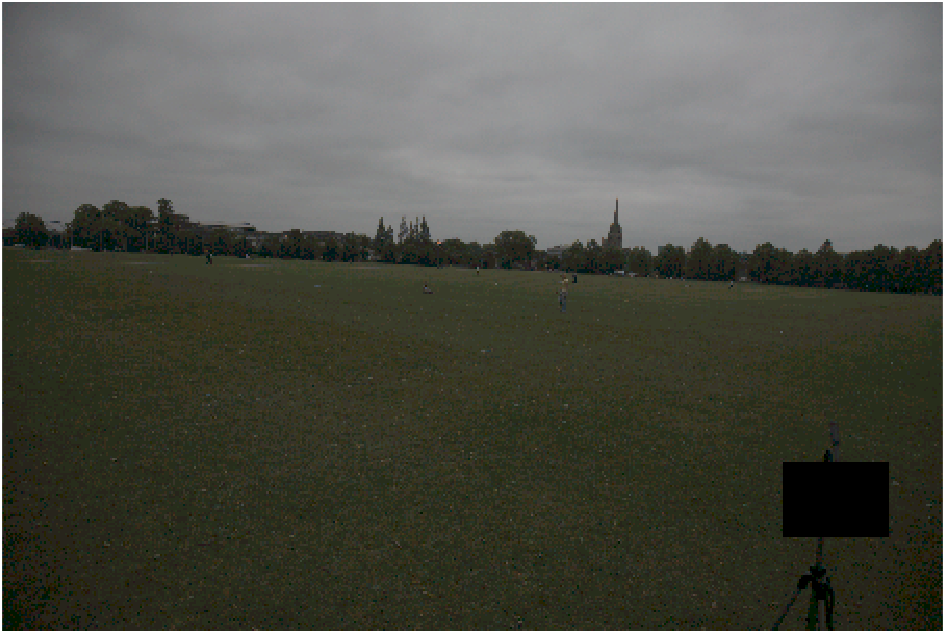}
        &\includegraphics[width=.234\linewidth, height = 1.25cm]{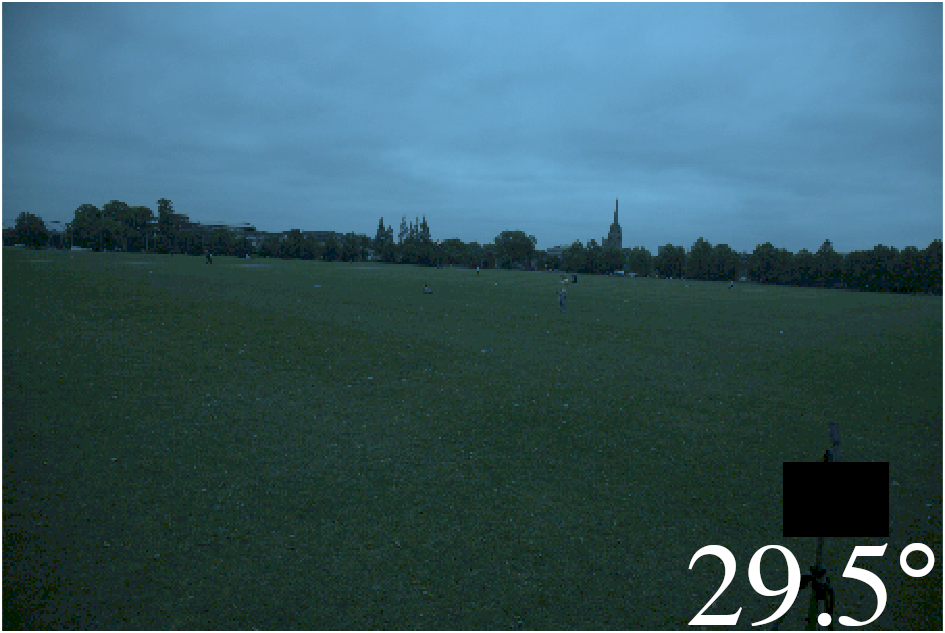}
        &\includegraphics[width=.234\linewidth, height = 1.25cm]{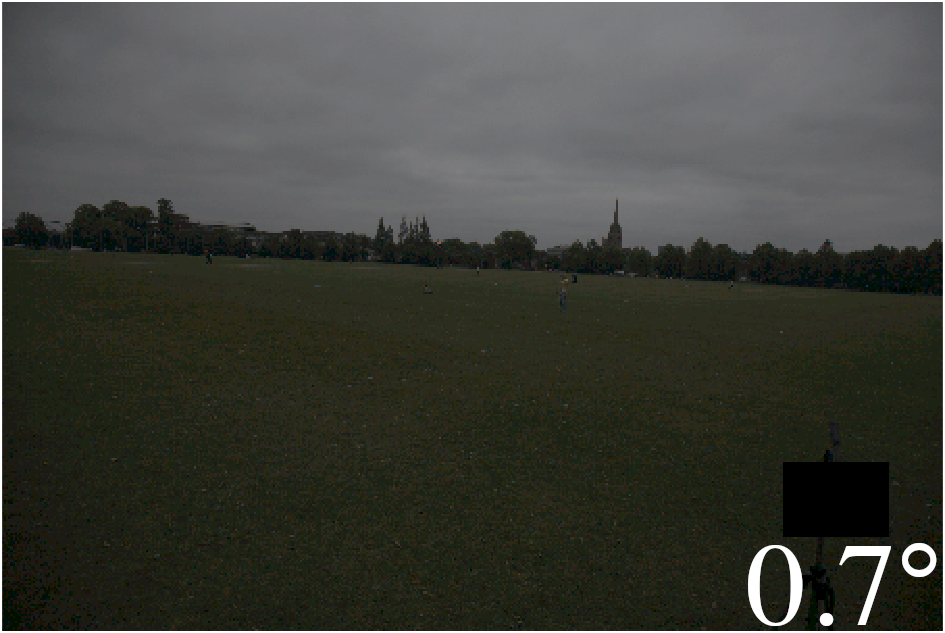}
        \\ 
        \includegraphics[width=.234\linewidth, height = 1.25cm]{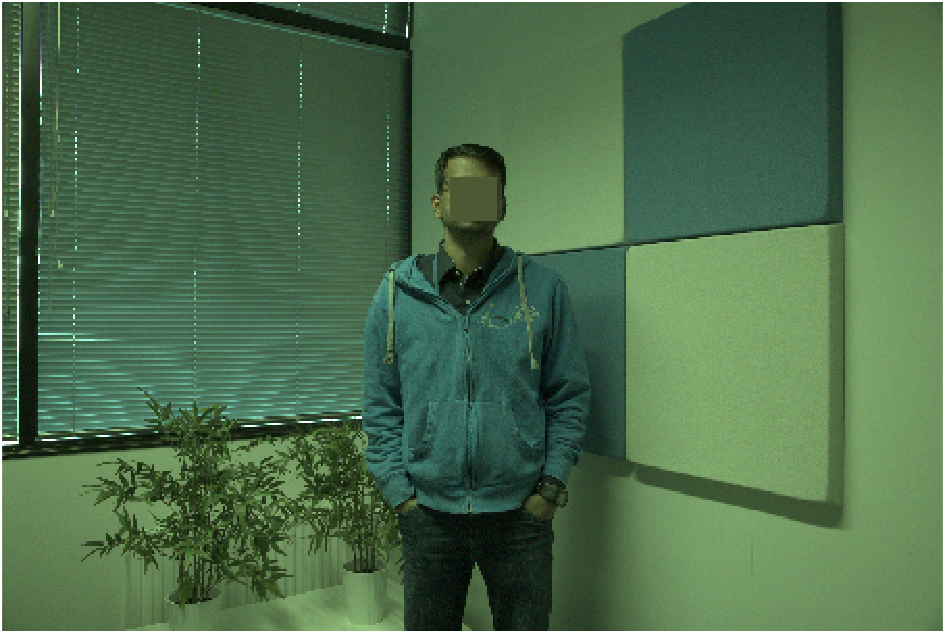} 
        &\includegraphics[width=.234\linewidth, height = 1.25cm]{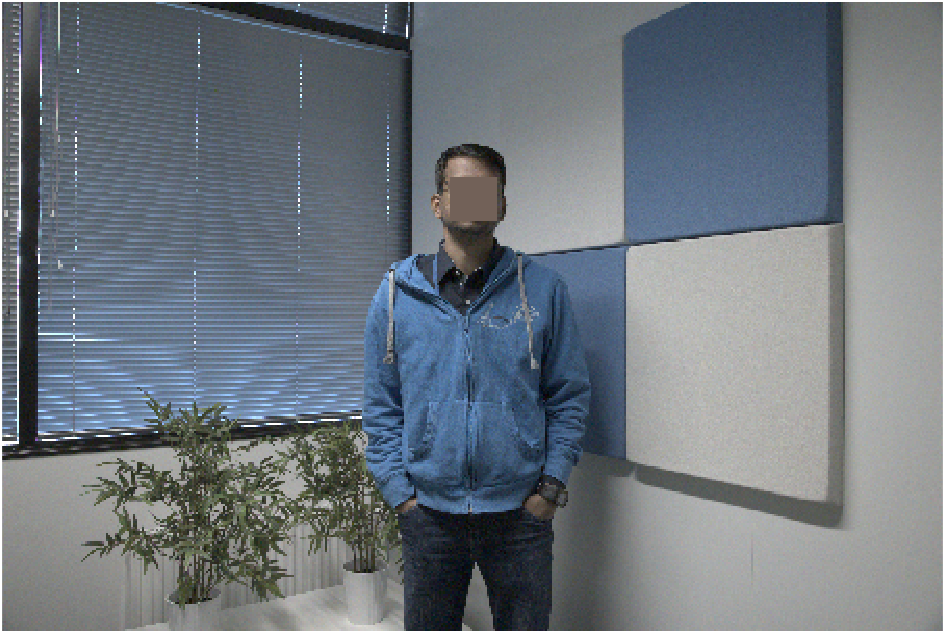} 
        &\includegraphics[width=.234\linewidth, height = 1.25cm]{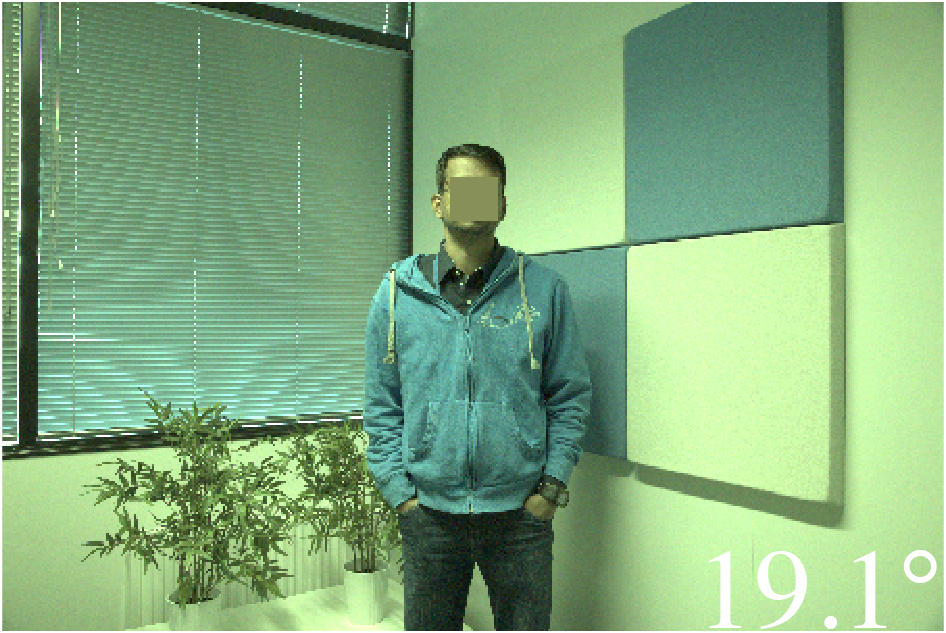}
        &\includegraphics[width=.234\linewidth, height = 1.25cm]{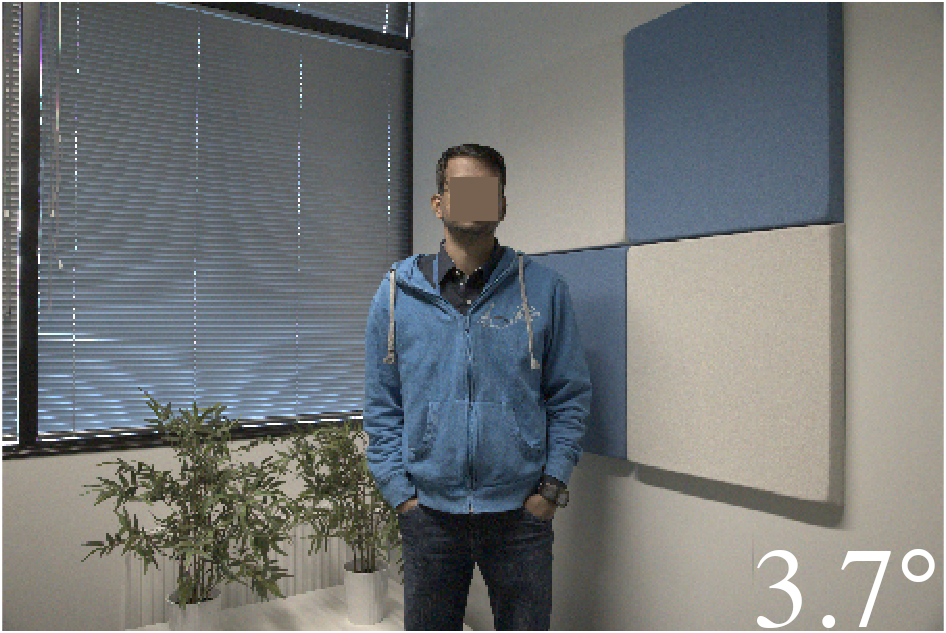}
        \\
         \includegraphics[width=.234\linewidth, height = 1.25cm]{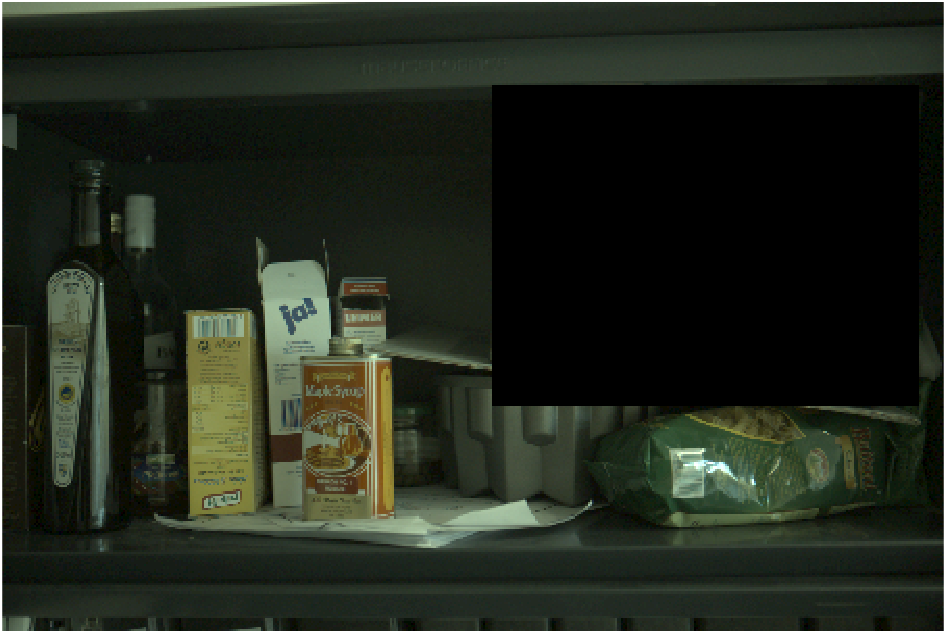} 
        &\includegraphics[width=.234\linewidth, height = 1.25cm]{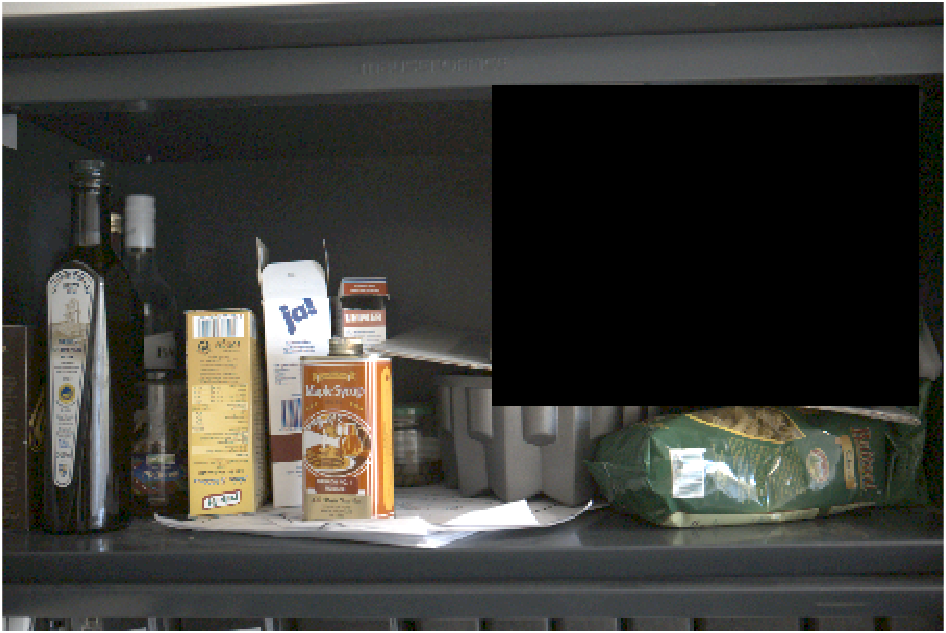} 
        &\includegraphics[width=.234\linewidth, height = 1.25cm]{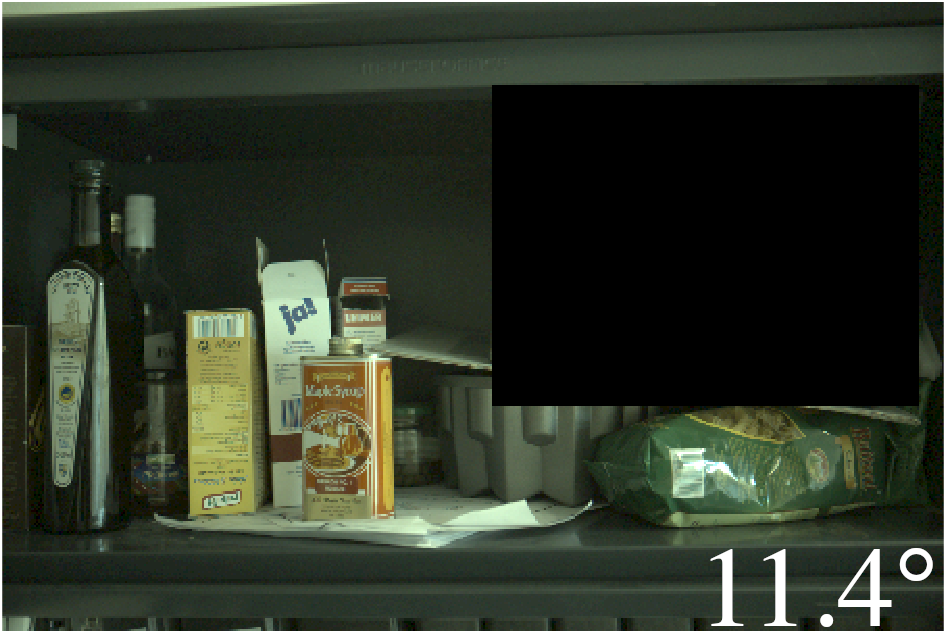}
        &\includegraphics[width=.234\linewidth, height = 1.25cm]{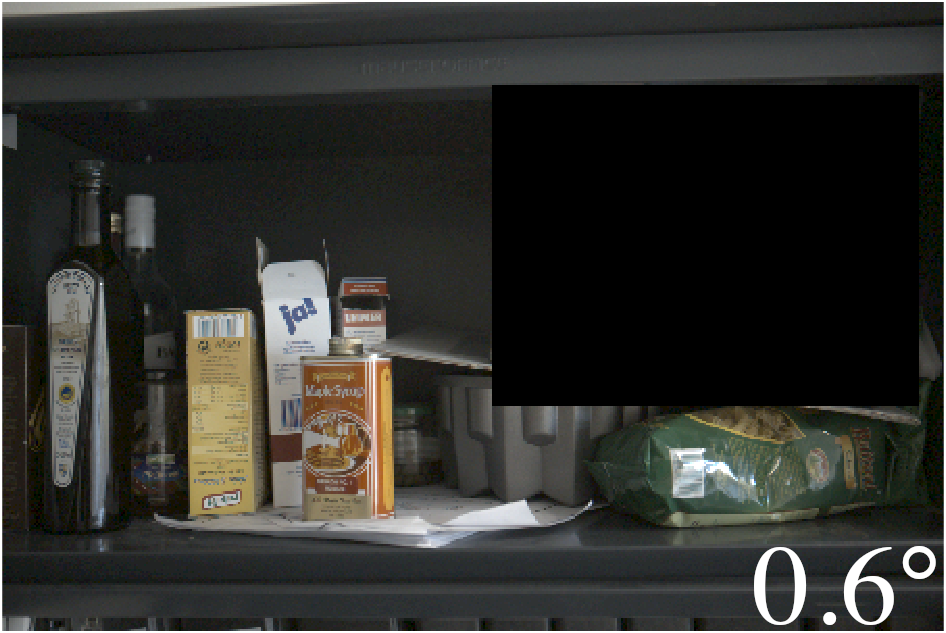}
        \\
        \includegraphics[width=.234\linewidth, height = 1.25cm]{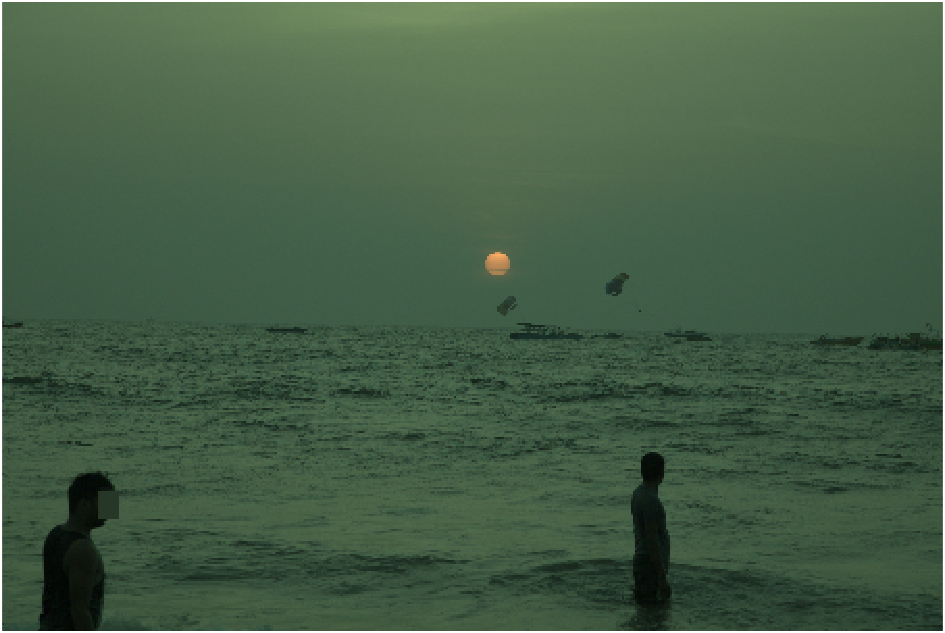} 
        &\includegraphics[width=.234\linewidth, height = 1.25cm]{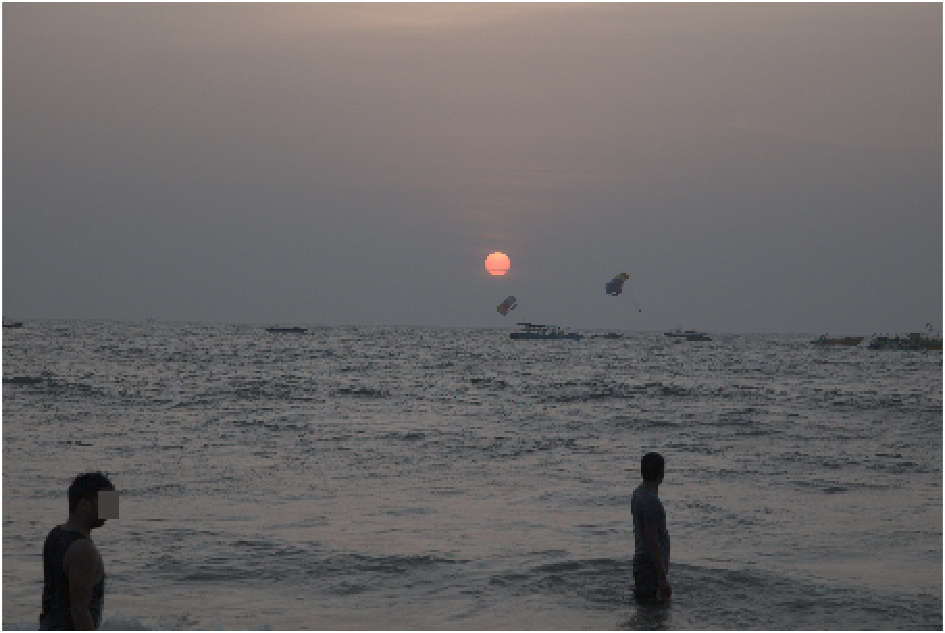} 
        &\includegraphics[width=.234\linewidth, height = 1.25cm]{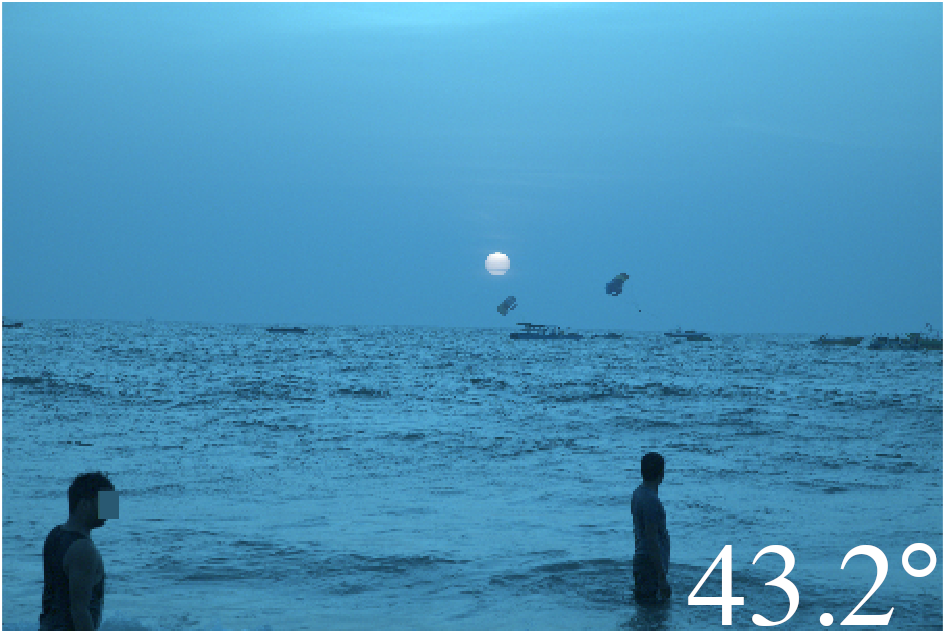}
        &\includegraphics[width=.234\linewidth, height = 1.25cm]{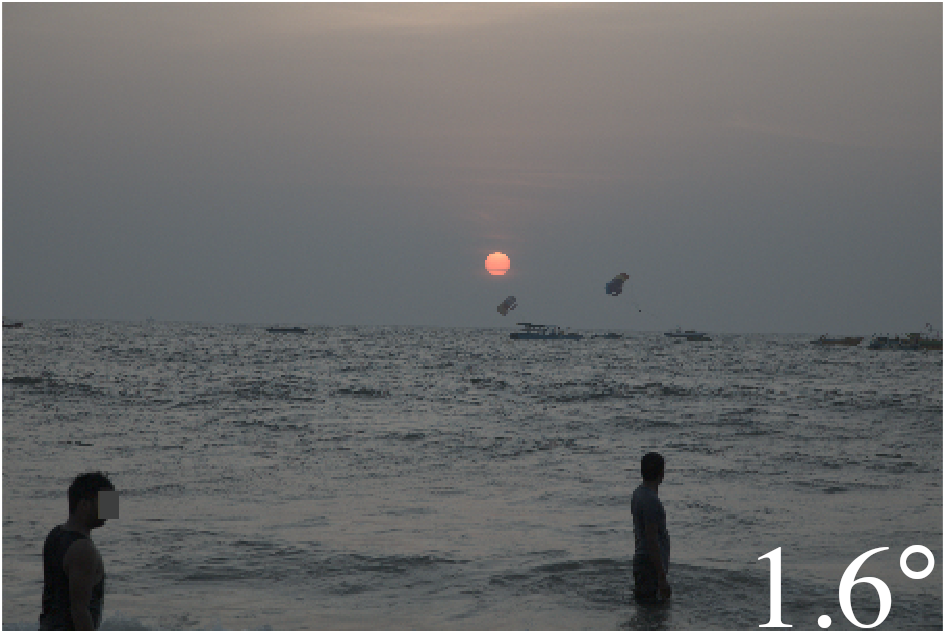}
    \end{tabular}
    \caption{Random worst case examples from global datasets. (First two rows: Left-to-right) Input, ground truth, the result of double-opponent cells based color constancy and proposed method with double-opponent cells based color constancy. (Last two rows: Left-to-right) Input, ground truth, the result of white-patch Retinex and proposed method with white-patch Retinex. Errors are given at the bottom right corner of the images.}
    \label{fig:single_comp_vis}
\end{figure}

As a final note, even though we extract hard parameters and use them to modify learning-free approaches, we are able to improve the methods' efficiency significantly. This outcome might indicate that easily producible color assimilation illusions might be used as input data to train the learning-based color constancy models for mixed-illumination conditions. Since the generated illusions will be free from the sensor specifications of the capturing device, they might prevent some of the current challenges of the learning-based models and benchmarks, i.e., data bias due to the camera sensor specifications and the type of illuminant~\cite{Ulucan/Ulucan/Ebner:2022b,Buzzelli/Zini/Bianco/Ciocca/Schettini/Tchobanou:2023}. When we consider the outcomes of this work, we may deduce that color illusions, especially color assimilation illusions, are valuable for color constancy, hence the connection between these two phenomena should be further investigated minutely. This investigation might lead us to design more robust models aiming at mimicking the abilities of the human visual system.

\section{\uppercase{Conclusion}} \label{sec:conc}
Color illusions and computational color constancy are two phenomena that can help us to reveal the structure of the brain, and design artificial systems that are one step closer to mimicking the human visual system. In this study, we investigated color illusions from the perspective of computational color constancy to find out what we can learn from the link between these two phenomena. We argued that if we design an approach that can reproduce the behavior of the human visual system on color illusions by utilizing global color constancy algorithms, this approach cannot only transform global color constancy methods into multi-illuminant color constancy algorithms but also increase their performance 
on single illuminant benchmarks. Thereupon, we have developed a simple yet effective method by making use of global color constancy algorithms to reproduce the behavior of the human visual system on color assimilation illusions. After we investigated color illusions, we used our findings to see whether our argument is valid or not. According to the experiments, our method is able to transform global color constancy algorithms into multi-illuminant color constancy algorithms, which produce competitive results compared to the state-of-the-art without requiring any prior information about the scene. Also, the effectiveness of global color constancy algorithms modified by our approach increases significantly in single illuminant color constancy benchmarks especially for the worst cases, compared to the original versions of the algorithms. From our findings, we suggest that these two phenomena should be further investigated together while taking more color illusions into account. 

\section*{\uppercase{Acknowledgment}}
This is the preprint version of the work accepted at VISAPP 2024. 

\bibliographystyle{apalike}
{\small
\bibliography{refs}}

\end{document}